\documentclass{article}

    \PassOptionsToPackage{numbers, sort&compress}{natbib}

    \usepackage[final]{neurips_2023}

\usepackage[utf8]{inputenc} % allow utf-8 input
\usepackage[T1]{fontenc}    % use 8-bit T1 fonts
\usepackage[autonum]{shortex}
\usepackage{xcolor}         % colors

\newcommand{\ELBO}[2]{\operatorname{ELBO}\left({#1}||{#2}\right)}

\begin{document}

\title{Embracing the chaos: analysis and diagnosis of numerical instability in variational flows}

\author{
  Zuheng Xu \qquad Trevor Campbell\\
  Department of Statistics\\
  University of British Columbia\\
  \texttt{[zuheng.xu | trevor]@stat.ubc.ca} 
}

\maketitle

\begin{abstract}
In this paper, we investigate the impact of numerical instability on the
reliability of sampling, density evaluation, and evidence lower bound (ELBO)
estimation in variational flows. We first empirically demonstrate that common
flows can exhibit a catastrophic accumulation of error: 
the numerical flow map deviates significantly from the exact map---which affects sampling---and 
the numerical inverse flow map does not accurately recover the initial input---which affects density and ELBO computations. 
Surprisingly though, we find that results produced by flows are
often accurate enough for applications despite the presence of serious numerical instability.
In this work, we treat variational flows as dynamical systems, and leverage \emph{shadowing theory}
to elucidate this behavior via theoretical guarantees 
on the error of sampling, density evaluation, and ELBO estimation.
Finally, we develop and empirically test a diagnostic procedure that can be used to validate results produced by 
numerically unstable flows in practice. 
\end{abstract}

\section{Introduction}\label{sec:introduction}

Variational families of probability distributions play a prominent role in generative modelling
and probabilistic inference. A standard construction of a variational family 
involves passing a simple reference distribution---such as a standard normal---through a sequence
of parametrized invertible transformations, i.e., a \emph{flow} \citep{Tabak13,Rezende15,Papamakarios21,Kobyzev21,Xu22mixflow,Chen22,Caterini18}. 
Modern flow-based variational families have the representational capacity to adapt to highly complex target distributions while 
still providing computationally tractable \iid draws and density evaluations, which in turn enables
convenient training by minimizing the Kullback-Leibler (KL) divergence \citep{Kullback51} 
via stochastic optimization methods \citep{Hoffman13,Kucukelbir17,Ranganath14}. 

In practice, creating a flexible variational flow often requires many flow
layers, as the expressive power of a variational flow generally increases with
the number of transformations \citep{kohler2021smooth,kong2020expressive,Xu22mixflow}. Composing
many transformations in this manner, however, can create numerical
instability \citep{behrmann2021understanding,zhang2021ivpf,Xu22mixflow,gomez17}, i.e.,
the tendency for numerical errors from imprecise
floating-point computations to be quickly magnified along the flow. This accumulation of error can result in
low-quality sample generation \citep{behrmann2021understanding},  
numerical non-invertibility \citep{behrmann2021understanding,zhang2021ivpf},
and unstable training \citep{gomez17}.
But in this work, 
we demonstrate that this is not always the case with a counterintuitive
empirical example: a flow that exhibits severe numerical instability---with error
that grows exponentially in the length of the flow---but which surprisingly still
returns accurate density values, \iid draws, and evidence lower bound (ELBO) estimates.

Motivated by this example, the goal of this
work is twofold: (1) to provide a theoretical understanding of how numerical instability in flows relates
to error in downstream results; and 
(2) to provide a diagnostic procedure such that users of normalizing flows can 
check whether their particular results are reliable in practice.
We develop a theoretical framework that investigates the influence of numerical instability using
\emph{shadowing theory} from dynamical systems \citep{coomes1996shadowing,
palmer2000shadowing, pilyugin2006shadowing, backes2019shadowing}. Intuitively, shadowing theory asserts that 
although a numerical flow trajectory may diverge quickly from its exact counterpart, there often exists another exact trajectory
nearby that remains close to the numerical trajectory (the \emph{shadowing property}). 
We provide theoretical error bounds for three
core tasks given a flow exhibiting the shadowing property---sample generation, density evaluation, and ELBO estimation---that show
that the error grows much more slowly than the error of the trajectory itself.
Our results pertain to both normalizing flows \citep{Rezende15, Papamakarios21} and
mixed variational flows \citep{Xu22mixflow}. We develop a diagnostic procedure for
practical verification of the shadowing property of a given flow.
Finally, we validate our theory and diagnostic procedure on 
MixFlow \citep{Xu22mixflow} on both synthetic and real data examples, 
and Hamiltonian variational flow \citep{Chen22} on synthetic examples.

\paragraph{Related work.} \citet{chang2018reversible} connected the numerical invertibility of deep
residual networks (ResNets) and ODE stability analysis by interpreting ResNets as
discretized differential equations, 
but did not provide a quantitative analysis of 
how stability affects the error of sample generation
or density evaluation. This approach also does not
apply to general flow transformations such as coupling layers.
\citet{behrmann2021understanding} analyzed the numerical inversion error of
generic normalizing flows via bi-Lipschitz continuity for each
flow layer, resulting in error bounds that grow exponentially with flow length.
These bounds do not reflect empirical results in which the error in downstream statistical procedures
(sampling, density evaluation, and ELBO estimation) tends to grow much more slowly.
Beyond the variational literature,  
\citet{bahsoun2014pseudo} investigated statistical properties of numerical
trajectories  by modeling numerical round-off errors as small random
perturbations, thus viewing the numerical trajectory as a sample path of a Markov chain. 
However, their analysis was constrained by the requirement that
the exact trajectory follows a time-homogeneous, strongly mixing
measure-preserving dynamical system, a condition that does not generally hold
for variational flows. 
\citet{tupper2009relation} 
established a relationship between shadowing and the weak distance between
inexact and exact orbits in dynamical systems.  
Unlike our work, their results are limited to quantifying the quality of
pushforward samples and do not extend to the evaluation of densities and ELBO
estimation. In contrast, our error analysis provides a more comprehensive
understanding of the numerical error of all three downstream statistical procedures.

\section{Background: variational flows}\label{sec:background}
A \emph{variational family}
is a parametrized set of probability distributions $\{q_\lambda : \lambda \in \Lambda\}$ on some space $\mcX$
that each enable tractable \iid sampling and density evaluation. The ability to obtain draws and compute density values
is crucial for fitting the family to data via maximum likelihood in generative modelling, as well as maximizing the ELBO
in variational inference,
\[
\ELBO{q_\lambda}{p} &=\int q_\lambda(x)\log\frac{p(x)}{q_\lambda(x)}\dee x,\label{eq:klmin}
\]
where $p(x)$ is an unnormalized density corresponding to a target probability distribution $\pi$.
In this work we focus on 
$\mcX \subseteq \reals^d$ endowed with its Borel $\sigma$-algebra and Euclidean norm $\|\cdot\|$, and parameter set 
$\Lambda \subseteq\reals^p$. We assume all distributions have densities with respect 
to a common base measure on $\mcX$, and will use the same symbol to denote a distribution and its
density. Finally, we will suppress the parameter subscript $\lambda$, since we consider 
a fixed member of a variational family throughout.

\paragraph{Normalizing flows.} One approach to building a variational distribution $q$
is to push a reference distribution, $q_0$,
through a measurable bijection on $\mcX$.
To ensure the function is flexible yet still enables tractable computation,
it is often constructed via the composition of simpler measurable, invertible ``layers'' $F_1, \dots, F_N$,
referred together as a (normalizing) \emph{flow} \citep{Tabak13,Rezende15,Kobyzev21}.
To generate a draw $Y\distas q$, we draw $X \distas q_0$ and evaluate 
$Y = F_N\circ F_{N-1}\circ \dots \circ F_1(X)$.
When each $F_n$ is a diffeomorphism,\footnote{A differentiable map with a differentiable inverse; we assume all flow maps in this work are diffeomorphisms.} 
the density of $q$ is
\[
	\forall x&\in\mcX & q(x) &= \frac{q_0(F_1^{-1}\circ \dots \circ F_N^{-1}(x))}{\prod_{n=1}^N J_n(F^{-1}_n\circ \dots \circ F_N^{-1}(x))}
	& J_n(x) &= \left| \det \nabla_x F_n(x)\right|. \label{eq:transformationvars}
\]
An unbiased ELBO estimate can be obtained using a draw from $q_0$ via 
\[
    X \distas q_0, \,\, E(X) \!=\! \log p(F_N \!\circ \dots \circ\! F_1 (X)) \!-\! \log q_0(X) \!+\! \sum_{n=1}^N \log J_n(F_n \!\circ \dots \circ\! F_1(X)).\label{eq:nfelbo}
\]
Common choices of $F_n$ are  
RealNVP \citep{Dinh17}, neural spline \citep{Durkan19}, Sylvester \citep{vansylvester}, Hamiltonian-based \citep{Chen22,Caterini18},
and planar and radial \citep{Rezende15} flows; see \citet{Papamakarios21} for a comprehensive
overview.

\paragraph{MixFlows.} Instead of using only the final pushforward distribution as the variational
distribution $q$, \citet{Xu22mixflow} proposed averaging over all the pushforwards along a flow
trajectory with identical flow layers,
i.e., $F_n = F$ for some fixed $F$. When $F$ is ergodic and
measure-preserving for some target distribution $\pi$, \citet{Xu22mixflow} show that averaged flows guarantee total variation
convergence to $\pi$ in the limit of increasing flow length. To generate a draw $Y\distas q$, we first draw $X \distas q_0$
and a flow length $K \distas \distUnif\{0, 1, \dots, N\}$, and 
then evaluate $Y = F^K(X) = F\circ F\circ \dots \circ F(X)$ via $K$ iterations of the map $F$.
The density formula is given by the mixture of intermediate flow densities,
\[
\forall x\in\mcX, \qquad q(x) = \frac{1}{N+1}\sum_{n=0}^{N} \frac{q_0(F^{-n}x)}{\prod_{j=1}^n J(F^{-j}x)},  \quad J(x) = \left|\det\nabla_x F(x)\right|. \label{eq:mixflowdensity}
\]
An unbiased ELBO estimate can be obtained using a draw from $q_0$ via
\[\label{eq:averagedelbo}
X \distas q_0, \quad 
E(X)  = \frac{1}{N+1}\sum_{n=0}^{N} \log  \frac{p(F^n X)}{q(F^n X)}.
\]

\section{Instability in variational flows and layer-wise error analysis}
\label{sec:problem}
\captionsetup[subfigure]{labelformat=empty}
\begin{figure*}[t!]
\centering 
\begin{subfigure}[b]{0.33\columnwidth} 
    \scalebox{1}{\includegraphics[width=\columnwidth]{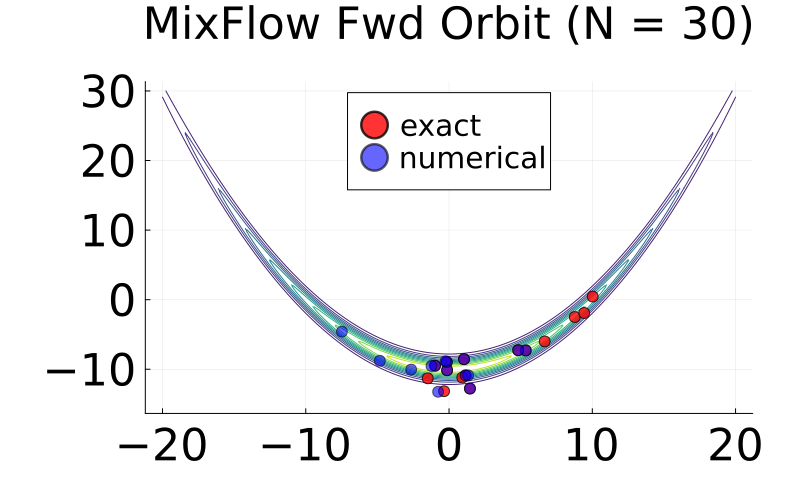}}
    \caption{(a) Banana: num. and exact orbits\label{fig:banana_orbit}}
\end{subfigure}
\hfill
\centering 
\begin{subfigure}[b]{0.32\columnwidth} 
    \scalebox{1}{\includegraphics[width=\columnwidth]{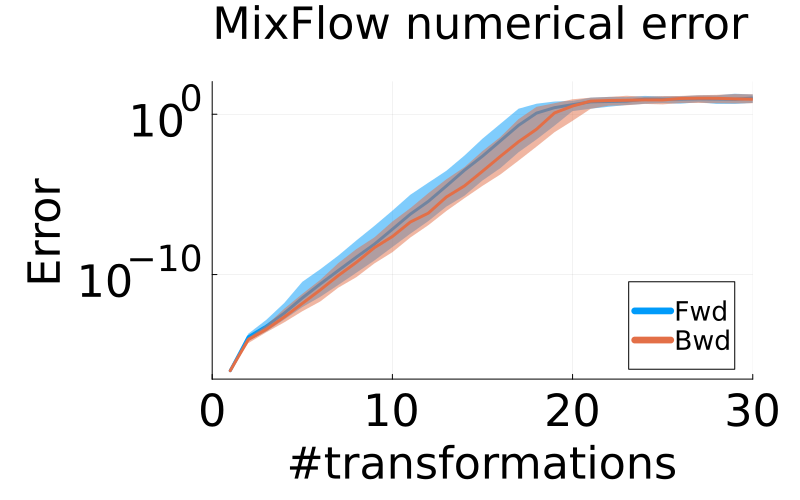}}
    \caption{(b) Banana: flow error \label{fig:banana_err}}
\end{subfigure}
\hfill
\centering 
\begin{subfigure}[b]{0.32\columnwidth} 
    \scalebox{1}{\includegraphics[width=\columnwidth]{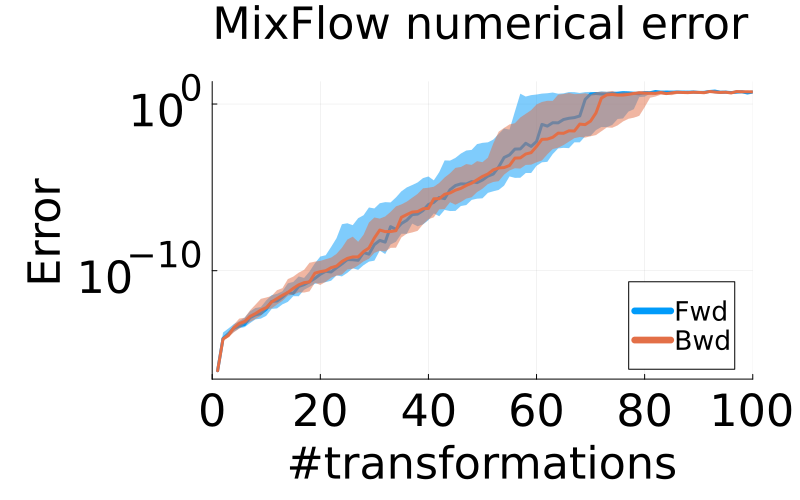}}
    \caption{(c) Lin. Reg.: flow error \label{fig:linreg_flow_err_log}}
\end{subfigure}
\hfill
\centering 
\begin{subfigure}[b]{0.33\columnwidth} 
    \scalebox{1}{\includegraphics[width=\columnwidth]{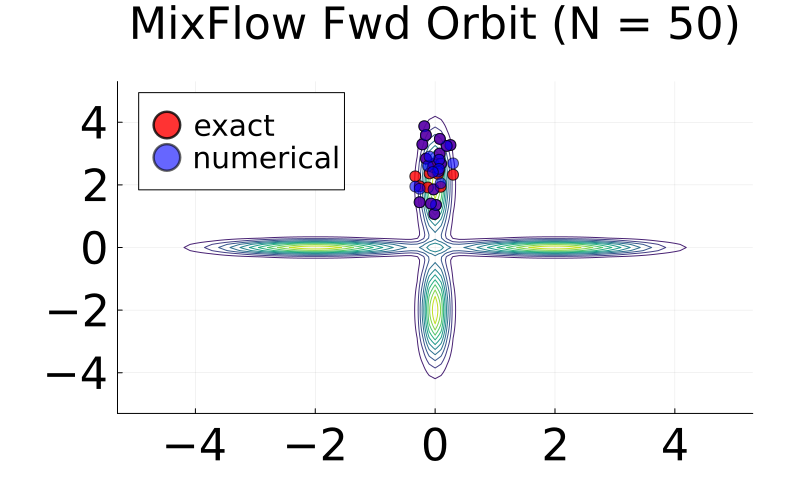}}
    \caption{(d) Cross: num. and exact orbits\label{fig:cross_orbit}}
\end{subfigure}
\hfill
\centering 
\begin{subfigure}[b]{0.32\columnwidth} 
    \scalebox{1}{\includegraphics[width=\columnwidth]{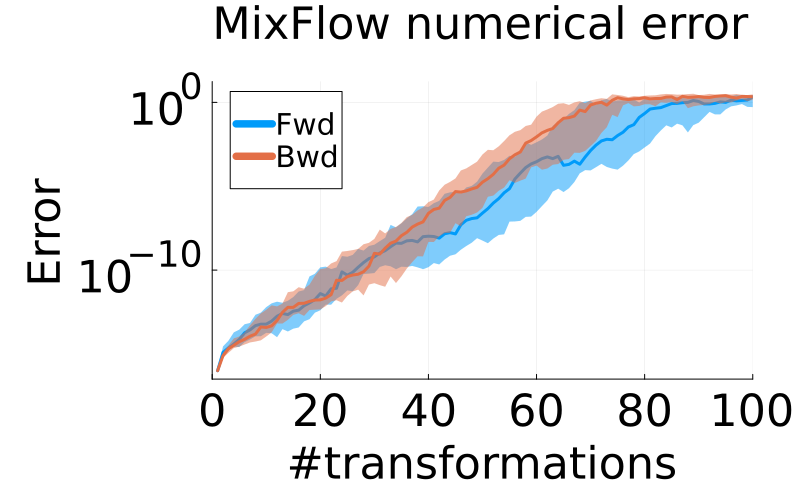}}
    \caption{(e) Cross: flow error \label{fig:cross_err}}
\end{subfigure}
\hfill
\centering 
\begin{subfigure}[b]{0.32\columnwidth} 
    \scalebox{1}{\includegraphics[width=\columnwidth]{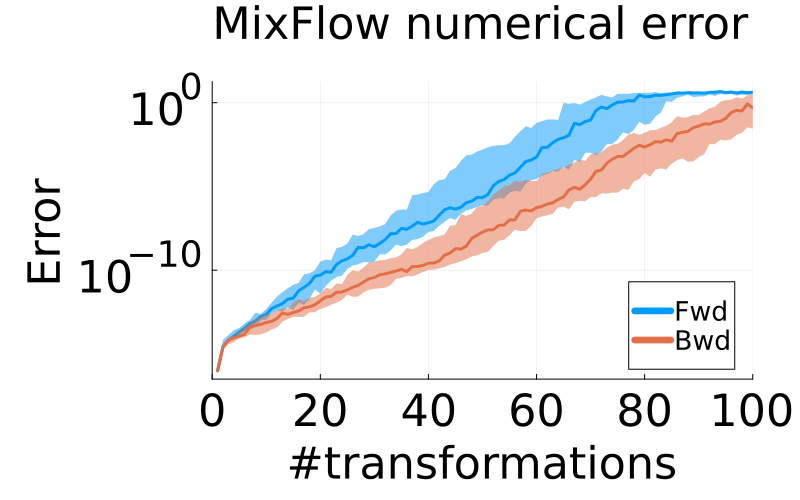}}
    \caption{(f) Log. Reg.: flow error \label{fig:logreg_flow_err_log}}
\end{subfigure}
\hfill
\caption{MixFlow forward (\texttt{fwd}) and backward (\texttt{bwd}) orbit errors
    on the Banana and Cross distributions, and on two real data examples.
    The first column visualizes the \texttt{exact} and \texttt{numerical} orbits with the same
starting point on the synthetic targets. 
For a better visualization, we display every
$2^\text{nd}$ and $4^\text{th}$ states of presented orbits 
(instead of the complete orbits) in \cref{fig:banana_orbit,fig:cross_orbit},
respectively.
\cref{fig:banana_err,fig:linreg_flow_err_log,fig:cross_err,fig:logreg_flow_err_log} show the median and upper/lower quartile 
forward error $\|F^k x - \hF^k x\|$ and backward error $\|B^k x - \hB^{k}x\|$
comparing $k$ transformations of the forward exact/approximate maps $F$, $\hF$
or backward exact/approximate maps $B$, $\hB$.
Statistics are plotted over 100 initialization draws from the reference distribution $q_0$.
For the \texttt{exact} maps we use a 2048-bit \texttt{BigFloat} representation, 
and for the \texttt{numerical} approximations we use 64-bit \texttt{Float} representation.
The ``exactness'' of \texttt{BigFloat} representation is justified in
\cref{fig:mfbigfloaterr} in \cref{apdx:moreresults}.}
\label{fig:trajerr}
\end{figure*}

\captionsetup[subfigure]{labelformat=empty}
\begin{figure*}[t!]
\centering 
\begin{subfigure}[b]{0.49\columnwidth} 
\centering \includegraphics[width=0.7\columnwidth]{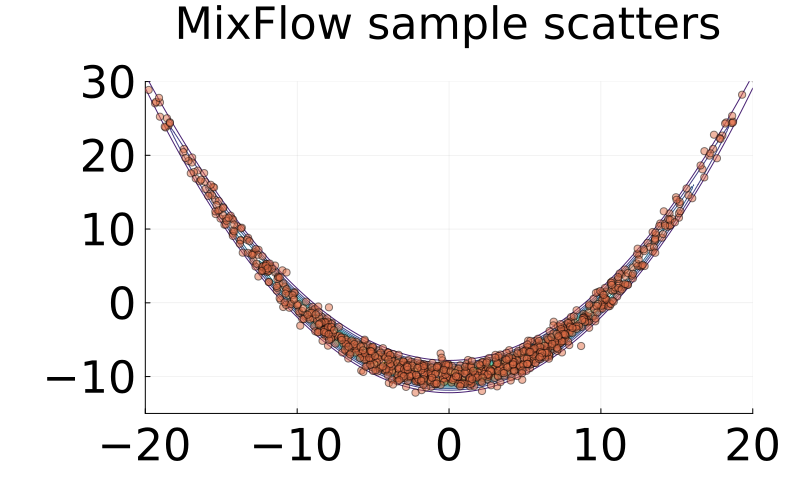}
    \caption{(a) Banana sample scatter\label{fig:banana_scatter}}
\end{subfigure}
\hfill
\centering 
\begin{subfigure}[b]{0.49\columnwidth} 
\centering\includegraphics[width=0.7\columnwidth]{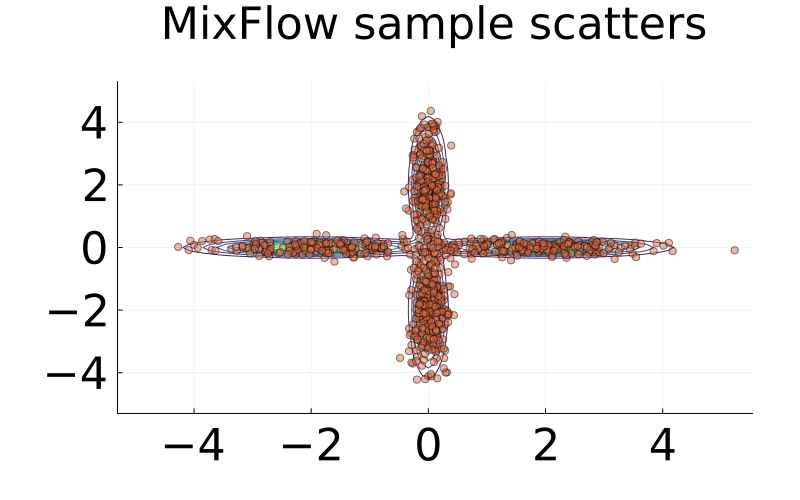}
    \caption{(b) Cross sample scatter\label{fig:cross_scatter}}
\end{subfigure}
\hfill
\centering 
\begin{subfigure}[b]{0.49\columnwidth} 
    \scalebox{1}{\includegraphics[width=0.325\columnwidth]{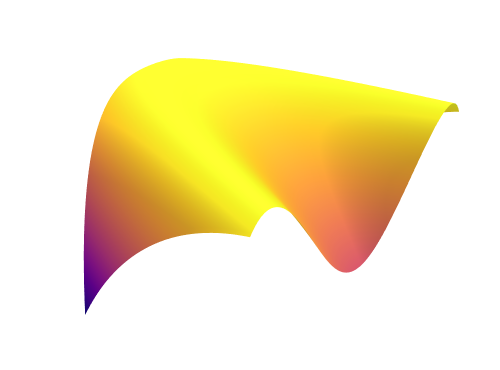}}
    \scalebox{1}{\includegraphics[width=0.325\columnwidth]{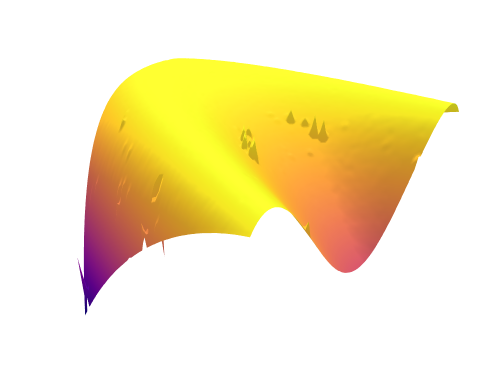}}
    \scalebox{1}{\includegraphics[width=0.325\columnwidth]{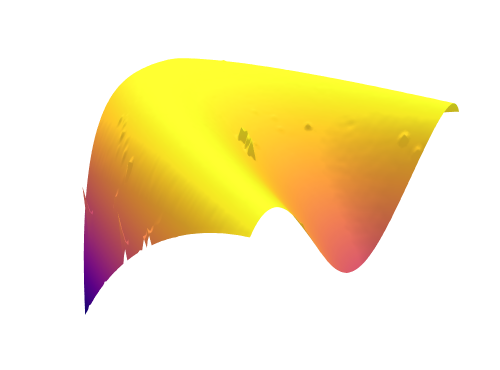}}
    \caption{(c) Banana log-densities\label{fig:banana_lpdf}}
\end{subfigure}
\hfill
\centering 
\begin{subfigure}[b]{0.49\columnwidth} 
    \scalebox{1}{\includegraphics[width=0.325\columnwidth]{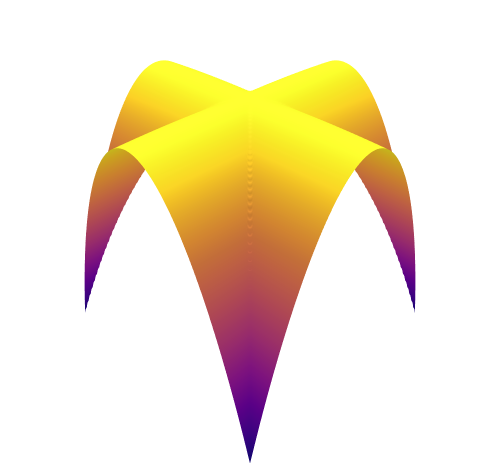}}
    \scalebox{1}{\includegraphics[width=0.325\columnwidth]{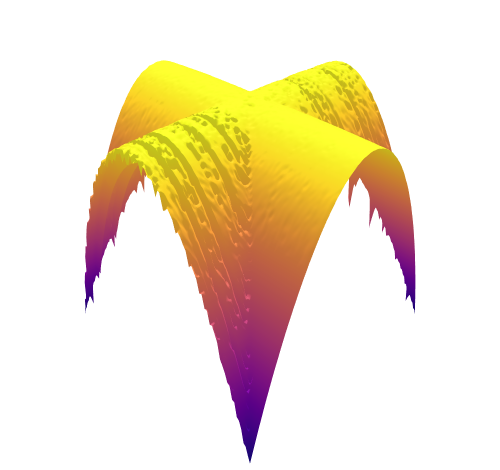}}
    \scalebox{1}{\includegraphics[width=0.325\columnwidth]{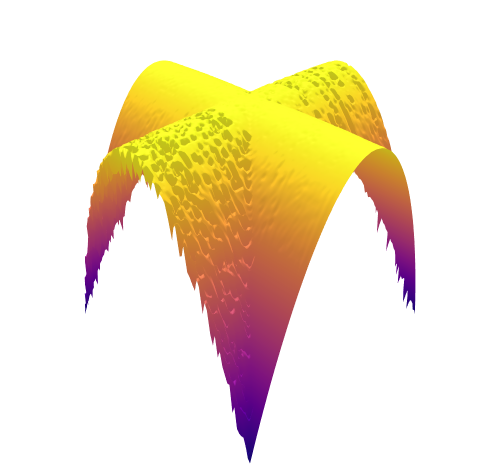}}
    \caption{(d) Cross log-densities\label{fig:cross_lpdf}}
\end{subfigure}
\hfill

\caption{ Figs.~2a and 2b respectively show sample scatters produced by the na\"ive application of MixFlow formulae 
    targeting at the banana and cross distributions, without accounting for numerical error.
  Figs.~2c and 2d display comparisons of log-densities on both synthetic
  examples. The true log-target is
  on the left, the exact MixFlow evaluation (computed via 2048-bit
  \texttt{BigFloat} representation) is in the middle, and the numerical MixFlow
  evaluation is on the right.}
\label{fig:sync_mf_lpdfs}
\end{figure*}

In practice, normalizing flows often involve tens of flow layers, with each
layer coupled with deep neural network blocks \citep{Rezende15,Kobyzev21,Dinh17}
or discretized ODE simulations \citep{Chen22,Caterini18}; past work on MixFlows
requires thousands of flow transformations \citep{Xu22mixflow}. These 
flows are typically chaotic and sensitive to small
perturbations, such as floating-point representation errors. As a result, the
final output of a numerically computed flow can significantly deviate from
its exact counterpart, which may cause downstream issues during training or evaluation of
the flow \citep{behrmann2021understanding,ramasinghe2021robust,kohler2021smooth}.
To understand the accumulation of error, \citet{behrmann2021understanding} assume that 
each flow layer $F_n$ and its inverse are Lipschitz 
and have bounded error, $\sup_x\|F_n(x) - \hF_n(x)\| \leq \delta$,
and analyze the error of each
layer individually. This layer-wise analysis tends to yield error bounds
that grow exponentially in the flow length $N$ (see \cref{sec:layerwiseerror}): 
\[
	\|F_N\circ \dots \circ F_1 (x) - \hF_N \circ \dots \circ \hF_1 (x)\| 
	 \leq \delta\sum_{n=0}^{N-1} \prod_{j=n+2}^{N} \mathrm{Lip}(F_j). \label{eq:layerwisebound}
\]
Given a constant flow map $F_j = F$ with Lipschitz constant $\ell$, the bound is $O(\delta \ell^N)$,
which suggests that the error accumulates exponentially in the length of the flow. 
In \cref{fig:trajerr}, we test this hypothesis empirically using MixFlows on 
two synthetic targets (the banana and cross distribution) and two real data
examples (Bayesian linear regression and logistic problems) taken from past work on MixFlows \cite{Xu22mixflow}. 
See \cref{apdx:expts} for the details of this test.
\cref{fig:banana_err,fig:linreg_flow_err_log,fig:cross_err,fig:logreg_flow_err_log} confirms that the exponential growth in the error bound \cref{eq:layerwisebound} is reasonable; 
the error does indeed grow exponentially quickly in practice. 
And \cref{fig:trajerrallex} (a)--(b) in \cref{apdx:moreresults}  further demonstrate that after fewer than
100 transformations both flows have error on the same order of magnitude as the scale of the target distribution. 
Na\"ively, this implies that sampling, density evaluation, and ELBO estimation may be corrupted badly by numerical error. 

But counterintuitively, in \cref{fig:sync_mf_lpdfs} we find that simply ignoring the issue of numerical error
and using the exact formulae yield reasonable-looking density evaluations and sample scatters. 
These results are of course only qualitative in nature; we provide corresponding quantitative results later in \cref{sec:expts}. But \cref{fig:sync_mf_lpdfs} appears to 
violate the principle of ``garbage in, garbage out;'' 
the buildup of significant numerical error in the sample trajectories themselves does not seem
to have a strong effect on the quality of downstream sampling and density evaluation.
The remainder of this paper is dedicated to resolving this counterintuitive behavior using shadowing theory \cite{pilyugin2011theory}.

\section{Global error analysis of variational flows via shadowing}
\label{sec:theory}

\subsection{The shadowing property}  \label{sec:orbits}
We analyze variational flows as finite discrete-time dynamical systems (see, e.g., \citep{pilyugin2011theory}). 
In this work, we consider a \emph{dynamical system} to be a sequence of diffeomorphisms on $(\mcX, \|\cdot\|)$.
We define the \emph{forward dynamics} $(F_n)_{n=1}^N$ to be the flow layer sequence,
and the \emph{backward dynamics} $(B_n)_{n=1}^N$ to be comprised of the inverted flow layers $B_n = F^{-1}_{N-(n-1)}$ in reverse order.   
An \emph{orbit} of a dynamical system starting at $x\in\mcX$ is the sequence of states produced by the sequence of maps
when initialized at $x$. Therefore, the forward and backward orbits initialized at $x\in\mcX$ are
\[
\text{Forward Orbit:} \quad &x = x_0 \overset{F_1}{\to} x_1 \overset{F_2}{\to} x_2 \to \dots \overset{F_N}{\to} x_N \\
x_{-N}\overset{B_N = F^{-1}_1}{\leftarrow}\dots \leftarrow x_{-2}\overset{B_2 = F^{-1}_{N-1}}{\leftarrow} x_{-1} \overset{B_1 = F^{-1}_N}{\leftarrow}  &x_0 = x \quad \text{: Backward Orbit.}
\]
Given numerical implementations $\hF_n \approx F_n$ and $\hB_n\approx B_n$ with tolerance $\delta >0$, i.e.,
\[
\forall x \in \mcX, \quad \|F_n(x) - \hF_n(x)\| \leq \delta, \quad \|B_n(x) - \hB_n(x)\| \leq \delta, \label{eq:onestepdelta}
\]
we define the \emph{forward} and \emph{backward pseudo-dynamics} to be $(\hF_n)_{n=1}^N$ and $(\hB_n)_{n=1}^N$, respectively,
along with their \emph{forward} and \emph{backward pseudo-orbits} initialized at $x\in\mcX$:
\[
\text{Forward Pseudo-Orbit:} \quad &x = \hx_0 \overset{\hF_1\approx F_1}{\to} \hx_1 \overset{\hF_2\approx F_2}{\to} \hx_2 \to \dots \overset{\hF_N\approx F_N}{\to} \hx_N \\
\hx_{-N}\overset{\hB_N\approx F^{-1}_{1}}{\leftarrow}\dots \leftarrow \hx_{-2}\overset{\hB_{2}\approx F^{-1}_{N-1}}{\leftarrow} \hx_{-1} \overset{\hB_1\approx F^{-1}_{N}}{\leftarrow}  &\hx_0=x \quad \text{: Backward Pseudo-Orbit.}
\]
For notational brevity, we use subscripts on elements of $\mcX$ throughout to denote forward/backward orbit states. 
For example, given a random element $Z\in\mcX$ drawn
from some distribution, $Z_k$ is the 
$k^\text{th}$ state generated by the forward dynamics $(F_n)_{n=1}^N$ initialized at $Z$, and $Z_{-k}$ is the $k^\text{th}$
state generated by the backward dynamics $(B_n)_{n=1}^N$ initialized at $Z$. 
Hat accents denote the same for the
pseudo-dynamics: for example, $\hZ_k$ is the $k^\text{th}$ state generated by $(\hF_n)_{n=1}^N$ when initialized at $Z$.

\begin{figure}
\centering
\includegraphics[width = 0.7\textwidth]{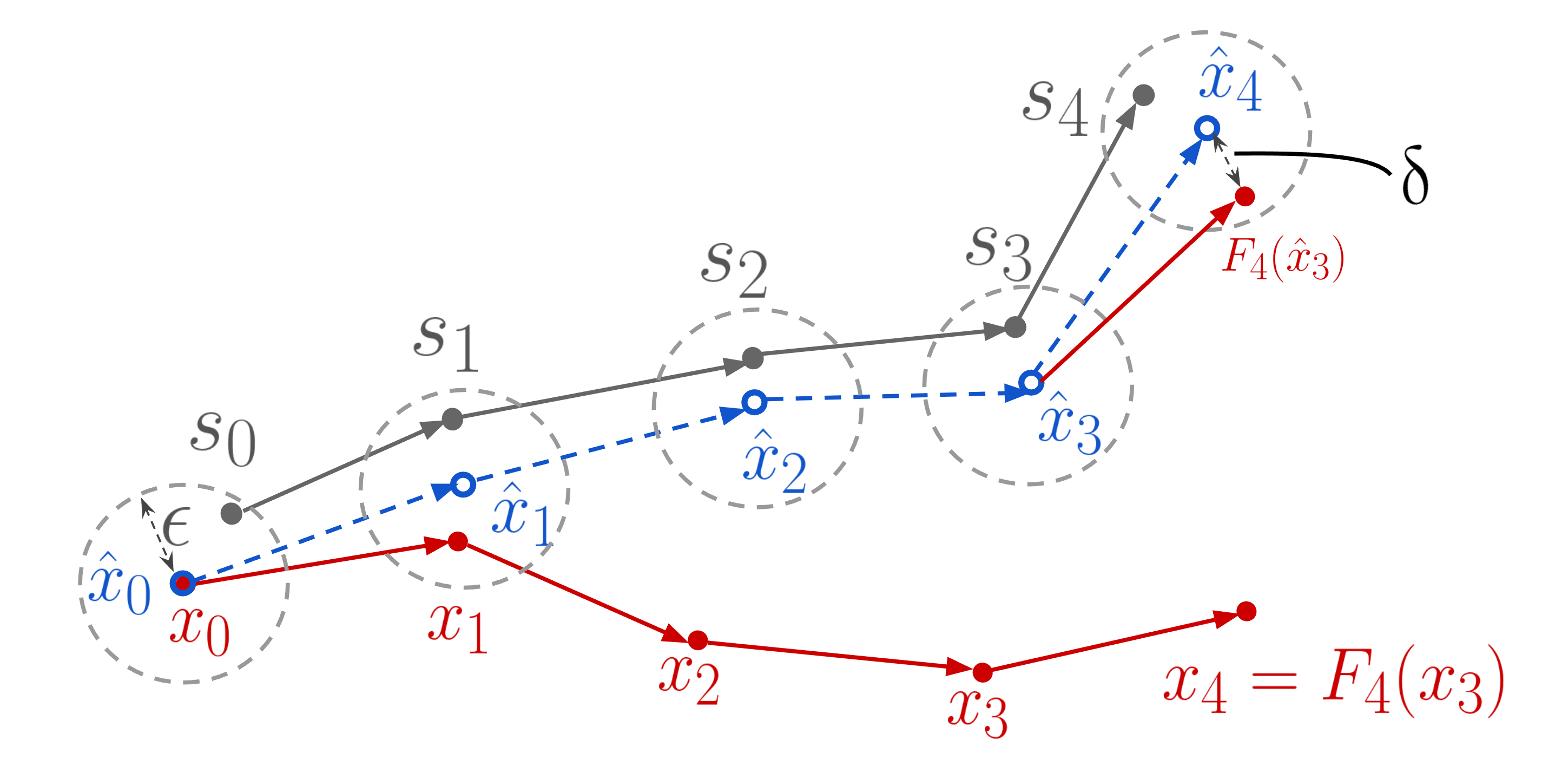}
\caption{
A visualization of a pseudo-orbit and shadowing orbit.
Solid arrows and filled dots indicate exact orbits, while dashed arrows and
open dots indicate pseudo-orbits (e.g., via numerical computations). 
Red indicates the
{\color[rgb]{0.8, 0, 0} exact orbit $(x_0, \dots, x_4)$} that one intends to
compute.  
Blue indicates the {\color[rgb]{0.067, 0.333, 0.80} numerically computed $\delta$-pseudo-orbit $(\hx_0, \dots, \hx_4)$.}
 Grey indicates the {\color[rgb]{0.333, 0.333, 0.334}corresponding $\eps$-shadowing orbit $(s_1, \dots, s_4)$}.
At the top right of the figure, $\|{\color[rgb]{0.067, 0.333,
0.80} \hx_4} - {\color[rgb]{0.8, 0, 0} F_4(\hx_3)} \| \leq \delta$ illustrates the $\delta$ numerical error at each step of the pseudo-orbit. 
The grey dashed circles demonstrate the {\color[rgb]{0.333, 0.333, 0.334} $\epsilon$-shadowing window}.  
}
\label{fig:shadowing}
\end{figure}
The forward/backward orbits satisfy $x_{k+1} = F_k(x_k)$ and $x_{-(k+1)} = B_k(x_{-k})$ exactly at each step.
On the other hand, the forward/backward pseudo-orbits incur a small amount of error,
\[
\|\hx_{k+1} - F_k(\hx_k)\| \leq \delta \quad \text{and}\quad \|\hx_{-(k+1)} - B_k(\hx_{-k})\| \leq \delta,
\]
which can be magnified quickly along the orbit.
However, there often exists another \emph{exact} orbit starting at some other point $s\in\mcX$ that remains in a close
neighbourhood of the numerical orbit.
This property, illustrated in \cref{fig:shadowing}, is referred to 
as the \emph{shadowing property} \citep{pilyugin2006shadowing,pilyugin2011theory,fakhari2010shadowing,corless1995approximate,meddaugh2021shadowing}.

\bdef[{Shadowing property \citep{pilyugin2011theory}}] \label{def:shadowing}
The forward dynamics $(F_n)_{n=1}^N$ has the $\bm{(\eps,\delta)}$\textbf{-shadowing property}
if for all $x\in\mcX$ and all $(\hF_n)_{n=1}^N$ satisfying \cref{eq:onestepdelta},
there exists an $s\in\mcX$ such that
\[
    \forall k=0, 1, \dots, N, \quad \| s_{k}- \hx_k\| < \eps.
\]  
\edef
An analogous definition holds for the backward dynamics $(B_n)_{n=1}^N$---where there is a shadowing
orbit $s_{-k}$ that is nearby the pseudo-orbit $\hx_{-k}$---and for the joint
forward and backward dynamics, where there is a shadowing orbit nearby both the backward and forward pseudo-orbits
simultaneously.
The key idea in this paper is that, intuitively, if the numerically computed pseudo-orbit
is close to \emph{some} exact orbit (the shadowing orbit),   
statistical computations based on the 
pseudo-orbit---e.g., sampling, density evaluation, and ELBO estimation---should be close to 
those obtained via that exact orbit.
We will defer the examination of when (and to what extent) shadowing holds
to \cref{sec:shadowing}; in this section, we will use it as an assumption
when analyzing statistical computations 
with numerically implemented variational flows.

\subsection{Error analysis of normalizing flows via shadowing} \label{sec:normalizingflow}
\paragraph{Sampling.} 
Our first goal is to relate the marginal distributions of $X_N$ and $\hX_N$
for $X\distas q_0$, i.e.,
to quantify the error in sampling due to numerical approximation.
Assume the normalizing flow has the $(\eps,\delta)$-shadowing property,
and let $\xi_0$ be the marginal distribution of the shadowing orbit start point.
We suspect that $\xi_0 \approx q_0$, in some sense; for example, we know that the 
bounded Lipschitz distance $\bld{\xi_0}{q_0}$ is at most $\epsilon$ due to shadowing.
And $\xi_0$ is indeed an implicit function of $q_0$; it is a fixed point of a
twice differentiable function involving the whole orbit starting at $q_0$
\citep[Page. ~176]{coomes1996shadowing}.
But the distribution $\xi_0$
is generally hard to describe more completely, and thus it is common
to impose additional assumptions.
For example, past work shows that  
the L\'evy-Prokhorov metric $\lpd{\hX_N}{X_N}$ is bounded by $\eps$,
under the assumption that $\xi_0 = q_0$ \citep{tupper2009relation}.
We provide a more general result (\cref{prop:nfsample}) without distributional
assumptions on $\xi_0$. We control $\bld{\cdot}{\cdot}$ rather than $\lpd{\cdot}{\cdot}$,
as its analysis is simpler and both metrize weak distance.

\bprop \label{prop:nfsample} 
Suppose the forward dynamics has the $(\eps,\delta)$-shadowing property,
and $X\!\!\distas\! q_0$. Then
\[
\sup_{f : |f|\leq U, \mathrm{Lip}(f) \leq \ell} \left| \EE f(X_N) - \EE f(\hX_N)\right| \leq \ell \eps + 2U \tvd{\xi_0}{q_0}.
\]
In particular, with $\ell = U = 1$, we obtain that $\bld{X_N}{\hX_N} \leq \eps + 2\tvd{\xi_0}{q_0}$.
\eprop
Recall the layerwise error bound from \cref{eq:layerwisebound}---which suggests
that the difference in orbit and pseudo-orbit grows exponentially in $N$---and
compare to \cref{prop:nfsample}, which asserts that the error is controlled by
the shadowing window size $\eps$. This window size may depend on $N$, but
we find in \cref{sec:expts} it is usually not much larger than $\delta$ in practice, which itself is 
typically near the precision of the relevant numerical representation.
We will show how to estimate $\eps$ later in \cref{sec:shadowing}.

\paragraph{Density evaluation.}
The exact density $q(x)$ follows \cref{eq:transformationvars},
while the approximation $\hq(x)$ is the same except that we use the backward
\emph{pseudo}-dynamics.
For $x\in\mcX$, a differentiable function $g: \mcX \mapsto \reals^+$, define the
\emph{local Lipschitz constant} for the logarithm of $g$ around $x$ as: 
\[\label{eq:locallipconstant}
L_{g, \eps}(x) = \sup_{\|y-x\|\leq \epsilon} \|\grad \log g(y)\|.
\]
\cref{thm:nflpdf} shows that, given the shadowing property,
the numerical error is controlled by the shadowing window size $\eps$ and the sum of the Lipschitz
constants along the pseudo-orbit. The constant $L_{q,\epsilon}$ occurs because we are
essentially evaluating $q(s)$ rather than $q(x)$, where $s\in\mcX$ is the backward shadowing orbit initialization.
The remaining constants occur because of the approximation of the shadowing orbit with 
the nearby, numerically-computed pseudo-orbit in the density formula.

\bthm \label{thm:nflpdf}
Suppose the backward dynamics has the $(\eps,\delta)$-shadowing property.
Then
\[
   \forall x \in \mcX, \quad |\log \hq(x) - \log q(x)| \leq  \epsilon \cdot
   \left(L_{q,\eps}(x) + L_{q_0,\eps}(\hx_{-N})+\sum_{n=1}^{N} L_{J_{N-n+1},\eps}(\hx_{-n})\right). 
\]
\ethm

\paragraph{ELBO estimation.}
 The exact ELBO estimation function is given in \cref{eq:nfelbo};
the numerical ELBO estimation function $\hE(x)$ is the same except that we use the forward
\emph{pseudo}-dynamics. 
The quantity $E(X)$, $X\distas q_0$ is an unbiased estimate of the exact ELBO, while $\hE(X)$ is 
biased by numerical error; \cref{thm:nfelbo} quantifies this error. 
Note that for simplicity we assume that the initial state distributions $q_0$, $\xi_0$ 
described earlier are identical. It is possible to obtain a bound including a $\tvd{q_0}{\xi_0}$ term, but this
would require the assumption that $\log p(x)/q(x)$ is uniformly bounded on $\mcX$.
\bthm \label{thm:nfelbo}
Suppose the forward dynamics has the $(\eps,\delta)$-shadowing property,
and $\xi_0 = q_0$.
Then
\[
\abs{\ELBO{q}{p} - \EE[\hE(X)]} \leq \eps\cdot \EE\left[L_{p,\eps}(\hX_N) + L_{q_0,\eps}(X) + \sum_{n=1}^N L_{J_n,\eps}(\hX_n)\right] \qquad \text{for } X\distas q_0.
\]
\ethm

\subsection{Error analysis of MixFlows via shadowing} \label{sec:mixflow}

The error analysis for MixFlows for finite length $N\in\nats$ parallels that of normalizing flows, except
that $F_n = F$, $B_n = B$, and $J_n = J$ for $n = 1, \dots, N$. 
However, when $F$ and $B$ are ergodic and measure-preserving for the target $\pi$, 
we provide asymptotically simpler bounds in the large $N$ limit that
do not depend on the difficult-to-analyze details of the Lipschitz constants along a pseudo-orbit.
These results show that the error of sampling tends to be constant in flow length,
while the error of density evaluation and ELBO estimation grows at most linearly.
We say the forward dynamics has the \emph{infinite $(\eps,\delta)$-shadowing property}
if $(F_n)_{n=1}^\infty$  has the $(\eps,\delta)$-shadowing property \cite{pilyugin1999shadowing}. 
Analogous definitions hold for both the backward and joint forward/backward dynamics.

\paragraph{Sampling.} Similar to \cref{prop:nfsample} for normalizing flows,
\cref{prop:mixsample} bounds the error in exact draws $Y$ and approximate draws $\hY$
from the MixFlow. The result demonstrates that error does not directly depend on the flow length $N$, 
but rather on the shadowing window $\epsilon$.
In addition, in the setting where the flow map $F$ is $\pi$-ergodic and measure-preserving,
we can (asymptotically) remove the total variation term between the initial shadowing distribution $\xi_0$
and the reference distribution $q_0$. Note that in the asymptotic bound,
the distributions of $Y$ and $\hY$ are functions of $N$.
\bprop \label{prop:mixsample}
Suppose the forward dynamics has the $(\epsilon,\delta)$-shadowing property, and $X\distas q_0$. 
Let $Y = X_K$, $\hY = \hX_K$ for $K\distas \distUnif\{0,1,\dots,N\}$. Then
\[
    \sup_{f:|f|\leq U, \mathrm{Lip}(f)\leq \ell} \left|\EE f(Y) - \EE f(\hY)\right| &\leq \ell \epsilon + 2U \tvd{\xi_0}{q_0}.
\]
In particular, if $\ell=U=1$, we obtain that $\bld{Y}{\hY} \leq \epsilon + 2\tvd{\xi_0}{q_0}$. 
If additionally the forward dynamics has the infinite $(\epsilon,\delta)$-shadowing property, 
$F$ is $\pi$-ergodic and measure-preserving, and $\xi_0, q_0 \ll \pi$, then
\[
    \limsup_{N\to\infty} \bld{Y}{\hY} \leq \eps.
\]
\eprop
A direct corollary of the second result in \cref{prop:mixsample} is that for $W \distas \pi$, $\lim_{N\to\infty} \bld{W}{\hY} \leq \epsilon$,
which results from the fact that $\bld{W}{Y}\to 0$ \cite[Theorem 4.1]{Xu22mixflow}. In other words, the bounded Lipschitz distance
of the approximated MixFlow and the target $\pi$ is asymptotically controlled by the shadowing window $\eps$. This improves
upon earlier theory governing approximate MixFlows, for which the best guarantee available
had total variation error growing as $O(N)$  \cite[Theorem 4.3]{Xu22mixflow}.

\paragraph{Density evaluation.}
The exact density follows  \cref{eq:mixflowdensity};
the numerical approximation is the same except that we use the backward \emph{pseudo}-dynamics. 
We obtain a similar finite-$N$ result as in \cref{thm:nflpdf}. Further, we show
that given infinite shadowing---where the shadowing window size
$\eps$ is independent of flow length $N$---the numerical approximation error 
asymptotically grows at most linearly in $N$, in proportion to $\epsilon$.
\bthm \label{thm:mflpdf}
Suppose the $(\eps,\delta)$-shadowing property holds for the backward dynamics.
Then
\[
\forall x\in\mcX, \quad |\log \hq(x) - \log q(x)| &\leq 
\epsilon \cdot \left(L_{q,\eps}(x) + \max_{0\leq n \leq N}
L_{q_0,\eps}(\hx_{-n})+\sum_{n=1}^{N} L_{J,\eps}(\hx_{-n})\right).
\]
If additionally the backward dynamics has the infinite $(\epsilon,\delta)$-shadowing property,
$F$ is $\pi$-ergodic and measure-preserving, $L_{q,\eps}(x) = o(N)$ as $N\to\infty$,
and $\xi_0 \ll \pi$, then for $q_0$-almost every $x\in\mcX$,
\[
\limsup_{N\to\infty} \frac{1}{N}|\log \hq(x) - \log q(x)| &\leq  \epsilon \cdot
\EE\left[L_{q_0,\eps}(X) + L_{J,\eps}(X)\right], \quad X\distas \pi.
\]
\ethm

\paragraph{ELBO estimation.} 
The exact ELBO formula for the MixFlow is given in \cref{eq:averagedelbo}.
Note that here we do not simply substitute the forward/backward pseudo-orbits as needed; the na\"ive approximation of 
the terms $q(F^n x)$ would involve $n$ applications of $\hF$ followed by $N$ applications of $\hB$,
which do not exactly invert one another. Instead, we analyze the method proposed in \cite{Xu22mixflow},
which involves simulating a single forward pseudo-orbit $\hx_1, \dots, \hx_N$ and backward pseudo-orbit
$\hx_{-1}, \dots, \hx_{-N}$ starting from $x\in \mcX$, and then caching these and using them as needed 
in the exact formula.

\bthm\label{thm:mfelbo}
Suppose the joint forward and backward dynamics has the $(\eps,\delta)$-shadowing property,
and $\xi_0 = q_0$. Then for $X\distas q_0$, 
\[
\left|\ELBO{q}{p}-\EE\left[\hE(X)\right]\right| \leq 
\eps\cdot\EE\left[\frac{1}{N+1}\sum_{n=0}^N L_{p,\eps}(\hX_n) +\!\!\!\!
\max_{-N\leq n \leq N}\!\! L_{q_0,\eps}(\hX_n) + \!\!\!\sum_{n=-N}^NL_{J,\eps}(\hX_{n})\right].
\]
If additionally the joint forward and backward dynamics has the infinite $(\eps,\delta)$-shadowing property,
$F$ is $\pi$-ergodic and measure-preserving, and for some $1 \leq m_1 < \infty$ and $1/m_1+1/m_2 = 1$
$L_{p,\eps}, L_{q_0,\eps}, L_{J,\eps} \in L^{m_1}(\pi)$ 
and $\der{q_0}{\pi} \in L^{m_2}(\pi)$, then
\[
\limsup_{N\to\infty}\frac{1}{N}\left|\ELBO{q}{p}-\EE\left[\hE(X)\right]\right|
&\leq 
2\eps\cdot\EE\left[L_{q_0,\eps}(X) + L_{J,\eps}(X)\right], \quad X\distas \pi.
\]
\ethm

\section{Computation of the shadowing window size} \label{sec:shadowing}
We have so far assumed the $(\eps,\delta)$-shadowing property throughout.
To make use of our results in practice, it is crucial to understand the shadowing window size $\eps$.
\cref{thm:finiteshadowing} presents sufficient conditions for a finite
dynamical system to have the shadowing property, and
characterizes the size of the shadowing window $\eps$. 
Note that throughout this section we focus on the forward dynamics; the backward and joint dynamics
can be treated identically.
Let $\|\cdot\|$ denote the spectral norm of a matrix.
\bthm[Finite shadowing theorem]
\label{thm:finiteshadowing}
Suppose the dynamics $(F_n)_{n=1}^N$ are $C^2$ diffeomorphisms on $\mcX$,
and the pseudo-dynamics $(\hF_n)_{n=1}^N$ satisfy \cref{eq:onestepdelta}.
For a given $x\in\mcX$, define the operator $A:\mcX^{N+1} \mapsto \mcX^{N}$
by
\[
    (A u)_k = u_{k+1} - \nabla F_{k+1}(\hx_k) u_k, \quad \text{for } \,u = (u_0, u_1, \dots, u_N) \in \mcX^{N}, \quad k = 0, 1, \dots, N-1.
\]
Let 
\[ \label{eq:existence}
   M \defas \max\left\{ \sup\nolimits_{\|v\|\leq 2\lambda\delta} \|\nabla^2 F_{n+1}(\hx_n + v)\|: n = 0, 1, \dots, N-1 \right\} \text{ where } \lambda = \lambda_{\text{min}}(AA^T)^{-1/2}.
\]
If $2M\lambda^2\delta \leq 1$, then the pseudo-orbit starting at $x$ is shadowed with $\epsilon = 2\lambda\delta$.
\ethm
\bprf
This result follows
directly by following the proof of \citep[Theorem 11.3]{palmer2000shadowing}
with nonconstant flow maps $F_n$, and then using 
the right-inverse norm from \citep[Corollary 4.2]{dokmanic2017beyond}.
\eprf

In order to apply \cref{thm:finiteshadowing} we need to (1) estimate $\delta$, e.g., by examining
numerical error of one step of the map in practice; (2)  compute $\lambda_{\min}(AA^T)$;
and (3) estimate $M$, e.g., by bounding the third derivative of $F_n$. While estimating $M$ and $\delta$ are problem-specific, 
one can employ standard procedures for computing $\lambda_{\min}(AA^T)$.
The matrix $AA^T$ has a block-tridiagonal form,
\[\label{eq:matrixLLT}
    AA^T = \bbmat 
    D_1 D_1^T + I& -D_2^T &  & &  \\ 
	-D_2 & D_2 D_2^T + I &  -D_3^T&  &\\
	&  \ddots & \ddots &\ddots &\\ 
	&  & -D_{N-1}& D_{N-1}D_{N-1}^T+I & -D_N^T \\ 
	&  & & -D_N &D_ND_N^T +I \\ 
	\ebmat,
\]
where $D_k = \nabla F_k(\hx_{k-1}) \in \reals^{d\times d}, \forall k \in [N]$.  
Notice that $AA^T$ is a symmetric positive definite sparse matrix with bandwidth
$d$ and so has $O(Nd^2)$ entries. The inherent structured sparsity can be
leveraged to design efficient eigenvalue computation methods, e.g., the inverse
power method \citep{schatzman2002numerical}, or tridiagonalization via Lanczos
iterations \citep{lanczos1950iteration} followed by divide-and-conquer
algorithms \citep{coakley2013379}.
However, in our experiments, directly calling the \texttt{eigmin} function provided
in Julia suffices; as illustrated in
\cref{fig:hamflowshadowing,fig:time}, the shadowing window computation requires
only a few seconds for low dimensional synthetic examples with hundreds flow layers,
and several minutes for real data examples. Hence, we didn't pursue specialized
methods, leaving that for future work.
It is also noteworthy that practical computations of $D_k$ can introduce
floating-point errors, influencing the accuracy of $\lambda$. 
To address this, one might consider adjusting the shadowing window size.
We explain how to manage this in \cref{sec:windowcomputeerror}, and why such numerical discrepancies
minimally impact results.

Finally, since $\eps$ depends on $\lambda_{\text{min}}(AA^T)^{-1/2}$, which itself
depends on $N$, the shadowing window size may potentially increase with $N$ in practice.
Understanding this dependence accurately for a general dynamical system
is challenging; there are examples where it remains constant, for instance (see
\cref{exa:scalingmap,exa:hyperbolic} in \cref{apdx:windowsizescaling}),
but $\eps$ may grow with $N$.
Our empirical results in next section show that on representative inferential examples
with over $500$ flow layers, $\eps$
scales roughly linearly with $N$ and is of a similar order of magnitude 
as the floating point representation error.
For further discussion on this topic, we refer readers to \cref{apdx:windowsizescaling}.

\section{Experiments}\label{sec:expts}

\captionsetup[subfigure]{labelformat=empty}
\begin{figure*}[t!]
\centering 
\begin{subfigure}[b]{0.24\columnwidth} 
    \scalebox{1}{\includegraphics[width=\columnwidth]{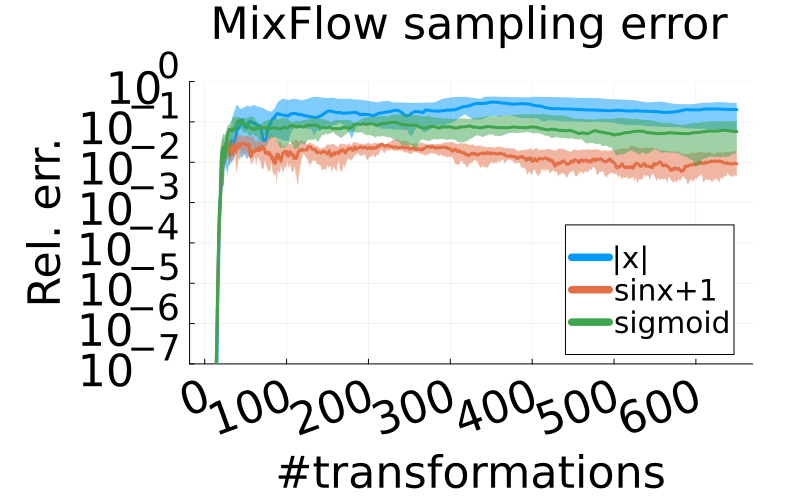}}
    \caption{(a) Banana}
\end{subfigure}
\hfill
\centering 
\begin{subfigure}[b]{0.24\columnwidth} 
    \scalebox{1}{\includegraphics[width=\columnwidth]{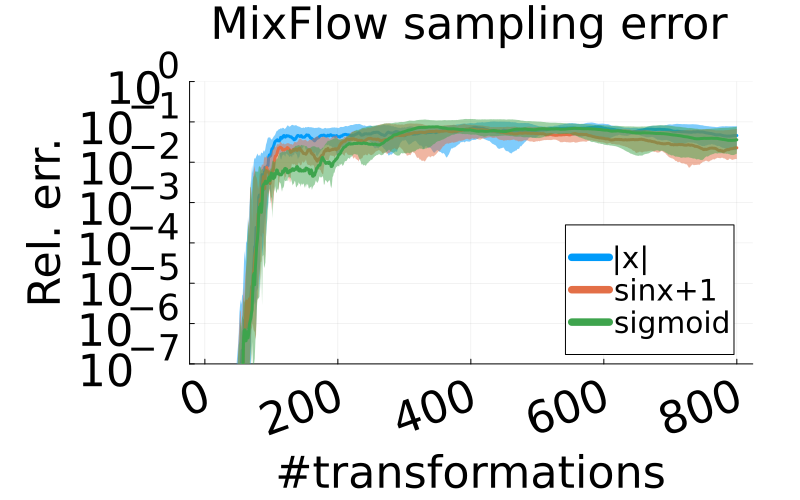}}
    \caption{(b) Cross}
\end{subfigure}
\hfill
\centering 
\begin{subfigure}[b]{0.24\columnwidth} 
    \scalebox{1}{\includegraphics[width=\columnwidth]{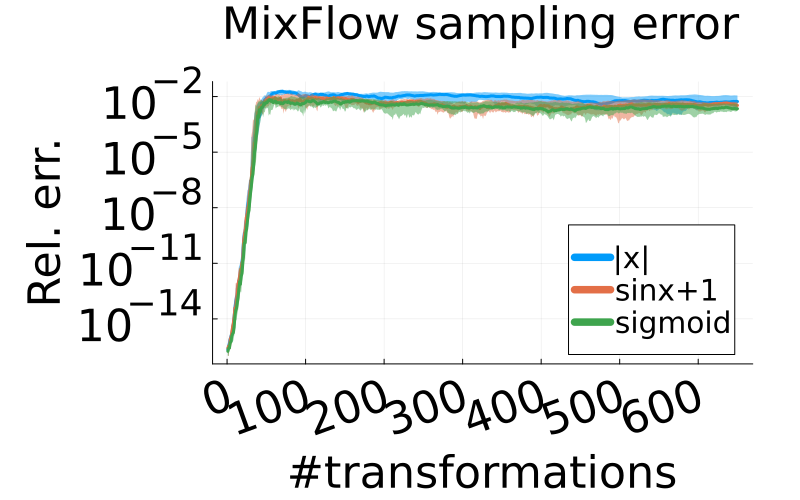}}
    \caption{(c) Lin. Reg.}
\end{subfigure}
\hfill
\centering 
\begin{subfigure}[b]{0.24\columnwidth} 
    \scalebox{1}{\includegraphics[width=\columnwidth]{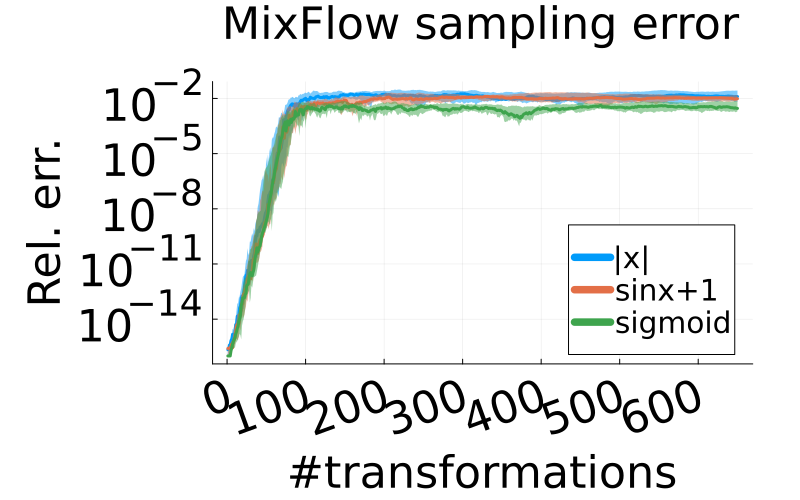}}
    \caption{(d) Log. Reg.}
\end{subfigure}
\hfill
\caption{MixFlow relative sample average computation error on three test functions:
    $\sum_{i = 1}^d[|x|]_i, \sum_{i = 1}^d[\sin(x) + 1]_i$ and $\sum_{i = 1}^d[\mathrm{sigmoid}(x)]_i$. 
The lines indicate the median, and error regions indicate $25^\text{th}$ to $75^\text{th}$ percentile from $20$ runs.}
\label{fig:mfsamplingerr}
\end{figure*}

\captionsetup[subfigure]{labelformat=empty}
\begin{figure*}[t!]
\centering 
\begin{subfigure}[b]{0.24\columnwidth} 
    \scalebox{1}{\includegraphics[width=\columnwidth]{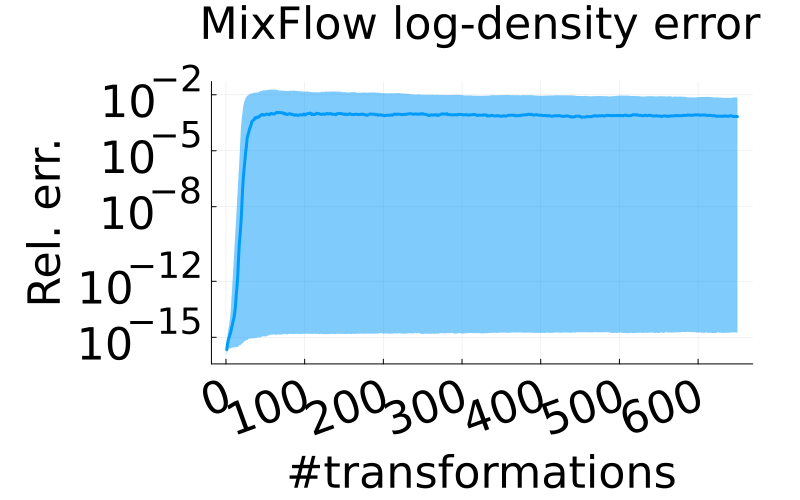}}
    \caption{(a) Banana}
\end{subfigure}
\hfill
\centering 
\begin{subfigure}[b]{0.24\columnwidth} 
    \scalebox{1}{\includegraphics[width=\columnwidth]{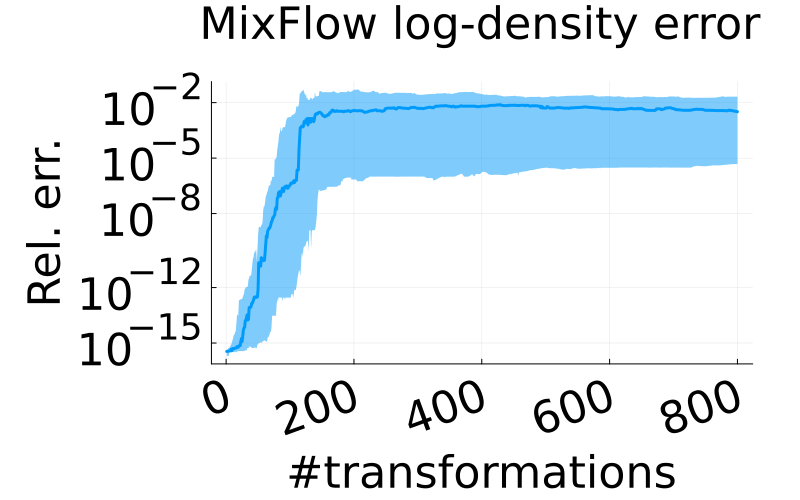}}
    \caption{(b) Cross}
\end{subfigure}
\hfill
\centering 
\begin{subfigure}[b]{0.24\columnwidth} 
    \scalebox{1}{\includegraphics[width=\columnwidth]{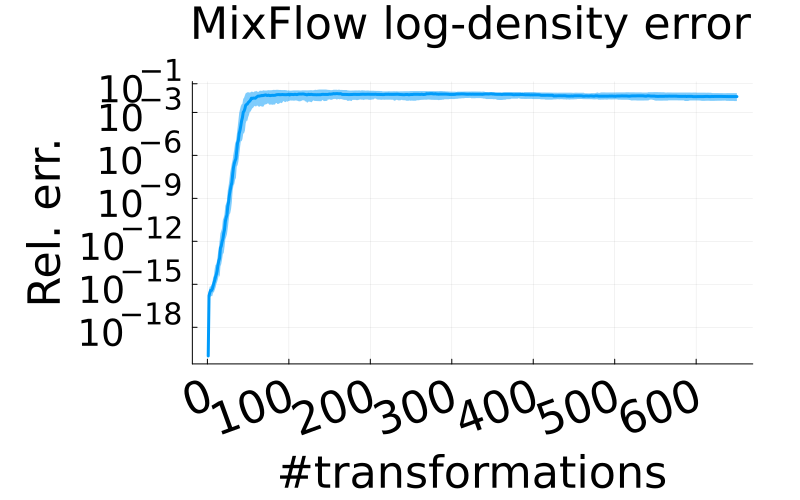}}
    \caption{(c) Lin. Reg.}
\end{subfigure}
\hfill
\centering 
\begin{subfigure}[b]{0.24\columnwidth} 
    \scalebox{1}{\includegraphics[width=\columnwidth]{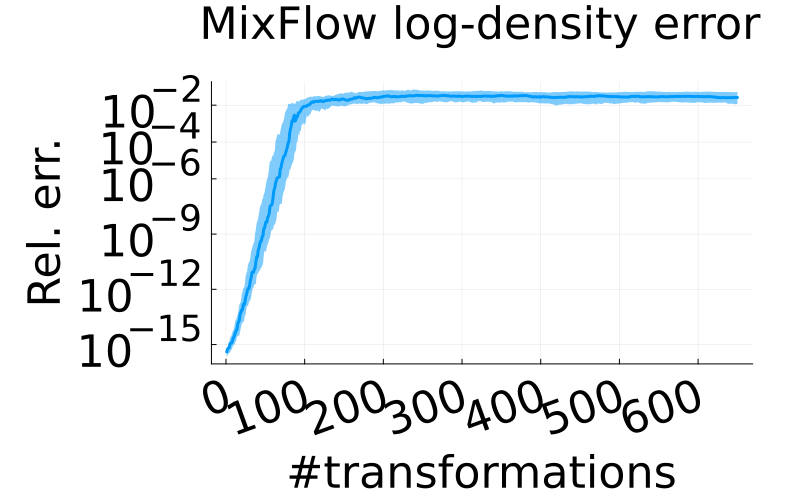}}
    \caption{(d) Log. Reg.}
\end{subfigure}
\hfill
\caption{MixFlow relative log-density evaluation error. 
The lines indicate the median, and error regions indicate $25^\text{th}$ to
$75^\text{th}$ percentile from $100$ evaluations on different positions.}
\label{fig:mflpdfsrelerr}
\end{figure*}

\captionsetup[subfigure]{labelformat=empty}
\begin{figure*}[t!]
\centering 
\begin{subfigure}[b]{0.24\columnwidth} 
    \scalebox{1}{\includegraphics[width=\columnwidth]{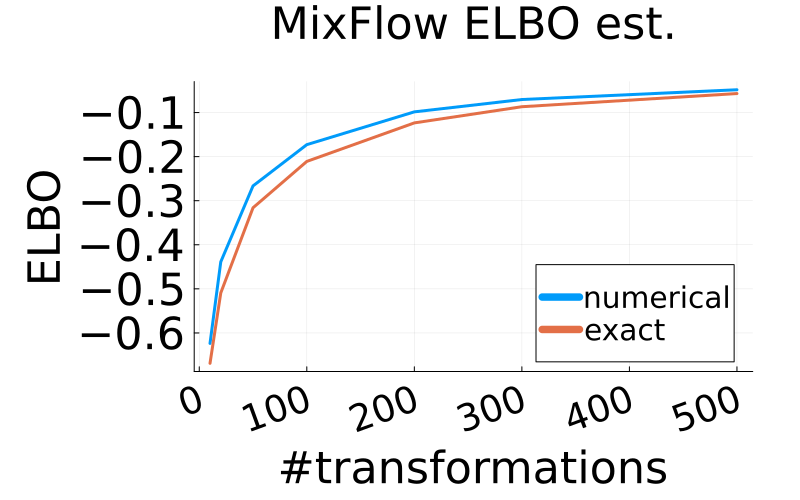}}
    \caption{(a) Banana}
\end{subfigure}
\hfill
\centering 
\begin{subfigure}[b]{0.24\columnwidth} 
    \scalebox{1}{\includegraphics[width=\columnwidth]{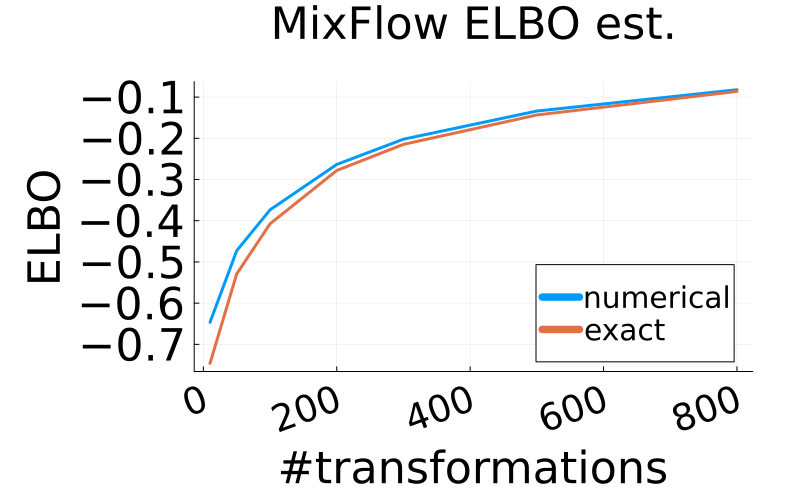}}
    \caption{(b) Cross}
\end{subfigure}
\hfill
\centering 
\begin{subfigure}[b]{0.24\columnwidth} 
    \scalebox{1}{\includegraphics[width=\columnwidth]{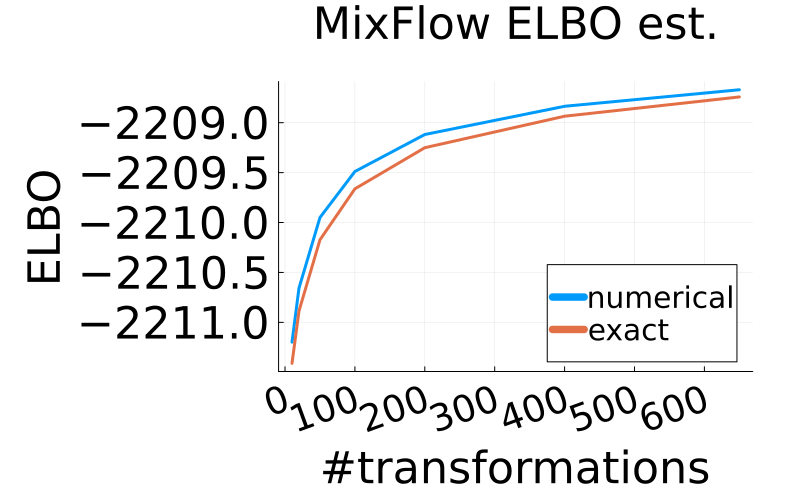}}
    \caption{(c) Lin. Reg.}
\end{subfigure}
\hfill
\centering 
\begin{subfigure}[b]{0.24\columnwidth} 
    \scalebox{1}{\includegraphics[width=\columnwidth]{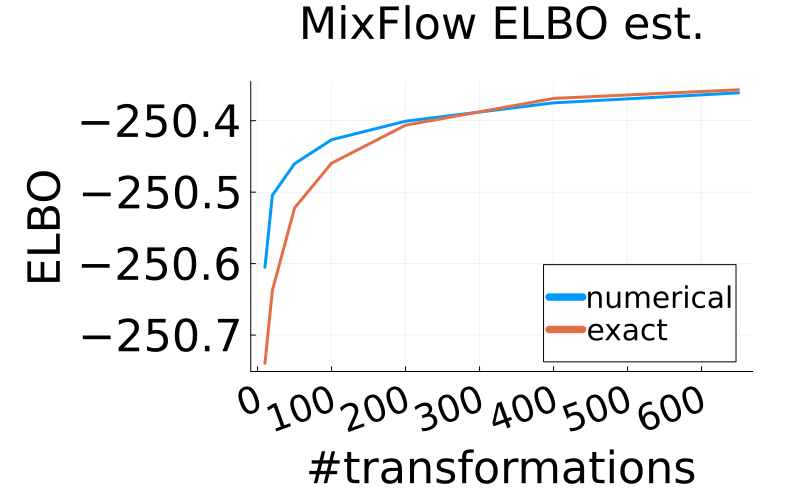}}
    \caption{(d) Log. Reg.}
\end{subfigure}
\hfill
\caption{MixFlow \texttt{exact} and \texttt{numerical} ELBO estimation over increasing flow length.
The lines indicate the averaged ELBO estimates over $200$ independent
forward orbits. The Monte Carlo error for the given estimates is sufficiently small
so we omit the error bar. }
\label{fig:mfelbos}
\end{figure*}

\captionsetup[subfigure]{labelformat=empty}
\begin{figure*}[t!]
\centering 
\begin{subfigure}[b]{0.24\columnwidth} 
    \scalebox{1}{\includegraphics[width=\columnwidth]{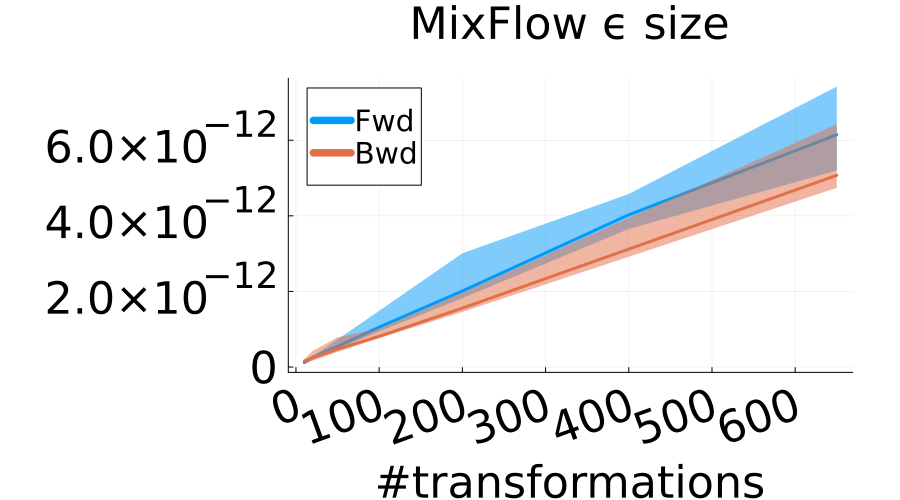}}
\caption{(a) Banana}
\end{subfigure}
\hfill
\centering 
\begin{subfigure}[b]{0.24\columnwidth} 
    \scalebox{1}{\includegraphics[width=\columnwidth]{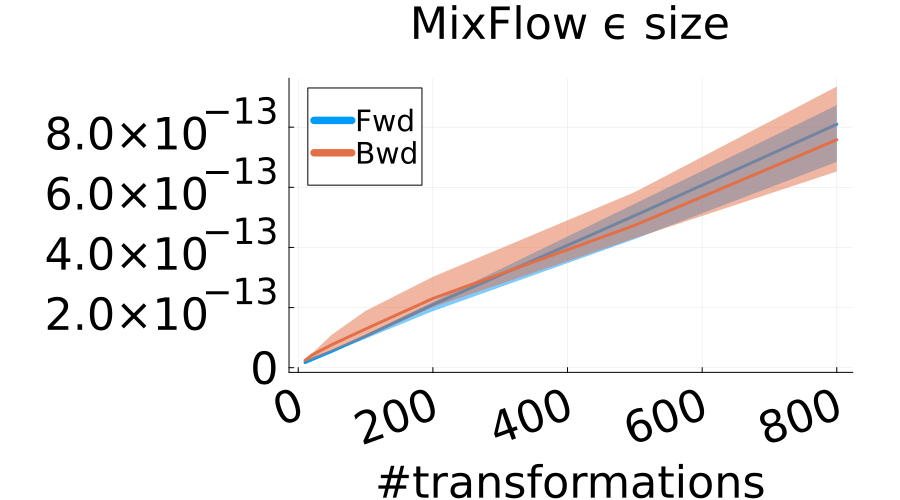}}
    \caption{(b) Cross}
\end{subfigure}
\hfill
\centering 
\begin{subfigure}[b]{0.24\columnwidth} 
    \scalebox{1}{\includegraphics[width=\columnwidth]{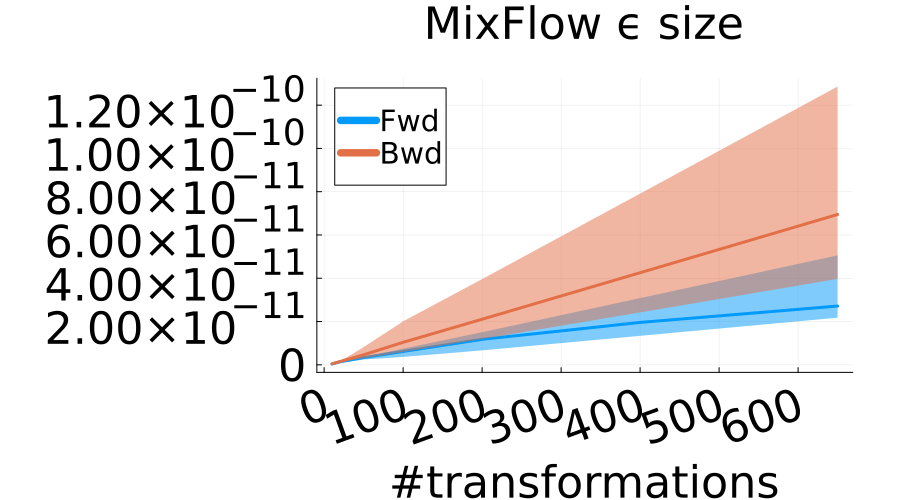}}
\caption{(c) Linear regression}
\end{subfigure}
\hfill
\centering 
\begin{subfigure}[b]{0.24\columnwidth} 
    \scalebox{1}{\includegraphics[width=\columnwidth]{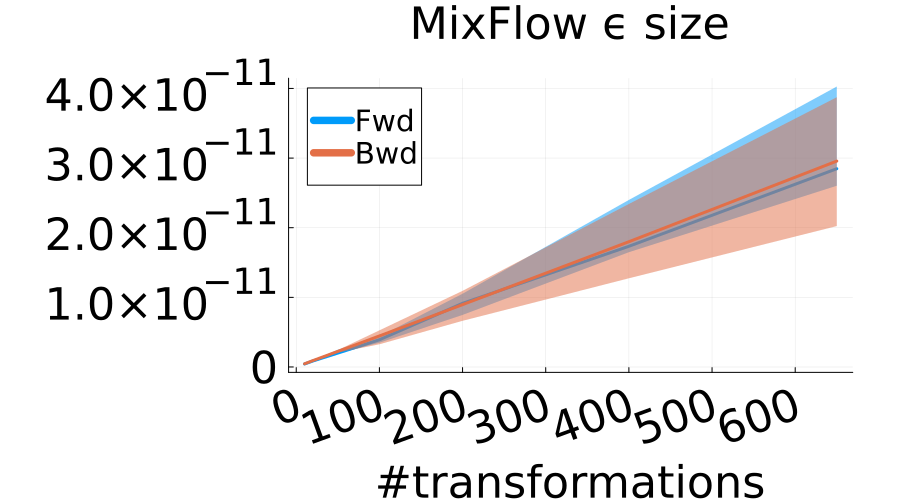}}
\caption{(d) Logistic regression }
\end{subfigure}
\hfill
\caption{MixFlow shadowing window size $\eps$ over increasing flow length. 
    The lines indicate the median, and error regions indicate $25^\text{th}$ to
$75^\text{th}$ percentile from $10$ runs. }
\label{fig:shadowingwindow}
\end{figure*}

In this section, we verify our error bounds and diagnostic procedure of MixFlow 
on the banana, cross, and 2 real data targets---a Bayesian linear regression and logistic
regression posterior; detailed model and dataset descriptions can be found in \cref{apdx:model}.  
We also provide a similar empirical investigation for Hamiltonian flow
\citep{Chen22} on the same synthetic targets in \cref{apdx:nfresults}.

We begin by assessing the error of trajectory-averaged estimates based on approximate
draws.  \cref{fig:mfsamplingerr} displays the relative
numerical estimation error compared to the exact estimation based on the same
initial draw. Although substantial numerical error was observed in orbit
computation (see \cref{fig:trajerrallex} in \cref{apdx:moreresults}), the relative
sample estimate error was around $1\%$ for all four examples,  
suggesting that the statistical properties of the forward orbits closely resemble those of the exact orbit.

We then focus on the density evaluation. 
For synthetic examples, we assessed densities at 100 evenly distributed points
within the target region (\cref{fig:banana_scatter,fig:cross_scatter}). For real
data, we evaluated densities at the locations of 100 samples from MCMC method
None-U-turn sampler (\texttt{NUTS}); detailed settings for \texttt{NUTS} is described in \cref{apdx:mixflowparameters}.
It is evident from the relative error shown in \cref{fig:mflpdfsrelerr} that the numerical density closely
matches the exact density evaluation,
with the relative numerical error ranging between $0.1\%$ and $1\%$, which is
quite small.
The absolute error can be found in \cref{fig:mflpdfserr} in
\cref{apdx:moreresults}, showing that the density evaluation error does not grow as substantially as
the orbit computation error (\cref{fig:trajerrallex} in \cref{apdx:moreresults}),  
which aligns with the bound in \cref{thm:mflpdf}.

\cref{fig:mfelbos} further demonstrates the numerical error for the ELBO estimations
(\cref{eq:averagedelbo}). In each example, both the averaged exact ELBO
estimates and numerical ELBO estimates are plotted against an increasing flow
length. Each ELBO curve is averaged over 200 independent forward orbits. Across
all four examples, the numerical ELBO curve remain closely aligned with the
exact ELBO curve, indicating that the numerical error is small in comparison to the scale of the ELBO
value. Moreover, the error does not grow with an increasing flow length, 
and even presents a decreasing trend as $N$ increases in the two synthetic examples
and the Logistic regression example. 
This aligns well with the error bounds provided in \cref{thm:mfelbo}.

Finally, \cref{fig:shadowingwindow} presents the size of the shadowing window
$\eps$ as the flow length $N$ increases.  As noted earlier, $\eps$
depends on the flow map approximation error $\delta$ and potentially $N$. We
evaluated the size of $\delta$ by calculating the approximation error of a
single $F$ or $B$ for 1000 \iid draws from the reference distribution. Boxplots
of $\delta$ for all examples can be found in \cref{fig:delta} of
\cref{apdx:moreresults}. These results show that in the four examples, $\delta$
is close to the floating point representation error (approximately $10^{-14}$).
Thus, we fixed $\delta$ at $10^{-14}$ when computing $\eps$.
As shown in \cref{fig:shadowingwindow},
$\eps$ for both the forward and backward orbits grows roughly linearly with the
flow length. This growth is significantly less drastic than the exponential
growth of the orbit computation error. 
And crucially, the shadowing window size is very small---smaller than $10^{-10}$ for MixFlow of length over
$1000$, which justifies the validity of the downstream statistical procedures.
In contrast, the orbit computation errors in the two synthetic examples rapidly
reach a magnitude similar to the scale of the target distributions, as shown in \cref{fig:trajerr}.

\section{Conclusion}

This work delves into the numerical stability of variational flows, drawing
insights from shadowing theory and introducing a diagnostic tool to assess the
shadowing property. 
Experiments corroborate our theory and demonstrate the effectiveness of the
diagnostic procedure. 
However, the scope of our current analysis is limited to downstream
tasks post-training of discrete-time flows. Understanding the impact of
numerical instability during the training phase or probing into recent
architectures like continuous normalizing flows or neural ODEs
\citep{zhuang2021mali,zhuang2020adaptive,chen2018neural,kidger2022neural,chang2018reversible},
remains an avenue for further exploration.
Additionally, 
while our theory is centered around error bounds that are proportional to the
shadowing window size $\eps$, we have recognized that $\eps$ can grow with $N$. 
Further study on its theoretical growth rate is needed.
Finally, in our experiments, we employed a basic method to calculate the minimum
eigenvalue of $AA^T$. 
Given the sparsity of this matrix, a deeper exploration into a more efficient computational procedure is merited.

\small
\bibliographystyle{unsrtnat} %imsart-nameyear
\bibliography{sources}

%%%%%%%%%%%%%%%%%%%%%%%%%%%%%%%%%%%%%%%%%%%%%%%%%%%%%%%%%%%%
\newpage
\appendix
\section{Proofs}

\subsection{Layer-wise error analysis}\label{sec:layerwiseerror}

The layer-wise error analysis follows by recursing the one-step bound
on the difference between the exact and numerical flow:
\[
	&\|F_N\circ \dots \circ F_1 x - \hF_N \circ \dots \circ \hF_1 x\|\\ 
	&\leq \|F_N\circ \dots \circ F_1 x - F_N \circ \hF_{N-1}\circ \dots \hF_1 x\| + \|F_N \circ \hF_{N-1}\circ \dots \hF_1 x - \hF_N \circ \dots \circ \hF_1 x\|\\
	&\leq \mathrm{Lip}(F_N)\|F_{N-1}\circ \dots \circ F_1 x - \hF_{N-1}\circ \dots \circ \hF_1 x\| + \delta\\
	&\leq \dots \leq \delta\sum_{n=0}^{N-1} \prod_{j=n+2}^{N} \mathrm{Lip}(F_j).
\]

\subsection{Error bounds for normalizing flows}
\bprfof{\cref{prop:nfsample}} 
Let $S$ be the random initial state of the orbit that shadows the pseudo-orbit of $X\distas q_0$.
By triangle inequality, 
\[  
\left|\EE f(X_N) - \EE f(\hX_N)\right| &\leq \left|\EE f(X_N) - \EE f(S_N)\right| + \left|\EE f(S_N) - \EE f(\hX_N)\right|.
\]
By $\eps$-shadowing and $\ell$-Lipshitz continuity of $f$, 
\[ 
\left|\EE f(S_N) - \EE f(\hX_N)\right| 
\leq \EE \left| f(S_N) - f(\hX_N)\right|
\leq \ell \EE \|S_N - \hX_N\|
\leq \ell \epsilon.
\] 
Next
\[
&\sup_{f : |f|\leq U, \mathrm{Lip}(f) \leq \ell}\left|\EE f(X_N) - \EE f(S_N)\right|\\
&\leq \sup_{f : |f|\leq U}\left|\EE f(X_N) - \EE f(S_N)\right|
= 2U\tvd{X_N}{S_N} = 2U\tvd{X}{S}.
\]
 The last equality is due to the fact that $X_N$ and $S_N$ are the 
 map of $X$ and $S$, respectively, under the bijection $F_N \circ \cdots \circ F_1$,
 and the fact that the total variation is invariant under bijections
 \citep[Theorem 1]{qiao2010study}.  
\eprfof

\bprfof{\cref{thm:nflpdf}}
Let $s\in\mcX$ be the initial state of the backward shadowing orbit.
By triangle inequality,  
\[
    |\log \hq(x) - \log q(x)| \leq   |\log q(x) - \log q(s)| + |\log \hq(x) - \log q(s)|.
\]
For the first term on the right-hand side, note that for some $y$ on the segment from $x$ to $s$,
\[
\left|\log q(x) - \log q(s)\right| &= \left|\grad \log q(y)^T(x-s)\right| \leq \eps \sup_{\|y-x\|\leq \eps} \|\grad\log q(y)\|. \label{eq:nflpdf1}
\]
For the second term,
\[
    \log \hq(x) - \log q(s) &=  \left|\log q_0(\hx_{-N}) - \log q_0(s_{-N}) \right| + 
    \sum_{n=1}^{N}\left| \log J_{N-(n-1)}(\hx_{-n}) - \log J_{N-(n-1)}(s_{-n}) \right|. 
\]
We apply the same technique to bound each term as in \cref{eq:nflpdf1}.
\eprfof

\bprfof{\cref{thm:nfelbo}}
By definition, $\ELBO{q}{p} = \EE\left[E(X)\right]$, and
by hypothesis the distribution $q_0$ of the initial flow state is equal
to the distribution $\xi_0$ of the shadowing orbit initial state, 
so $\EE\left[E(X)\right] = \EE\left[E(S)\right]$.
Hence $\left|\ELBO{q}{p} - \EE\left[\hE(X)\right]\right| \leq \EE\left[\left|E(S) - \hE(X)\right|\right]$.
Finally,
\[
&\left|E(s) - \hE(x)\right|\\
&= \left| \left(\log p(s_N) - \log q_0(s) + \sum_{n=1}^N \log J_n(s_n)\right) - \left(\log p(\hx_N) - \log q_0(x) + \sum_{n=1}^N \log J_n(\hx_n)\right)\right|\\
&\leq \left|\log p(s_N) - \log p (\hx_N)\right| 
+ \left|\log q_0(s) - \log q_0(x)\right|
+ \sum_{n=1}^N \left|\log J_n(s_n) - \log J_n(\hx_n)\right|.
\]
The proof is completed by bounding each difference with the local Lipschitz
constant around the pseudo-orbit times the shadowing window size $\epsilon$,
and applying the expectation.
\eprfof

\subsection{Error bounds for MixFlows}

\bprfof{\cref{prop:mixsample}}
Let $S$ be the initial point of the random shadowing orbit. By the definition of MixFlows,
\[
\left| \EE f(Y) - \EE f(\hY)\right| &= \left| \EE\left[ \frac{1}{N+1}\sum_{n=0}^N f(X_n) - f(S_n) + f(S_n) - f(\hX_n) \right]\right|\\
&\leq \left| \EE\left[\frac{1}{N+1}\sum_{n=0}^N f(X_n) - f(S_n) \right]\right| + \EE\left|  \frac{1}{N+1}\sum_{n=0}^Nf(S_n) - f(\hX_n)\right|\\
&\leq \left| \EE\left[\frac{1}{N+1}\sum_{n=0}^N f(X_n) - f(S_n) \right]\right| + \frac{1}{N+1}\sum_{n=0}^N\EE\left| f(S_n) - f(\hX_n)\right|.
\]
To bound the second term, we apply shadowing and the Lipschitz continuity of $f$ 
to show that each $\left|f(S_n) - f(\hX_n)\right| \leq \ell \epsilon$, and hence their average has the same bound.
For the first term, in the case where $N$ is fixed and finite,
\[
\sup_{f:|f|\leq U, \mathrm{Lip}(f)\leq \ell}\left|\EE\left[ \frac{1}{N+1}\sum_{n=0}^N f(X_n) - f(S_n)\right]\right| 
&\leq \frac{1}{N+1}\sum_{n=0}^N \sup_{f:|f|\leq U, \mathrm{Lip}(f)\leq \ell}\left| \EE\left[f(X_n) - f(S_n)\right]\right|\\ 
&\leq \frac{1}{N+1}\sum_{n=0}^N 2U \frac{1}{2}\sup_{f:|f|\leq 1}\left|\EE\left[f(X_n) - f(S_n)\right]\right|\\
&= \frac{1}{N+1}\sum_{n=0}^N 2U \tvd{X_n}{S_n}\\
&= \frac{1}{N+1}\sum_{n=0}^N 2U \tvd{X}{S}\\
&= 2U \tvd{\xi_0}{q_0}.
\]
We now consider the large-$N$ limiting case when $F$ is $\pi$-ergodic and measure-preserving,
$q_0 \ll \pi$ and $\xi_0\ll \pi$,
and $U=\ell=1$. Let $Z = S_K$, $K\distas \distUnif\{0,1,\dots,N\}$, 
and let $W \distas \pi$. In this case, by the triangle inequality,
\[
\sup_{f:|f|\leq U, \mathrm{Lip}(f)\leq \ell}\left| \EE\left[\frac{1}{N+1}\sum_{n=0}^N f(X_n) - f(S_n)\right]\right|
&\leq \bld{Y}{W} + \bld{Z}{W}.
\]
By \cite[Theorem 4.1]{Xu22mixflow}, both terms on the right-hand side converge to 0 as $N\to\infty$.
\eprfof

\blem \label{lem:gibbsineq}
If $\forall i \in [n], a_i, b_i > 0$, then
\[
    \left|\log\frac{a_1 + \dots + a_n}{b_1 + \dots + b_n}\right| \leq \max_{1
    \leq i \leq n} \left\{\left|\log\frac{a_i}{b_i}\right|\right\}
\]
\elem
\bprf 
Define $A \defas \sum_{i=1}^n a_i$, 
\[
    \log\frac{a_1 + \dots + a_n}{b_1 + \dots + b_n} 
     =  - \log \frac{b_1 + \dots + b_n}{a_1 + \dots + a_n}  
     = - \log \sum_{i = 1}^n \frac{a_i}{A} \frac{b_i}{a_i} 
\]
Jensen's inequality yields that
\[
    \log\frac{a_1 + \dots + a_n}{b_1 + \dots + b_n} 
    \leq \sum_{i =1}^n \frac{a_i}{A} \log \frac{a_i}{b_i} \leq \max_{1 \leq i \leq n}
    \left\{\log\frac{a_i}{b_i}\right\}.
\]
Applying the same technique to the other direction yields the result.
\eprf

\bprfof{\cref{thm:mflpdf}}
First, shadowing and the triangle inequality yields
\[ \label{eq:mflpdfb0} 
    |\log \hq(x) - \log q(x)| &\leq |\log q(x) - \log q(s)| + |\log \hq(x) - \log q(s)|\\
    &\leq \eps L_{q,\eps}(x) + |\log \hq(x) - \log q(s)|.
\]
By \cref{lem:gibbsineq}, we have
\[ \label{eq:mflpdfb0} 
    |\log \hq(x) - \log q(s)| &\leq \max_{0 \leq n \leq N} \left|\log \frac{q_0(\hx_{-n})/\prod_{j = 1}^n J(\hx_{-j})}{q_0(s_{-n})/\prod_{j = 1}^n J(s_{-j})}\right|\\ 
    &= \max_{0 \leq n \leq N} \left| \log \hq_n(x) - \log q_n(s)\right|,
\]
where $q_n$ is the density of the length-$n$ normalizing flow with constant forward flow map $F$, 
and $\hq_n$ is its approximation via the pseudo-orbit. Using the same technique as in the proof of \cref{thm:nflpdf},
\[
\max_{0\leq n\leq N}\left| \log \hq_n(x) - \log q_n(s)\right|
&\leq \max_{0\leq n\leq N}\epsilon\cdot
\left(L_{q_0,\eps}(\hx_{-n})+\sum_{j=1}^{n} L_{J,\eps}(\hx_{-j})\right)\\
&\leq \epsilon \cdot \left(\max_{0\leq n \leq
N}L_{q_0,\eps}(\hx_{-n})+\sum_{n=1}^{N} L_{J,\eps}(\hx_{-n})\right).
\]
Combining this with the earlier bound yields the first stated result.
To obtain the second result in the infinite shadowing
setting, we repeat the process of bounding $\max_{0\leq n\leq N}\left| \log \hq_n(x) - \log q_n(s)\right|$,
but rather than expressing each supremum around the pseudo-orbit $\hx_{-n}$, we express it around the shadowing orbit $s_{-n}$
to find that 
\[
\max_{0\leq n\leq N}\left| \log \hq_n(x) - \log q_n(s)\right|
&\leq \epsilon \cdot \left(\max_{0\leq n \leq
N}L_{q_0,\eps}(s_{-n})+\sum_{n=1}^{N} L_{J,\eps}(s_{-n})\right).
\]
We then bound $\max_{0\leq n \leq N}$ with a sum and merge this with the first bound to find that
\[
\limsup_{N\to\infty} \frac{1}{N}|\log \hq(x) - \log q(x)| &\leq 
 \epsilon \cdot \limsup_{N\to\infty}\left( \frac{1}{N} L_{q,\eps}(x) +
 \frac{1}{N} \sum_{n=0}^N L_{q_0,\eps}(s_{-n}) + \frac{1}{N} \sum_{n=1}^{N}
L_{J,\eps}(s_{-n})\right).
\]
Since $L_{q,\eps}(x) = o(N)$ as $N\to\infty$, the first term decays to 0.
The latter two terms are ergodic averages under the backward dynamics initialized at $s$;
if the pointwise ergodic theorem \cite{Birkhoff31}, \cite[p.~212]{Eisner15} holds at $s$,
then the result follows.
However, the pointwise ergodic theorem holds only $\pi$-almost surely for each
of $L_{q_0,\eps}$ and $L_{J,\eps}$.
Denote $\mcZ$ to be the set of $\pi$-measure zero for which the theorem does not apply.
Suppose there is a set $\mcA \subseteq \mcX$ such that $q_0(\mcA) > 0$, and $x\in \mcA$ implies
that $s \in \mcZ$. But then $\xi_0(\mcZ) > 0$, which contradicts $\xi_0 \ll \pi$.
Therefore the pointwise ergodic theorem applies to $s$ for $q_0$-almost every $x\in\mcX$.
\eprfof

\bprfof{\cref{thm:mfelbo}}
By definition, $\ELBO{q}{p} = \EE\left[E(X)\right]$, and
by hypothesis the distribution $q_0$ of the initial flow state is equal
to the distribution $\xi_0$ of the joint shadowing orbit initial state, 
so $\EE\left[E(X)\right] = \EE\left[E(S)\right]$.
Hence $\left|\ELBO{q}{p} - \EE\left[\hE(X)\right]\right| \leq \EE\left[\left|E(S) - \hE(X)\right|\right]$.
So applying the triangle inequality, we find that
\[
\left|E(s) - \hE(x)\right|
	&\leq \left|\frac{1}{N+1}\sum_{n=0}^N \log\frac{p(s_n)}{p(\hx_n)}\right| + \left|\frac{1}{N+1}\sum_{n=0}^N \log\frac{\frac{1}{N+1}\sum_{j=0}^N \frac{q_0(\hx_{n-j})}{\prod_{i=1}^j J(\hx_{n-i})}}{\frac{1}{N+1}\sum_{j=0}^N \frac{q_0(s_{n-j})}{\prod_{i=1}^j J(s_{n-i})}}\right| 
\]
For the first sum, each term is bounded using the local Lipschitz constant,
\[
\left|\frac{1}{N+1}\sum_{n=0}^N \log\frac{p(s_n)}{p(\hx_n)}\right|
&\leq  \frac{1}{N+1}\sum_{n=0}^N \left|\log p(s_n) - \log p(\hx_n)\right|\\
&\leq  \frac{1}{N+1}\sum_{n=0}^N \eps L_{p,\eps}(\hx_n).
\]
For the second, we apply \cref{lem:gibbsineq} and then use the local Lipschitz constants,
\[
\left|\frac{1}{N+1}\sum_{n=0}^N\log\frac{\frac{1}{N+1}\sum_{j=0}^N \frac{q_0(\hx_{n-j})}{\prod_{i=1}^j J(\hx_{n-i})}}{\frac{1}{N+1}\sum_{j=0}^N \frac{q_0(s_{n-j})}{\prod_{i=1}^j J(s_{n-i})}}\right|
&\leq \frac{1}{N+1}\sum_{n=0}^N\left|\log\frac{\frac{1}{N+1}\sum_{j=0}^N \frac{q_0(\hx_{n-j})}{\prod_{i=1}^j J(\hx_{n-i})}}{\frac{1}{N+1}\sum_{j=0}^N \frac{q_0(s_{n-j})}{\prod_{i=1}^j J(s_{n-i})}}\right|\\ 
&\leq \frac{1}{N+1}\sum_{n=0}^N\max_{0\leq j\leq N}\left|\log\frac{\frac{q_0(\hx_{n-j})}{\prod_{i=1}^j J(\hx_{n-i})}}{\frac{q_0(s_{n-j})}{\prod_{i=1}^j J(s_{n-i})}}\right|\\ 
&\leq \frac{1}{N+1}\sum_{n=0}^N\max_{0\leq j\leq N}\left|\log q_0(\hx_{n-j}) - \log q_0(s_{n-j})\right| + \sum_{i=1}^j\left|\log J(\hx_{n-i}) - \log J(s_{n-i})\right|\\
&\leq \eps\left(\frac{1}{N+1}\sum_{n=0}^N\max_{0\leq j\leq N}
L_{q_0,\eps}(\hx_{n-j}) + \sum_{i=1}^j L_{J,\eps}(\hx_{n-i})\right)\\
&\leq \eps\left(\max_{-N\leq n \leq N} L_{q_0,\eps}(\hx_n) +
\frac{1}{N+1}\sum_{n=0}^N\sum_{i=1}^N L_{J,\eps}(\hx_{n-i})\right)\\
&\leq \eps\left( \max_{-N\leq n \leq N} L_{q_0,\eps}(\hx_n) +
\sum_{n=-N}^NL_{J,\eps}(\hx_{n})\right).
\]
Combining this with the previous bound and taking the expectation yields the first result,
\[
\left|\ELBO{q}{p}-\EE\left[\hE(X)\right]\right| \leq 
\eps\cdot\EE\left[\frac{1}{N+1}\sum_{n=0}^N L_{p,\eps}(\hX_n) +\!\!\!\!
\max_{-N\leq n \leq N}\!\! L_{q_0,\eps}(\hX_n) + \!\!\!\sum_{n=-N}^NL_{J,\eps}(\hX_{n})\right].
\]
Once again to arrive at the second result, we redo the analysis but center the supremum Lipschitz constants around
the shadowing orbit points, and replace the maximum with a sum, yielding the bound
\[
\left|\ELBO{q}{p}-\EE\left[\hE(X)\right]\right| &\leq 
\eps\cdot\EE\left[\frac{1}{N+1}\sum_{n=0}^N L_{p,\eps}(S_n) +\!\!\!\!
\max_{-N\leq n \leq N}\!\! L_{q_0,\eps}(S_n) + \!\!\!\sum_{n=-N}^NL_{J,\eps}(S_{n})\right]\\
&\leq \eps\cdot\EE\left[\frac{1}{N+1}\sum_{n=0}^N L_{p,\eps}(S_n) +\!\!\!\!
\sum_{n=-N}^N\!\! L_{q_0,\eps}(S_n) + \!\!\!\sum_{n=-N}^NL_{J,\eps}(S_{n})\right].
\]
Now since $\xi_0 = q_0$ by assumption, $S_n \eqd X_n$ for each $n\in\nats$, so
\[
\left|\ELBO{q}{p}-\EE\left[\hE(X)\right]\right| &\leq 
\eps\cdot\EE\left[\frac{1}{N+1}\sum_{n=0}^N L_{p,\eps}(X_n) +\!\!\!\!
\sum_{n=-N}^N\!\! L_{q_0,\eps}(X_n) + \!\!\!\sum_{n=-N}^NL_{J,\eps}(X_{n})\right].
\]
We divide by $N$ and take the limit supremum:
\[
&\limsup_{N\to\infty}\frac{1}{N}\left|\ELBO{q}{p}-\EE\left[\hE(X)\right]\right| \\
&\leq \eps \limsup_{N\to\infty} \cdot\EE\left[\frac{1}{N(N+1)}\sum_{n=0}^N
L_{p,\eps}(X_n) +\frac{1}{N}\sum_{n=-N}^N\!\! L_{q_0,\eps}(X_n)
+\frac{1}{N}\sum_{n=-N}^NL_{J,\eps}(X_{n})\right].
\]
Consider just the term
\[
    \EE\left[\frac{1}{N}\sum_{n=0}^N L_{J,\eps}(X_n)\right]
&= \int \frac{1}{N}\sum_{n=0}^N L_{J,\eps}(F^n x) q_0(\dee x)
= \int \frac{1}{N}\sum_{n=0}^N L_{J,\eps}(F^n x) \der{q_0}{\pi}(x) \pi(\dee x).
\]
\cite[Theorem 8.10 (vi)]{Eisner15} 
asserts that as long as $L_{J,\eps} \in L^{m_1}(\pi)$ and $\der{q_0}{\pi} \in L^{m_2}(\pi)$ for $1/m_1+1/m_2 = 1$, $1 \leq m_1 < \infty$,
then
\[
    \lim_{N\to\infty} \int \frac{1}{N}\sum_{n=0}^N L_{J,\eps}(F^n x) \der{q_0}{\pi}(x) \pi(\dee x)
    = \int L_{J,\eps}(x) \pi(\dee x) \int \der{q_0}{\pi} \pi(\dee x) =
    \EE\left[L_{J,\eps}(X)\right], \,\, X\distas \pi.
\]
Applying this result to each term above yields the stated result.
\eprfof

\subsection{Norm of matrix right inverse}
Notice that the operator $A$ in \cref{thm:finiteshadowing} has a following matrix
representation: 
\[
\label{eq:matrixL}
A = \bbmat 
-D_1 & I & & & \\ 
    & -D_2 & I & & \\
    &  & \ddots & \ddots & \\ 
    &  & & -D_N & I \\ 
    \ebmat
\in \reals^{dN \times d(N+1)}, \text{ where } D_k = \nabla F_k(\hx_{k-1}), \forall k
\in [N].   
\] 
\bprop[{\citep[Corollary 4.2]{dokmanic2017beyond}}] \label{prop:svdbound}
Let $A$ be as defined in \cref{eq:matrixL}.
Suppose $(F_n)_{n = 1}^N$ are all differentiable. Then 
\[ \label{eq:bestinverse}
    A^{\dagger}\defas  A^T(AA^T)^{-1} \in \arg\min_{X}\|X\|, \quad \text{subject
    to } A X = I,
\]
and $\|A^\dagger\|= \sigma_{\min}(A)^{-1}$, where $ \sigma_{\min}(A)$ denotes
the smallest singular value of $A$.
\eprop

\bprfof{\cref{prop:svdbound}} 
Note that $A$ has full row rank due to the identity blocks. In this case, 
$AA^{T}$ is invertible, thus $A^T(AA^T)^{-1}$ is a valid right inverse of $A$.
\cref{eq:bestinverse} is then obtained by a direct application of \citep[Corollary 4.2]{dokmanic2017beyond}.

To obtain the norm of $A^\dagger$, since $A$ has full row rank,
\[
    A = U\Sigma V^T, \quad \text{where } 
    &U \in \reals^{dN\times dN}, V\in
    \reals^{d(N+1)\times d(N+1)} \text{ are orthonormal matrices}, \\	
    &\Sigma \in \reals^{dN \times d(N+1)} \text{ is a rectangular diagonal
    matrix with full row rank}.
\]
Then 
\[
    \|A^\dagger\| 
    &= \left\|V\Sigma^T U^T \left( U\Sigma V^TV\Sigma^T U^T \right)^{-1}\right\|  \\ 
    & = \left\| V \Sigma^T U^T U (\Sigma \Sigma^T)^{-1} U^T\right\|\\ 
    &= \|\Sigma(\Sigma\Sigma^T)^{-1}\| \\ 
    &= \sigma_{\min}(A)^{-1}
\]
Notice that since $AA^T$ is an invertible Hermitian matrix, $\sigma_{\min}(A)$
is strictly positive.
\eprfof

\section{Discussion about the shadowing window $\eps$}

\subsection{Numerical error when computing $\eps$} \label{sec:windowcomputeerror}
As mentioned in \cref{sec:shadowing}, due to the floating-point error $\delta$ involved
in evaluating $\nabla F_k$, the evaluation of $A$ is perturbed by the numerical
error. This consequntely leads to an inaccuracy of the estimation of $\lambda= \|A_r^{-1}\|$. 
Here $A_r^{-1}$ denotes the right inverse of $A$.
In this section, we explain how to calibrate the calculation of $\lambda$ to
take into account of this error.

This computational challenge, rooted in numerical inaccuracies, was previously
acknowledged in \citet[Section 3.4]{coomes1996shadowing}, with a provided
remedy.
Suppose we compute terms in $\nabla F_k(\hx_{k-1})$ with a floating-point error
$\delta$. This introduces $O(\sqrt{N}\delta)$ Frobenius norm error into the
practically calculated $A$. Here we denote the digital computed $A$ by $\tilde
A$, $\tilde\lambda := \|\tilde A_r^{-1}\|$, and $\tilde\delta := \|A - \tilde A\|$.
According to \citet[Eq. ~45]{coomes1996shadowing}, we can employ the following upper bound
on $\lambda$:
\[
    \lambda \leq (1 - \tilde\delta \tilde\lambda)^{-1}\tilde\lambda 
    = \left(1 - O(\sqrt{N\delta}) \tilde\lambda\right)^{-1}\tilde\lambda.
\]
Substituting values of $\delta$, $N$, and computed $\tilde \lambda$ from our
experiments (refer to \cref{fig:shadowingwindow}), this amounts to an inconsequential
relative error of around $(1 - 10^{-9})^{-1} \approx 1$ on $\lambda$.

It is also worth discussing the the shadowing window computation methods
proposed in \citet[Section 3.4]{coomes1996shadowing}. As opposed to our
generic eigenvalue computation introduced in \cref{sec:shadowing}, the computation
procedure presented in \citet[Section 3.4]{coomes1996shadowing} can potentially
offer better scalability.
However, the methodology in \citet{coomes1996shadowing} operates under the heuristic
that a dynamical system with shadowing is intrinsically hyperbolic, or nearly so.
Specifically, the ``hyperbolic splitting'' discussed in \citet[P.413, Appendix
B]{coomes1995rigorous} (i.e., the choice of $\ell$), and the ``hyperbolic
threshold'' choice in \citet[P.415, Appendix B]{coomes1995rigorous} (i.e., the
choice of $p$), both implicitly assume that the dynamics either exhibit hyperbolicity or tend closely to it. 
Such presuppositions might be overly restrictive for generic variational flows,
given that the underlying  dynamics are general time-inhomogeneous systems.
When the underlying dynamics are not (nearly) hyperbolic, applying strategies
from \citet{coomes1996shadowing} could potentially lead to a poor estimation of the shadowing window.

\subsection{Dependence of $\eps$ on $N$}
\label{apdx:windowsizescaling}

One may notice that $\eps$ depends on $N$ implicitly through
$\lambda_{\min}(AA^T)^{-1/2}$.
Recall that $A$ can be represented as a matrix
\[
    A = \bbmat 
    -D_1 & I &  & &  \\ 
	& -D_2 & I &  &\\
	&  & \ddots &\ddots &\\ 
	&  & & -D_N &I \\ 
	\ebmat
    \in \reals^{dN \times d(N+1)}, \text{ where }D_k = \nabla F_k(\hx_{k-1}), \forall k
    \in [N].   
\] 
And the shadowing window is given by $\eps = 2\delta \lambda_{\min}(AA^T)^{-1/2}$.
If $\lambda_{\min}(AA^T)^{-1/2}$ scales badly with $N$, our $O(N\eps)$ error bounds become vacuous. 
Therefore, in this section, 
we collect some insights about the scaling of $\eps$ over increasing $N$.
It is important to note that understanding the explicit quantitative
relationship between $\eps$ and $N$ for general nonautonomous dynamical systems
is quite challenging. 
Although current literature of dynamical systems focuses on studying
conditions so that $\eps$ does not scale with $N$, these conditions are
typically difficult to verify in practice or are only applicable in restrictive
settings such as time-homogeneous linear dynamical systems. 
For a more systematic study, see \citep{backes2019shadowing,coomes1996shadowing,pilyugin2011theory}.
The examples provided below merely serve as synthetic
illustrations for intuition and should not be taken as definitive statements for
all cases.

\bexa[Scaling map in $\reals$] \label{exa:scalingmap}
Consider the time-homogeneous dynamics formed by 
$T: \reals \mapsto \reals$ such that $\forall x\in \reals, T(x) = C x$ for some $C >0$.
Then the corresponding shadowing window of a $\delta$-pseudo-orbit is
\[ \label{eq:scalingmap}
\eps = 2\delta  \left[(C - 1)^2 + 2C\left(1- \cos \frac{\pi}{N+1}\right)
    \right]^{-\frac{1}{2}}.
\]
Therefore, for all $C \neq 1$, as $N \to \infty$, $\eps = O(\delta)$. 
For $C = 1$, $\eps = O(N\delta)$.  
The formula for $\lambda_{\min}(AA^T)$ above 
arises because in this case, $AA^T$ has the following symmetric tridiagonal form,  
\[
    AA^T = \bbmat 
    C^2 + 1 & -C &  & &  \\ 
	-C & C^2+1 &  -C &  &\\
	&  \ddots & \ddots &\ddots &\\ 
	&  & -C & C^2+1 & -C \\ 
	&  & & -C & C^2+1 \\ 
	\ebmat, 
\]
whose eigenvalues have the following closed form \citep[Page.~19--20]{elliott1953characteristic}:
\[
\lambda_{k} (AA^T) =  C^2+1 - 2C \cos \frac{k \pi}{N+1}, \quad k = 1, \dots, N.
\]
\eexa

\bexa[Uniform hyperbolic linear map{\citep[Theorem 12.3 and p.~171]{pilyugin2019spaces}}]
\label{exa:hyperbolic}
Consider the time-homogeneous linear dynamical system in $\reals^d$, $T: x\mapsto
Ax$, where $A \in \reals^{d\times d}$ is similar to a block-diagonal matrix
$\diag(A_1, A_2)$ such that $\|A_1\| \leq \lambda$ and $\|A_2^{-1}\|\leq
\lambda$  for some $\lambda \in (0, 1)$.
Then for all $N \in \nats$, any $\delta$-pseudo-orbit is $\eps$-shadowed,
with $\eps = \frac{1+\lambda}{1-\lambda}\delta$.
\eexa

As illustrated in \cref{exa:scalingmap,exa:hyperbolic}, it is possible
for the $\eps$-shadowing property to hold in a way that $\eps$ does not depend
on the flow length $N$ and remains in a similar magnitude as $\delta$. This
phenomenon is referred to as the \emph{infinite shadowing property}
\citep{coomes1996shadowing,pilyugin1999shadowing,pilyugin2006shadowing} as
mentioned in the beginning of \cref{sec:mixflow}.
The infinite shadowing property is shown to be \emph{generic} in the space of all dynamical
systems \citep{pilyugin1999shadowing}, meaning that dynamical systems exhibiting
this property form a dense subset of the collection of all dynamical systems.
However, while genericity offers valuable insights
into the typical behavior of dynamical systems, it does not imply that all
systems possess the property. As previously explained, only a small class of
systems has been shown to have the infinite shadowing property, and verifying
this property for a specific given system remains a challenging task.

\section{Normalizing flow via Hamiltonian dynamics} \label{sec:hamiltonianvi}
While past work on normalizing flows
\citep{Dinh17,Durkan19,vansylvester,Rezende15,Papamakarios21} typically builds a flexible flow family by composing
numerous parametric flow maps that are agnostic to the target distribution, 
recent target-informed variational flows focus on designing architectures that
take into account the structures of the target distribution. A common
approach for constructing such target-informed flows is to use Langevin and Hamiltonian
dynamics as part of the flow transformations
\citep{Caterini18,Chen22,Geffner21,Zhang21,geffner2023langevin,Thin21}.
In this section, we begin with a concise overview of two Hamiltonian-based
normalizing flows: Hamiltonian variational auto-encoder (HVAE)
\citep{Caterini18} and Hamiltonian flow (HamFlow) \citep{Chen22}.
Subsequently, we present corollaries derived from \cref{thm:nflpdf,thm:nfelbo},
specifically applying these general results to HVAE and
HamFlow. Finally, we offer empirical evaluations of our theory for
HamFlow on the same synthetic examples as those examined for MixFlow.

Hamiltonian dynamics (\cref{eq:hamiltonian}) describes the evolution of a
particle's position $\theta_t$ and momentum $\rho_t$ within a physical system. This
movement is driven by a differentiable negative potential energy $\log p(\theta_t)$
and kinetic energy $\frac{1}{2} \rho_t^T\rho_t$:
\[\label{eq:hamiltonian}
\frac{\mathrm{d} \rho_t}{\mathrm{~d} t}=\nabla \log \pi\left(\theta_t\right) \quad
\frac{\mathrm{d} \theta_t}{\mathrm{~d} t}=\rho_t.
\]
For $t\in \reals$, we define the mappings $H_t:\reals^{2d}\to \reals^{2d}$, 
transforming $(\theta_s, \rho_s) \mapsto (\theta_{t+s}, \rho_{t+s})$ according
to the dynamics of \cref{eq:hamiltonian}. 
Intuitively, traversing the trajectory of $H_t, t\in \reals$ allows particles
to efficiently explore the distribution of interest's support, thereby
motivating the use of $H_t$ as a flow map.
In practice, the exact Hamiltonian flow $H_t$ is replaced
by its numerical counterpart---leapfrog integration---which involves $K$
iterations of the subsequent steps:
\[
\hat{\rho}\gets\rho+\frac{\epsilon}{2} \nabla \log \pi\left(\theta\right) &&
\theta \gets \theta+\epsilon \hat{\rho} && \rho \gets \hat{\rho}+\frac{\epsilon}{2}
\nabla \log \pi \left(\theta\right).
\]
Here $\eps$ denotes the step size of the leapfrog integrator. 
We use $T_{K, \eps}: \reals^{2d} \to \reals^{2d}$ to define the map by
sequencing the above three steps $K$ times.
There are two key properties of $T_{K, \eps}$ that make it suitable for normalizing flow
constructions. 
$T_{K, \eps}$ is invertible (i.e., $T_{K, \eps}^{-1} = T_{K, -\eps}$) and
has a simple Jacobian determinant of the form: $|\det \nabla T_{K, \eps}| = 1$. 
This enables a straightforward density transformation formula: 
for any density $q_0(\theta, \rho)$ on $\reals^{2d}$, pushforward of $q$ under
$T_{K, \eps}$ has the density $q_0(T_{K, -\eps}(\theta, \rho))$.

Given the above properties, \citet{Caterini18} and \citet{Chen22} developed flexible 
normalizing flows---HVAE and HamFlow, respectively—utilizing
flow transformations through alternating compositions of multiple $T_{K,
\epsilon}$ and linear transformations. 
Notably, the momentum variable $\rho$ introduced in the Hamiltonian dynamics
\cref{eq:hamiltonian} necessitates that both HVAE and HamFlow
focus on approximating the \emph{augmented target distribution} $\bpi(\theta,
\rho) \propto\pi(\theta) \cdot \exp(-\frac{1}{2}\rho^T \rho)$ on $\reals^{2d}$. This represents the
joint distribution of the target $\pi$ and an independent momentum distribution
$\distNorm(0, I_d)$.

Specifically, flow layers of HVAE alternate between the leapfrog transformation
$T_{K, \eps}$ and a momentum scaling layer:
\[\label{eq:momscaling}
    (\theta, \rho) \mapsto (\theta, \gamma \rho), \quad \gamma \in \reals^+.
\]
And those of HamFlow alternate between $T_{K, \eps}$ and a momentum
standardization layer:
\[ \label{eq:momstandard}
    (\theta, \rho) \mapsto (\theta, \Lambda(\rho - \mu)), \quad \Lambda \in
    \reals^{d\times d} \text{ is lower
    triangular with positive diagonals }, \mu \in \reals^d.
\]

\subsection{Error analysis for Hamiltonian-based flows} \label{sec:hamflowtheory}
We have provided error analysis for generic normalizing flows in
\cref{sec:normalizingflow}. In this section, we specialize the error bounds for
density evaluation (\cref{thm:nflpdf}) and ELBO estimation (\cref{thm:nfelbo}) 
for the two Hamiltonian-based flows. $\pi$ denotes the target distribution
of interest on $\reals^d$, and $\bp$ denotes the augmented target distribution
on $\reals^{2d}$. $p$ and $\bp$ respectively denote the unnormalized density of
$\pi$ and $\bpi$.

\bcor \label{cor:hamflowlpdf}
Let $q$ be either \emph{HVAE} or \emph{HamFlow} of $N$ layers with $q_0  =
\distNorm(0, I_{2d})$. 
Suppose the backward dynamics has the $(\eps,\delta)$-shadowing property.
Then 
\[
    \forall x \in \reals^{2d}, \quad |\log \hq(x) - \log q(x)| \leq  \epsilon \cdot
   \left(L_{q,\eps}(x) + \|\hx_{-N}\| \right) + \eps^2. 
\]
\ecor
\bprfof{\cref{cor:hamflowlpdf}}
Since the Jacobian determinant of each HVAE/HamFlow layer is constant,
\[
    L_{J_n,\eps}(x) = 0, \qquad \forall n \in [N] \text{ and } x \in \reals^{2d}. \label{eq:hamflowJ} 
\]
And since $q_0$ is a standard normal distribution, 
\[
    L_{q_0, \eps} = \sup_{\| y -x\|\leq \eps} \|y\| \leq \|x\| + \eps.
    \label{eq:hamflowq0}
\]
The proof is then complete by directly applying \cref{eq:hamflowJ,eq:hamflowq0}
to the error bound stated in \cref{thm:nflpdf}.
\eprfof

\bcor \label{cor:hamflowelbo}
Let $q$ be either \emph{HVAE} or \emph{HamFlow} of $N$ layers with $q_0  =
\distNorm(0, I_{2d})$.
Suppose the forward dynamics has the $(\eps,\delta)$-shadowing property,
and $\xi_0 = q_0$. Then
\[
    \abs{\ELBO{q}{\bp} - \EE[\hE(X)]} \leq \eps \cdot \left(\EE\left[
    L_{\bp,\eps}(\hX_N) \right] + \sqrt{2d}\right)+ \eps^2 \qquad \text{for } X\distas q_0.
\]
If additionally $\log p$ is $L$-smooth (i.e., $\forall x, y, \|\nabla \log p(x)
- \nabla \log p(y)\|\leq L\|x-y\|$), then 
\[
    \abs{\ELBO{q}{\bp} - \EE[\hE(X)]} \leq \eps \cdot \left(\EE\left[ \|\nabla
    \log \bp(\hX_N)\| \right] + \sqrt{2d}\right)+ (L \vee 1 +1)\eps^2 \qquad \text{for } X\distas q_0.
\]
\ecor

\bprfof{\cref{cor:hamflowelbo}}
Similar to the proof of \cref{fig:cross_lpdf}, applying \cref{eq:hamflowJ,eq:hamflowq0}
to the error bound stated in \cref{thm:nfelbo} yields that
\[
    \abs{\ELBO{q}{\bp} - \EE[\hE(X)]} \leq \eps \cdot \EE\left[
	L_{\bp,\eps}(\hX_N) + \|X\|\right] + \eps^2 \qquad \text{for } X\distas q_0.
\]
The first result is then obtained by employing the following bound via Jensen's
inequality:
\[
    \EE[\|X\|] \leq \sqrt{\EE[\|X\|^2]} = \sqrt{2d} \qquad \text{for } X\distas
    q_0 = \distNorm(0, I_{2d}).
\]

To arrive at the second result, we apply the following bound on $L_{p, \eps}$ using
the fact that $\nabla \log p$ is $L$-Lipschitz continuous: 
\[
    L_{\bp, \eps} \leq \|\nabla \log \bp(x)\| + \sup_{\|y - x\|\leq \eps}\|\nabla \log
    \bp(x) - \nabla \log \bp(y)\| \leq \|\nabla \log \bp(x)\| + L\vee 1 \cdot\eps.
\]
\eprfof

\subsection{Empirical results for Hamiltonian flow } \label{apdx:nfresults}
\begin{figure*}[t!]
\centering 
\begin{subfigure}[b]{0.48\columnwidth} 
    \scalebox{1}{\includegraphics[width=\columnwidth]{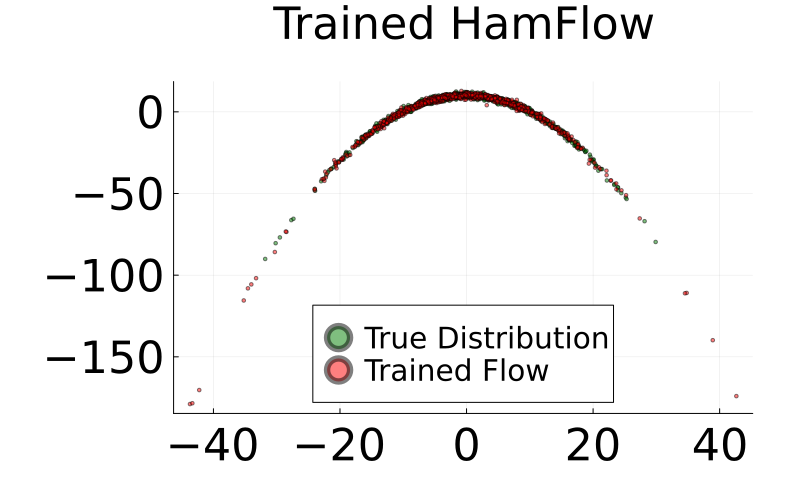}}
    \caption{(a) Banana: \texttt{HamFlow} sample scatters} \label{fig:banana_hamflowscatter}
\end{subfigure}
\hfill
\centering 
\begin{subfigure}[b]{0.48\columnwidth} 
    \scalebox{1}{\includegraphics[width=\columnwidth]{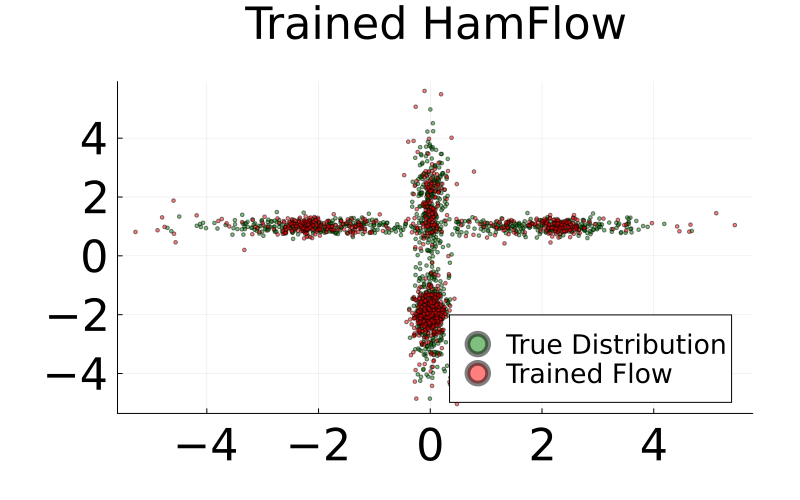}}
    \caption{(b) Cross: \texttt{HamFlow} sample scatter}\label{fig:cross_hamflowscatter}
\end{subfigure}
\hfill
\centering 
\begin{subfigure}[b]{0.24\columnwidth} 
    \scalebox{1}{\includegraphics[width=\columnwidth]{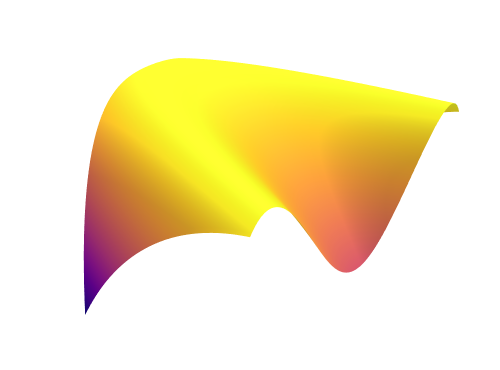}}
    \caption{(c) True log-pdf} \label{fig:banana_hamflow_truelpdf}
\end{subfigure}
\hfill
\centering 
\begin{subfigure}[b]{0.24\columnwidth} 
    \scalebox{1}{\includegraphics[width=\columnwidth]{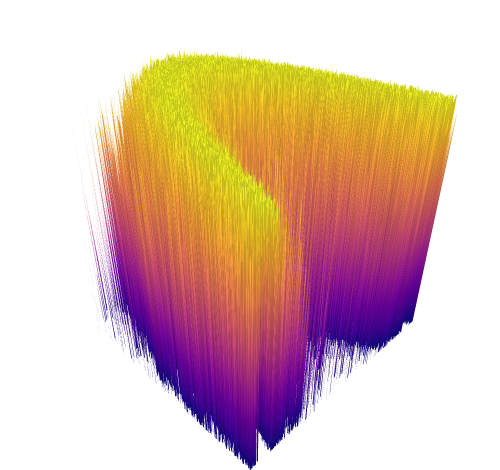}}
    \caption{(d) \texttt{HamFlow} log-pdf}\label{fig:banana_hamflow_lpdfest}
\end{subfigure}
\hfill
\centering 
\begin{subfigure}[b]{0.24\columnwidth} 
    \scalebox{1}{\includegraphics[width=\columnwidth]{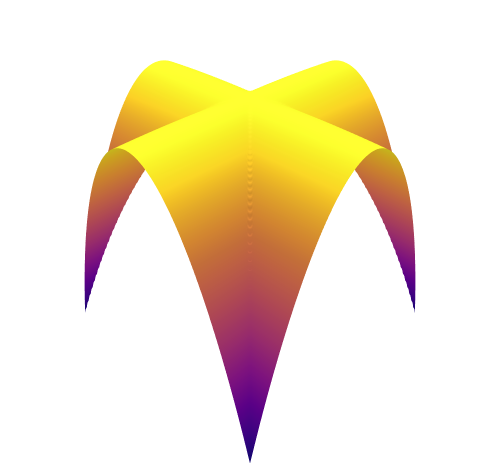}}
    \caption{(e) True log-pdf} \label{fig:cross_hamflow_true}
\end{subfigure}
\hfill
\centering 
\begin{subfigure}[b]{0.24\columnwidth} 
    \scalebox{1}{\includegraphics[width=\columnwidth]{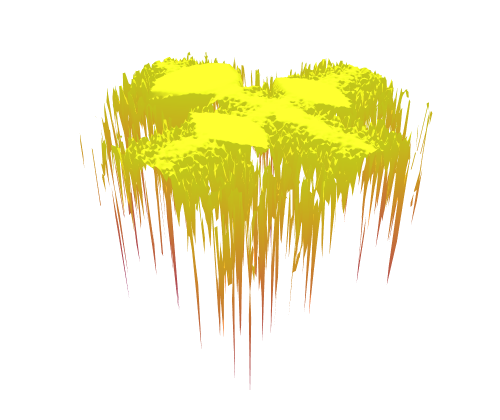}}
    \caption{(f) \texttt{HamFlow} log-pdf} \label{fig:cross_hamflow_lpdfest}
\end{subfigure}
\hfill
\caption{
    Visualization of sample scatters and density of \texttt{HamFlow}. Figure
    (a)--(b) each shows 1000 \iid draws from \texttt{HamFlow}
    distribution ({\color{red}red}) and target distribution ({\color{green}green}).
    (c) and (e) show respectively the exact log-density of the Banana and the
    cross distribution, while (d) and (f) show respectively the sliced 
    \texttt{HamFlow} log-density evaluation at $\rho = (0, 0)$ on those two
    targets. The \texttt{HamFlow} log-density evaluations are computed via $2048$-big \texttt{BigFloat}
    representation.
} \label{fig:hamflowvis}
\end{figure*}

In this section, we empirically validate our error bounds and the diagnostic
procedure of HamFlow using two synthetic targets: the banana and cross
distributions. We employ a warm-started HamFlow, as detailed in
\citep[Section 3.4]{Chen22}. This is achieved through a single pass of 10,000
samples from the reference distribution, $q_0$, without additional optimization.
Specifically, the $\mu$ and $\Lambda$ values for each flow standardization layer
are determined by the sample mean and the Cholesky factorization of the sample
covariance, respectively, of the batched momentum samples processed through the layer. 
For the banana distribution, we set the
leapfrog step $K$ at $200$ and a step size $\eps$ of $0.01$. For the cross
distribution, we set $K$ to $100$ with an $\eps$ value of $0.02$.
\cref{fig:banana_hamflowscatter,fig:cross_hamflowscatter} showcase a comparison between $1000$ \iid
samples from the warm-started HamFlow with $400$ layers and the true targets.  
The results illustrate that HamFlow generates samples of decent quality.

\begin{figure*}[t!]
\centering 
\begin{subfigure}[b]{0.24\columnwidth} 
    \scalebox{1}{\includegraphics[width=\columnwidth]{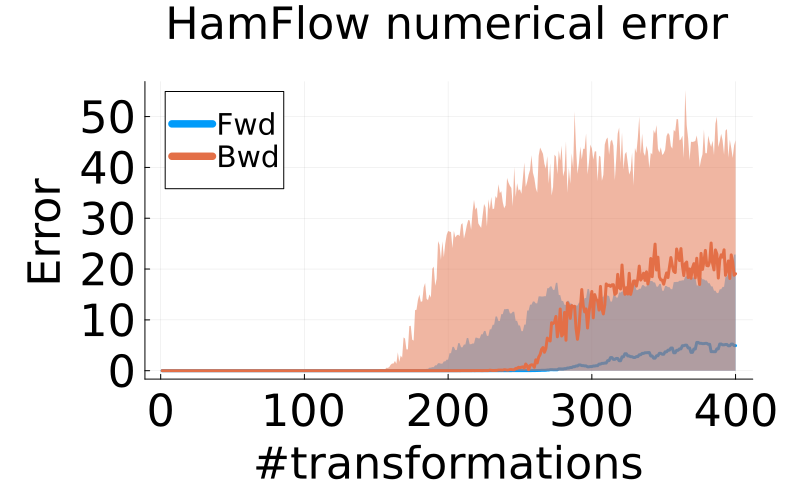}}
\caption{(a) Banana}
\end{subfigure}
\hfill
\centering 
\begin{subfigure}[b]{0.24\columnwidth} 
    \scalebox{1}{\includegraphics[width=\columnwidth]{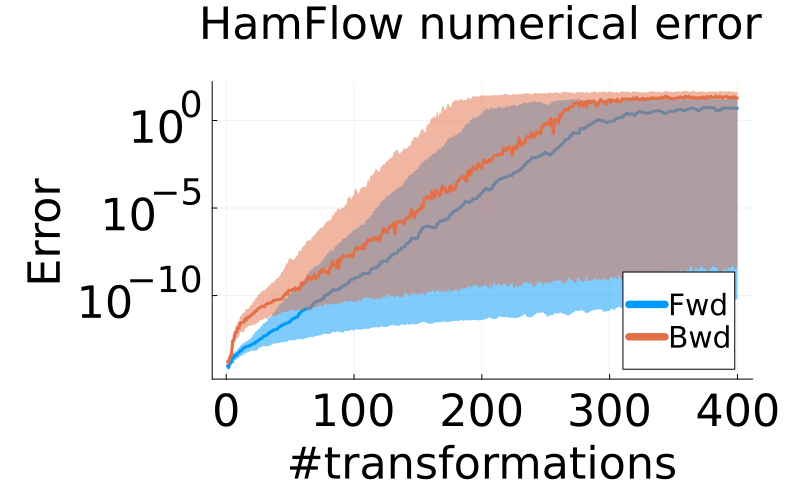}}
\caption{(b) Banana (log-scale)}
\end{subfigure}
\hfill
\centering 
\begin{subfigure}[b]{0.24\columnwidth} 
    \scalebox{1}{\includegraphics[width=\columnwidth]{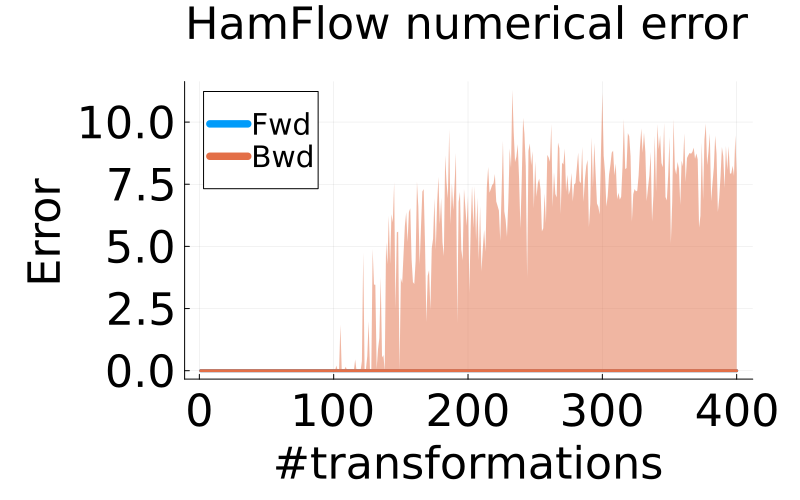}}
\caption{(c) Cross}
\end{subfigure}
\hfill
\centering 
\begin{subfigure}[b]{0.24\columnwidth} 
    \scalebox{1}{\includegraphics[width=\columnwidth]{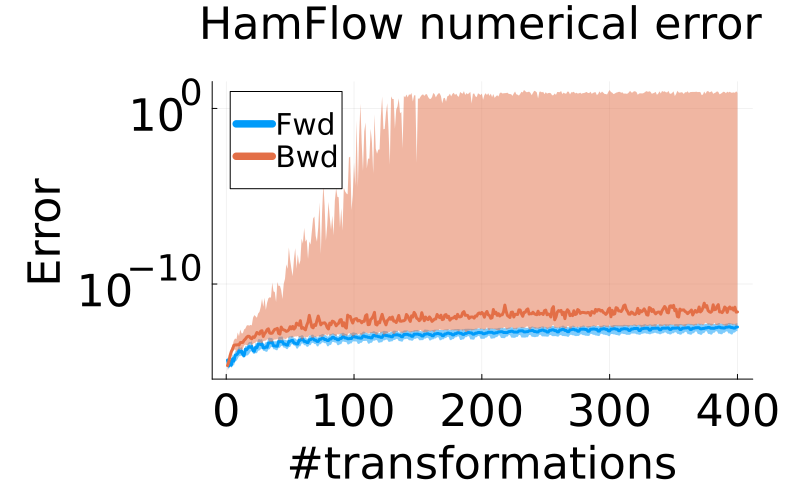}}
    \caption{(d) Cross (log-scale)}
\end{subfigure}
\hfill
\caption{
    HamFlow forward (\texttt{fwd}) and backward (\texttt{bwd}) orbit
    errors on synthetic examples.
(a) and (c) show the median and upper/lower quartile 
forward error $\|F^k x - \hF^k x\|$ and backward error $\|B^k y - \hB^{k}y\|$
comparing $k$ transformations of the forward exact/approximate maps $F$, $\hF$
or backward exact/approximate maps $B$, $\hB$.
And (b) and (d) respectively display (a) and (c) in a logarithmic scale, intended to provide a clearer
illustration of the exponential growth of the error.
The lines indicate the median, and error regions indicate $25^\text{th}$ to
$75^\text{th}$ percentile from $100$ independent runs, which take an \iid draw
of $x$ from $q_0$ and draw $y$ from the actual target distribution $p$.
} \label{fig:hamfloworbiterr}
\end{figure*}

\begin{figure*}[t!]
\centering 
\begin{subfigure}[b]{0.24\columnwidth} 
    \scalebox{1}{\includegraphics[width=\columnwidth]{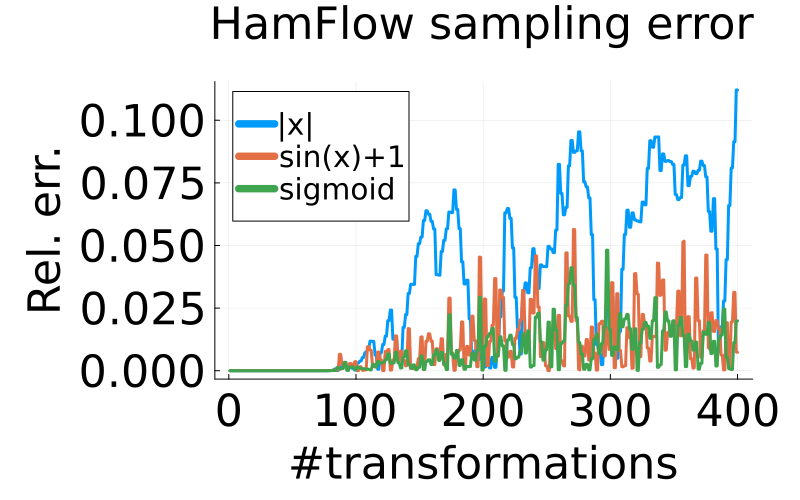}}
\caption{(a) Banana}
\end{subfigure}
\hfill
\centering 
\begin{subfigure}[b]{0.24\columnwidth} 
    \scalebox{1}{\includegraphics[width=\columnwidth]{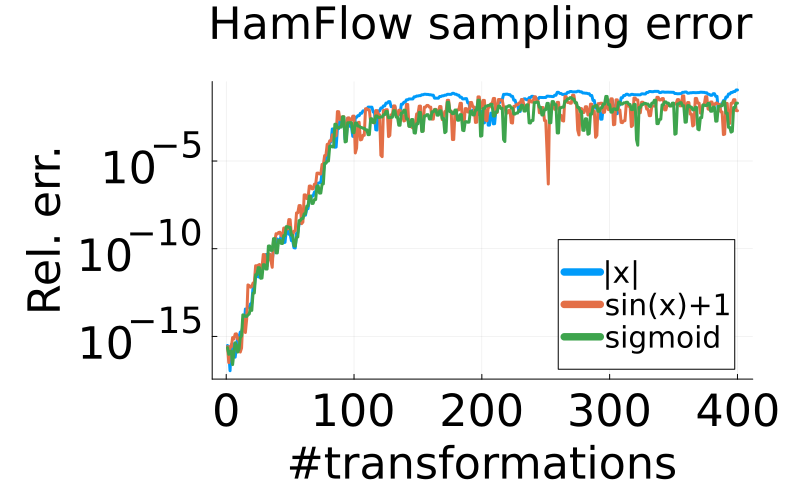}}
\caption{(b) Banana (log-scale)}
\end{subfigure}
\hfill
\centering 
\begin{subfigure}[b]{0.24\columnwidth} 
    \scalebox{1}{\includegraphics[width=\columnwidth]{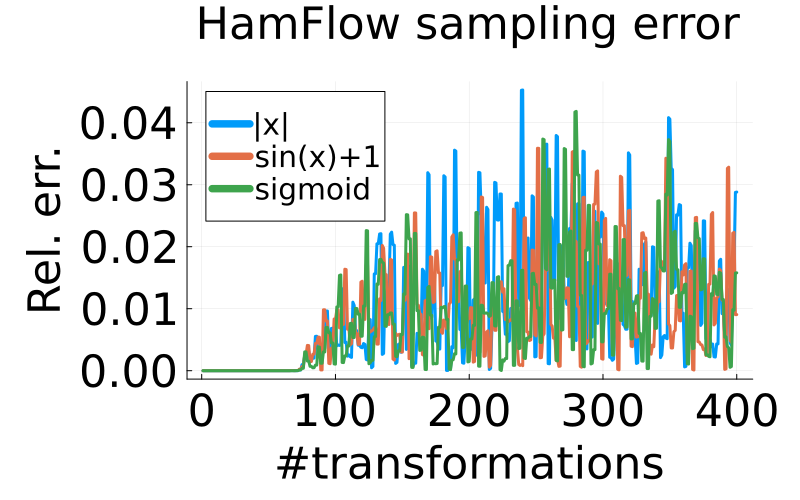}}
\caption{(c) Cross}
\end{subfigure}
\hfill
\centering 
\begin{subfigure}[b]{0.24\columnwidth} 
    \scalebox{1}{\includegraphics[width=\columnwidth]{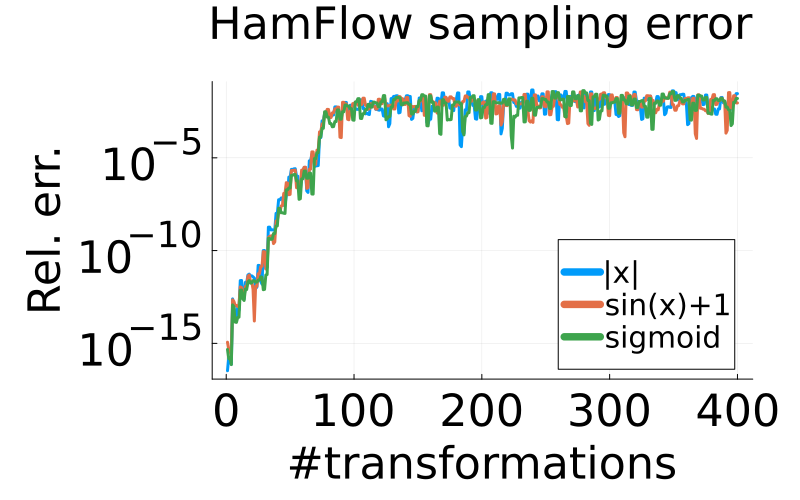}}
\caption{(d) Cross (log-scale)}
\end{subfigure}
\hfill
\caption{
    HamFlow relative sample average computaion error on three test
    functions: 
    $\sum_{i = 1}^d[|x|]_i, \sum_{i = 1}^d[\sin(x) + 1]_i$ and $\sum_{i = 1}^d[\mathrm{sigmoid}(x)]_i$. 
Each line denotes the numerical error computed from Monte Carlo average of the three test
functions via $100$ independent forward orbits starting at \iid draws from
$q_0$.  
} \label{fig:hamflowsample}
\end{figure*}

\begin{figure*}[t!]
\centering 
\begin{subfigure}[b]{0.45\columnwidth} 
    \scalebox{1}{\includegraphics[width=\columnwidth]{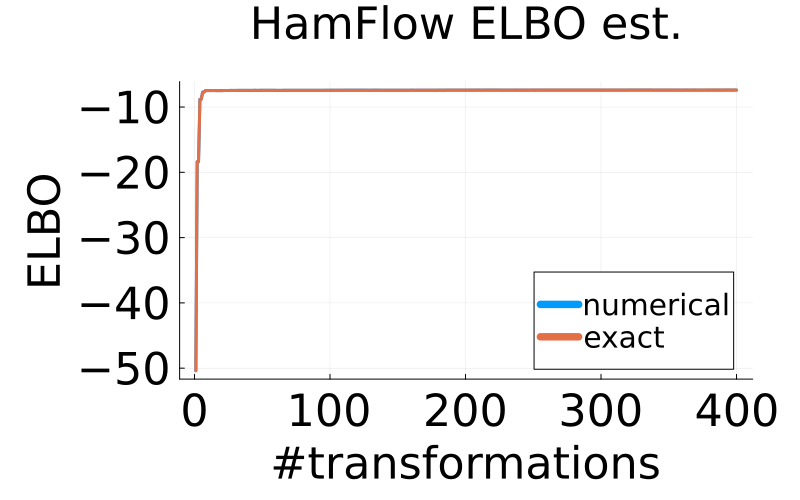}}
\caption{(a) Banana}
\end{subfigure}
\centering 
\begin{subfigure}[b]{0.45\columnwidth} 
    \scalebox{1}{\includegraphics[width=\columnwidth]{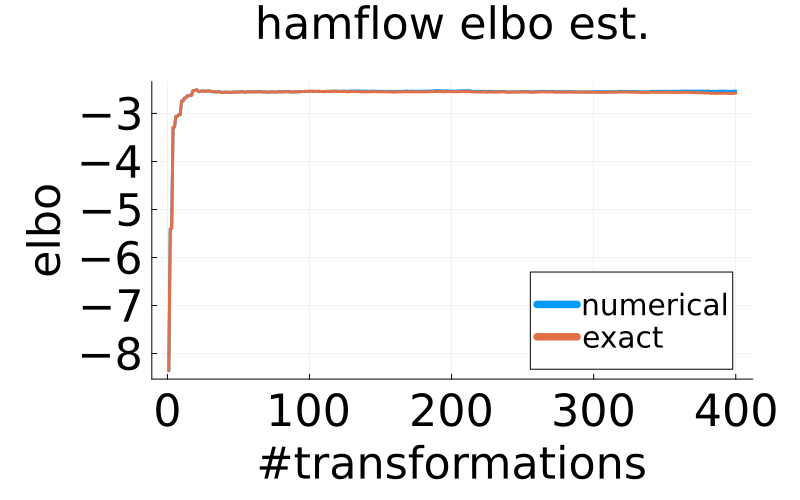}}
    \caption{(b) Cross}
\end{subfigure}
\hfill
\caption{
HamFlow \texttt{exact} and \texttt{numerical} ELBO estimation over increasing flow length.
The lines indicate the averaged ELBO estimates over $100$ independent
forward orbits. The Monte Carlo error for the given estimates is sufficiently small
so we omit the error bar.
} \label{fig:hamflowelbo}
\end{figure*}

We start by examining the numerical orbit computaiton error. As demonstrated in 
\cref{fig:hamfloworbiterr}, the pointwise evaluation error grows
exponentially with increasing flow depth, and quickly reaches the same order of
magnitude as the scale of the target distributions. However, 
similar to what we have observed in the experiments of MixFlow (described in
\cref{sec:expts}), the substantial numerical error of orbit computation does not
impact the accuracy of Monte Carlo estimations of expectations and ELBO
estimations significantly. 
\cref{fig:hamflowsample} shows the relative numerical error of Monte Carlo
estimates of the expectations of three test functions. The relative samples
estimation error remains within $5\%$ for the two bounded Lipschitz continuous
test functions, aligning with our error bound in \cref{prop:nfsample}. 
Moreover, even for the unbounded test function, the error stays within reasonable limits: 
$10\%$ for the banana and $5\%$ for the cross examples.
\cref{fig:hamflowelbo} then compares the numerical ELBO estimates and exact
ELBO estimates for HamFlow over increasing flow length. In both
examples, the numerical ELBO curve is almost identical to the exact one, 
notwithstanding the considerable sample computation error of the forward orbit.

\begin{figure*}[t!]
\centering 
\begin{subfigure}[b]{0.24\columnwidth} 
    \scalebox{1}{\includegraphics[width=\columnwidth]{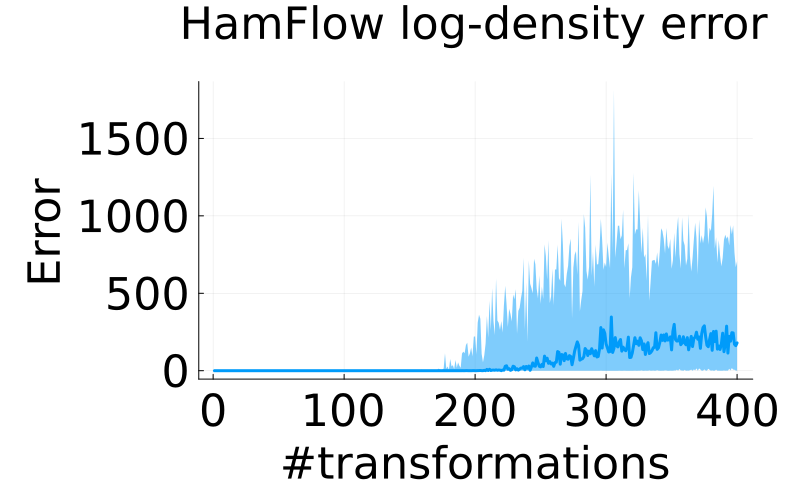}}
\caption{(a) Banana}
\end{subfigure}
\hfill
\centering 
\begin{subfigure}[b]{0.24\columnwidth} 
    \scalebox{1}{\includegraphics[width=\columnwidth]{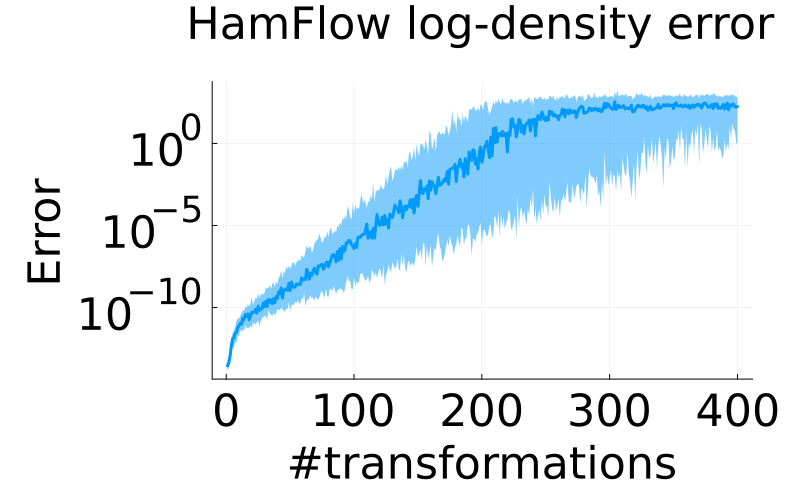}}
\caption{(b) Banana (log-scale)}
\end{subfigure}
\hfill
\centering 
\begin{subfigure}[b]{0.24\columnwidth} 
    \scalebox{1}{\includegraphics[width=\columnwidth]{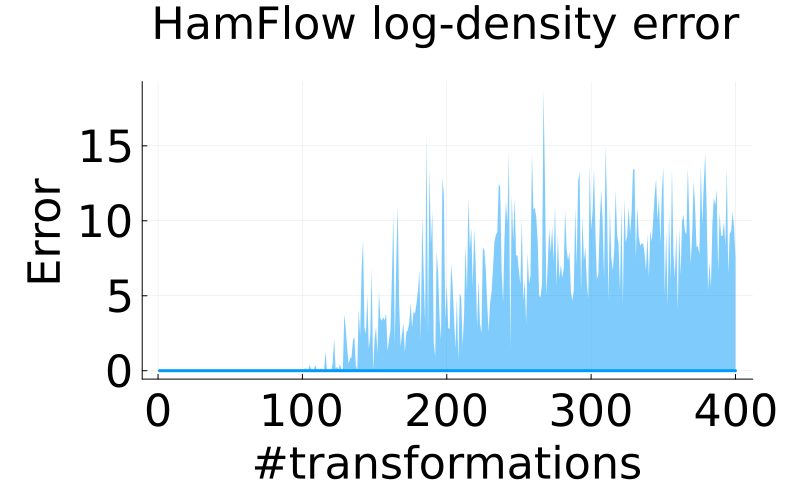}}
\caption{(c) Cross}
\end{subfigure}
\hfill
\centering 
\begin{subfigure}[b]{0.24\columnwidth} 
    \scalebox{1}{\includegraphics[width=\columnwidth]{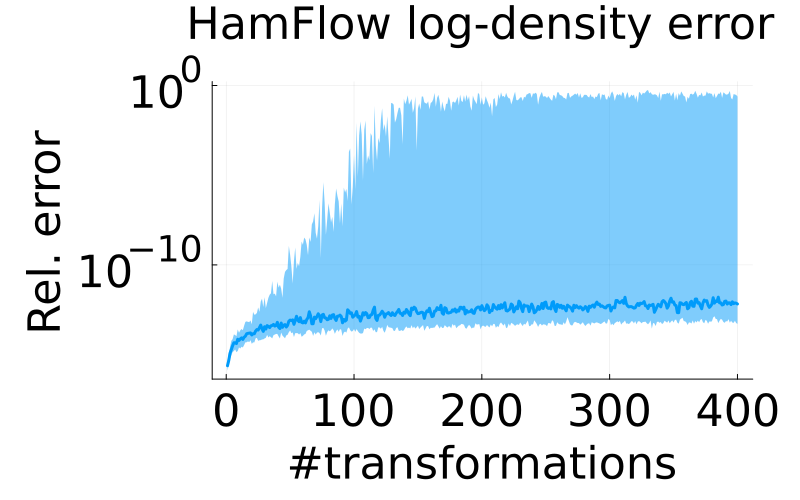}}
    \caption{(d) Cross (log-scale)}
\end{subfigure}
\hfill
\caption{
    HamFlow log-density evaluation error on synthetic examples. 
Log-densities are assessed at positions of 100 independent samples from the
target distribution. (b) and (d) respectively display (a) and (c) in a
logarithmic scale for a better illustration of the growth of error over
increasing flow length.
The lines indicate the median, and error regions indicate $25^\text{th}$ to
$75^\text{th}$ percentile from the $100$ evaluations.
} \label{fig:hamflowlpdf}
\end{figure*}

However, unlike the small log-density evaluation error we observed in the
MixFlow experiments, HamFlow's log-density shows significant issues because of
the numerical errors of backward orbits computation.
In \cref{fig:hamflowlpdf}, we measure the numerical error for the log-density
evaluation of HamFlow at 100 \iid points from the target distribution. As we add
more flow layers, the error grows quickly. For the banana example, the
log-density evaluation error can even go above $1000$.
This raises two main questions:
(1) Why is the ELBO estimation error different from the log-density evaluation error in HamFlow?
(2) Why is the scaling of the log-density evaluation error different between MixFlow and HamFlow?
Our theory presented in \cref{sec:theory} and \cref{sec:hamflowtheory} 
offers explanation to these questions.

\begin{figure*}[t!]
\centering 
\begin{subfigure}[b]{0.24\columnwidth} 
    \scalebox{1}{\includegraphics[width=\columnwidth]{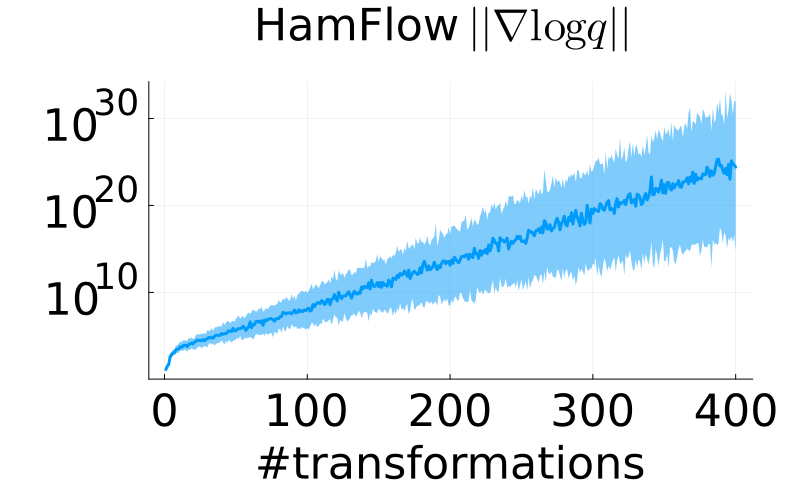}}
\caption{(a) Banana} \label{fig:banana_lq}
\end{subfigure}
\hfill
\centering 
\begin{subfigure}[b]{0.24\columnwidth} 
    \scalebox{1}{\includegraphics[width=\columnwidth]{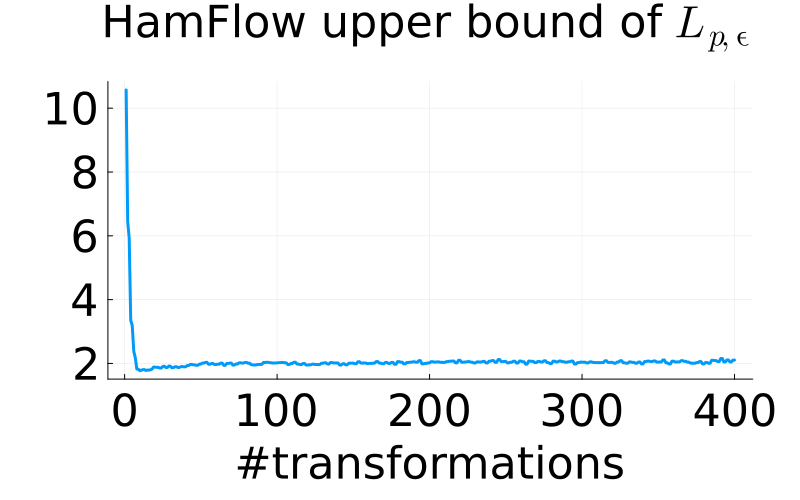}}
\caption{(c) Banana}\label{fig:banana_lp}
\end{subfigure}
\hfill
\centering 
\begin{subfigure}[b]{0.24\columnwidth} 
    \scalebox{1}{\includegraphics[width=\columnwidth]{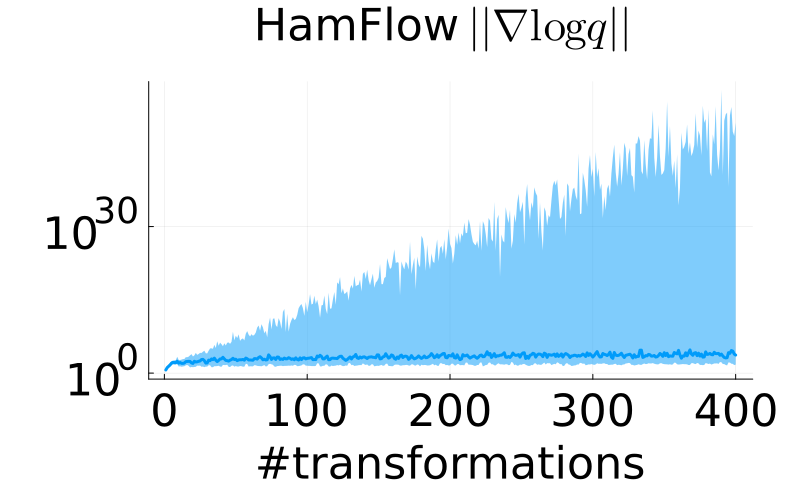}}
\caption{(a) Cross} \label{fig:cross_lq}
\end{subfigure}
\hfill
\centering 
\begin{subfigure}[b]{0.24\columnwidth} 
    \scalebox{1}{\includegraphics[width=\columnwidth]{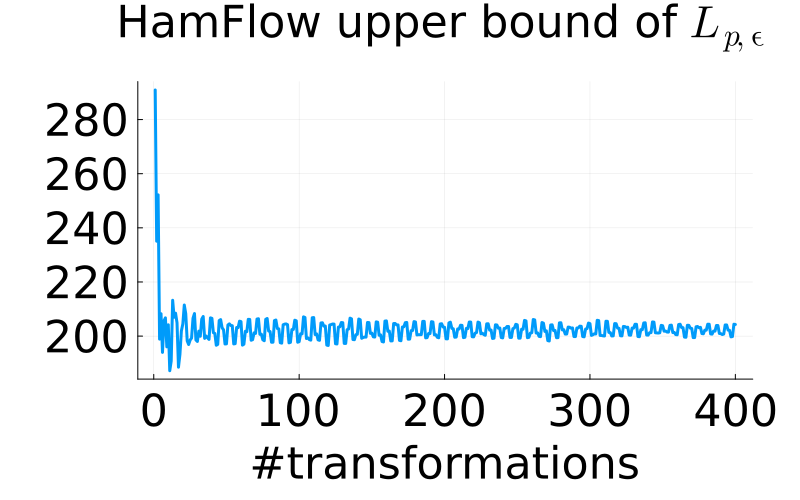}}
\caption{(c) Cross}\label{fig:cross_lp}
\end{subfigure}
\hfill
\caption{
    Comparison of HamFlow's $L_{q, \eps}$ and $\EE[L_{\bp, \eps}(\hX_N)]$ on
    the banana and cross distributions. \cref{fig:banana_lq,fig:cross_lq} plot $\|\nabla \log q(x)\|$
    against flow length $N$; the lines indicate the median, and error regions indicate $25^\text{th}$ to
$75^\text{th}$ percentile from \iid 100 independent draws of $x$ from $\bp$.
    \cref{fig:banana_lp,fig:cross_lp} plot estimated upper bounds of
    $\EE[L_{\bp, \eps}(\hX_N)]$ (averaging over $100$ independent draws from $q$) against flow length
    $N$. The choice and detailed derivation of the upper bounds of
    $L_{\bp,\eps}$ can be found in \cref{sec:lpbounds}
}
    \label{fig:lipconstant}
\end{figure*}

To address the first question, we compare the bounds for HamFlow on log-density
evaluation (\cref{cor:hamflowlpdf}) and ELBO estimation
(\cref{cor:hamflowelbo}). The main difference between these bounds is in the
local Lipschitz constants.
The error in log-density evaluation is based on $L_{q, \eps}$ while the error in
ELBO estimation is based on $\EE[L_{\bp, \eps}]$.
These two quantities--- $L_{q, \eps}$ and $\EE[L_{\bp, \eps}]$---relate to the smoothness of the
learned log-density of HamFlow and log-density of the target. 
\cref{fig:banana_hamflow_truelpdf,fig:banana_hamflow_lpdfest,fig:cross_hamflow_true,fig:cross_hamflow_lpdfest} 
display $\log p$ and $\log q$  side-by-side for both synthetic examples. 
Indeed, in both cases, $\log q$ markedly deviates from $\log p$ and presents 
drastic fluctuations, indicating that $L_{q, \eps}$ can be substantially larger than
$\EE[L_{\bp, \eps}]$.
\cref{fig:lipconstant} further provides a quantitative comparison on how $L_{q,
\eps}$ and $\EE[L_{\bp, \eps}]$ change as we increase the flow length.
Note that since both $L_{q, \eps}$ and $L_{\bp, \eps}$ are intractable in
general, we instead focus on the scaling of $\|\nabla \log q\|$ (which is a
lower bound of $L_{q, \eps}$) and an upper bound of $\EE[L_{\bp, \eps}]$ (derived from
case-by-case analysis for both the banana and cross distributions; see
\cref{sec:lpbounds} for detailed derivation).
Results show that in both synthetic examples, $L_{q, \eps}$ can increase exponentially over increasing flow
length and is eventually larger than $\EE[L_{\bp, \eps}]$ by $30$ orders of
magnitude, while $\EE[L_{\bp, \eps}]$ does not grow and remains within a
reasonable scale.

Similarly, 
the differences in log-density errors between MixFlow and HamFlow come from how
smooth the learned log-density is in each method,  
according to the error bounds presented in \cref{thm:nflpdf,thm:mflpdf}. 
\cref{fig:sync_mf_lpdfs} visualizes the learned log-densities of MixFlow, which
matches closely to the actual target distribution, and has a smoother output
than HamFlow. 
This matches what was noted in \citet[Figure 3]{Xu22mixflow}, where MixFlow was
shown to approximate the target density better than standard normalizing flows.

\begin{figure*}[t!]
\centering 
\begin{subfigure}[b]{0.32\columnwidth} 
    \scalebox{1}{\includegraphics[width=\columnwidth]{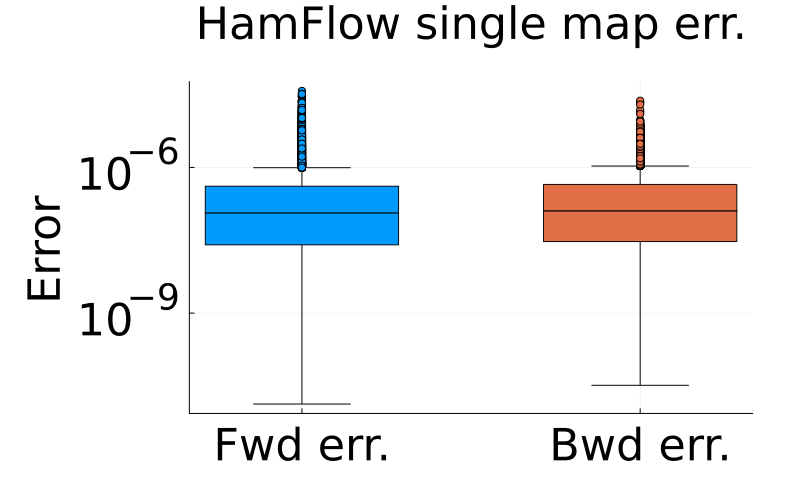}}
    \caption{(a) Banana} \label{fig:hamflow_banana_delta}
\end{subfigure}
\hfill
\centering 
\begin{subfigure}[b]{0.32\columnwidth} 
    \scalebox{1}{\includegraphics[width=\columnwidth]{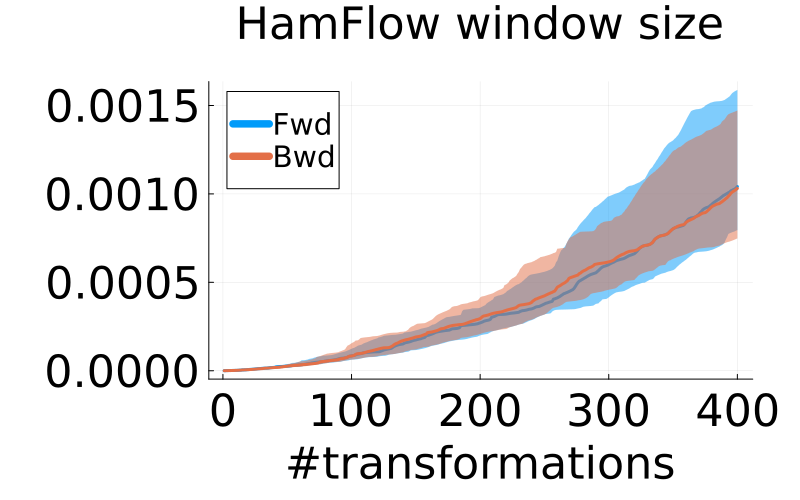}}
\caption{(b) Banana} \label{fig:hamflow_banana_window}
\end{subfigure}
\hfill
\centering 
\begin{subfigure}[b]{0.32\columnwidth} 
    \scalebox{1}{\includegraphics[width=\columnwidth]{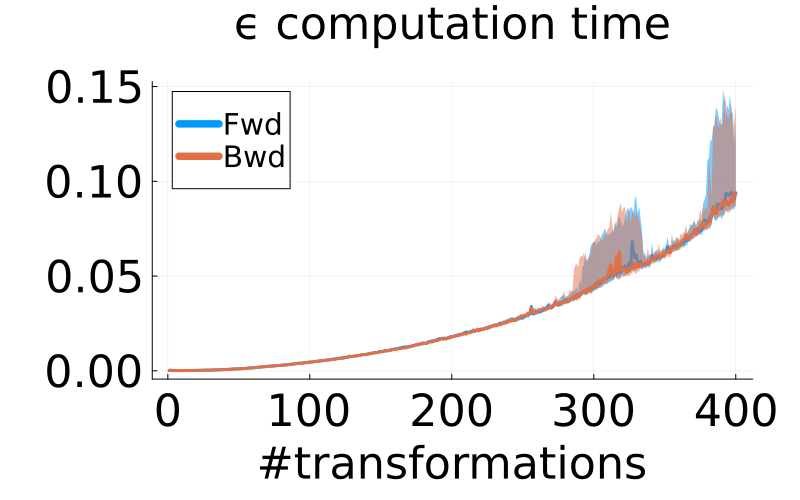}}
\caption{(c) Banana}  \label{fig:hamflow_banana_time}
\end{subfigure}
\hfill
\centering
\begin{subfigure}[b]{0.32\columnwidth} 
    \scalebox{1}{\includegraphics[width=\columnwidth]{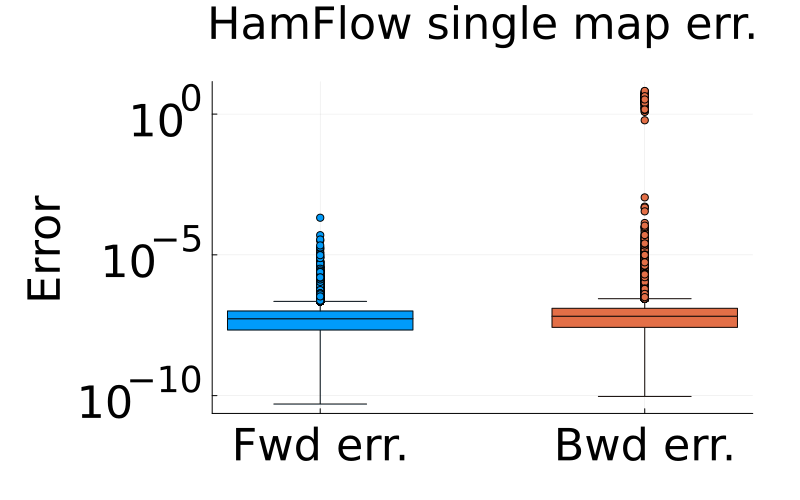}}
\caption{(d) Cross} \label{fig:hamflow_cross_delta}
\end{subfigure}
\hfill
\centering 
\begin{subfigure}[b]{0.32\columnwidth} 
    \scalebox{1}{\includegraphics[width=\columnwidth]{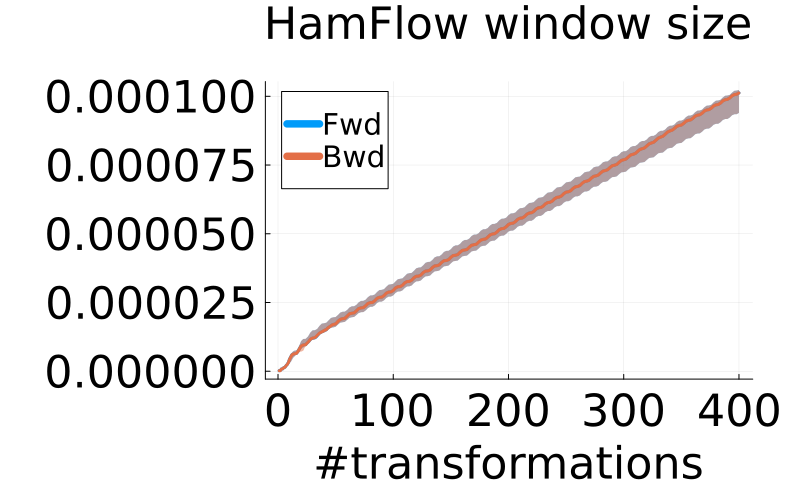}}
\caption{(e) Cross} \label{fig:hamflow_cross_window}
\end{subfigure}
\hfill
\centering 
\begin{subfigure}[b]{0.32\columnwidth} 
    \scalebox{1}{\includegraphics[width=\columnwidth]{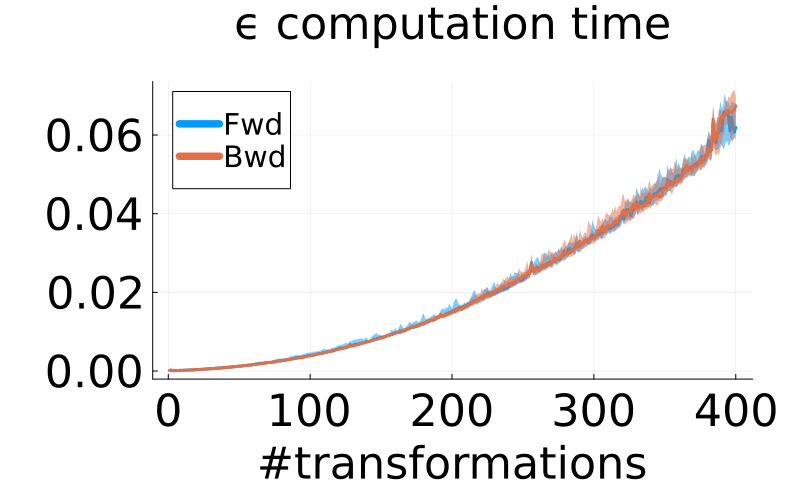}}
\caption{(f) Cross} \label{fig:hamflow_cross_time}
\end{subfigure}
\hfill
\caption{HamFlow forward (\texttt{fwd}) and backward (\texttt{bwd}) single transformation
error $\delta$ on both synthetic examples (Figure (a) and (d)), shadowing window size $\eps$ over
increasing flow length (Figure (b) and (e)), and shadowing window computation time
(Figure (c) and (f)). Figure. (a) and (d) show the distribution of forward error  
$\left\{\|F^n x_0 - \hF\circ F^{n-1} x_0\|: n = 1, \dots, N\right\}$ (\texttt{Fwd err.}) 
and backward error
$\left\{ \|F^{n} x_0 - \hB \circ F^{n+1} x_0\|: n = 1, \dots, N\right\}$
(\texttt{Bwd err.}).
These errors are computed along precise forward trajectories that originate at
$x_0$, which is sampled 50 times independently and identically from the
reference distribution $q_0$. $N$ here denotes the flow length for each example.
Figure (b) and (e) show $\eps$ over increasing flow length. Figure (c) and (f)
show the wall time of computing shadowing window over increasing flow length. 
The lines in Figure (b)--(c) and (e)--(f) indicate the median, and error regions
indicate $25^\text{th}$ to $75^\text{th}$ percentile from $50$ runs. 
} \label{fig:hamflowshadowing}
\end{figure*}

Finally, \cref{fig:hamflowshadowing} presents the single step evaluation error
$\delta$ of HamFlow and the size of the shadowing window $\eps$ as the flow length $N$
increases. Results show that for both synthetic examples, $\delta$ is
approximately $10^{-7}$, hence we fix $\delta$ to be this value when computing
the shadowing window $\eps$. Compared to the significant orbit evaluation error
when $N$ is large, the size of $\eps$ remains small for both the forward and
backward orbits (in the scale of $10^{-4}$ to $10^{-3}$ when $N = 400$). And
more importantly, the size of $\eps$ grows less drastically than the orbit
computation error with the flow length---roughly quadratically in the banana
example, and linearly in the cross example, which is similar to what is observed
in the MixFlow experiments.

\subsection{Upper bounds of $L_{\bp, \eps}$ for synthetic examples}
\label{sec:lpbounds}
Recall that we denote $\bp(x) \propto p([x]_{1:2})\cdot
\exp(-\frac{1}{2}[x]_{3:4}^T [x]_{3:4})$ the augmented target and $p$ denotes
the 2d-synthetic target we are interested.

Notice that 
\[
    L_{\bp, \eps}(x) =\sup_{\|y - x\|\leq \eps} \left\|\nabla \log p([y]_{1:2})
    - [y]_{3:4}\right\| \leq L_{p, \eps}([x]_{1:2}) + \|[x]_{3:4}\| + \eps, 
\]
where $L_{p, \eps}$ requires a case-by-case analysis.
Therefore, in this section, we focus on bounding the local Lipschitz constant
for $\log p$.

\bprop\label{prop:banana} 
For the 2-dimensional banana distribution, i.e., 
\[
    p(x) \propto \exp\left( -\frac{1}{2} \frac{x_1^2}{\sigma^2} -
    \frac{1}{2}\left(x_2 + bx_1^2 - \sigma^2 b\right) \right), 
\]
we have that for all $x =:\bbmat x_1\\ x_2\ebmat \in \reals^2$,
\[
L_{p, \eps}(x) \leq \|\nabla \log p(x)\| +
\eps \cdot \left( 2b|x_1| + \max\left\{ \left|6b^2 x_1^2 + 2b x_2 - 2\sigma^2 b^2 +
    \frac{1}{\sigma^{2}}\right|, 1\right\}  + 2b\max \left\{ 6b|x_1| + 6b(\eps +
    \eps^2), 1 \eps\right\}\right)
\]
\eprop
\bprfof{\cref{prop:banana}}
By the mean value theorem, 
\[
    L_{p, \eps}(x) 
    &\leq \|\nabla \log p(x)\| + \sup_{\|y-x\|\leq \eps} \|\nabla \log p(y) -
    \nabla \log p (x)\| \\
   &\leq \|\nabla \log p(x)\| + \sup_{\|y - x\|\leq \eps}\|\nabla^2 \log p(y)\| \cdot \eps.
\]
The proof will then complete by showing that
\[ \label{eq:bananahessian}
\sup_{\|y - x\|\leq \eps}\|\nabla^2 \log p(y)\| \leq
2b|x_1| + \max\left\{ \left|6b^2 x_1^2 + 2b x_2 - 2\sigma^2 b^2 +
\frac{1}{\sigma^{2}}\right|, 1\right\}  + 2b\max \left\{ 6b|x_1| + 6b(\eps +
\eps^2),  \eps\right\}.
\]
To verify \cref{eq:bananahessian}, note that 
\[
    \nabla^2\log p(x) = -\bbmat 
    6b^2 x_1^2 +2bx_2- 2\sigma^2 b^2 + \frac{1}{\sigma^2} & 2bx_1\\
    2bx_1 & 1
    \ebmat
\]
The Gershgorin circle theorem then yields that 
\[ 
    \|\nabla^2\log p(x)\| \leq 2b|x_1| + \max\left\{ \left|6b^2 x_1^2 + 2b x_2 - 2\sigma^2 b^2 +
    \frac{1}{\sigma^{2}}\right|, 1\right\}.
\]
Additionally, we have
\[
    \sup_{\|y_1 - x_1\|\leq \eps} |y_1| \leq |x_1| + \eps
\]
and
\[
\sup_{\|y - x\|\leq \eps} \left|6b^2 y_1^2 + 2b y_2 - 2\sigma^2 b^2 + \frac{1}{\sigma^{2}}\right|
\leq \left|6b^2 x_1^2 + 2b x_2 - 2\sigma^2 b^2 +
    \frac{1}{\sigma^{2}}\right| + 2b\max \left\{ 6b|x_1| + 6b(\eps + \eps^2),
    \eps\right\}.
\]
The proof is then completed. 
\eprfof

\bprop \label{prop:cross}
For the cross distribution, a 4-component Gaussian mixture of the form 
\[
    p(x) = & \frac{1}{4}\distNorm\left(x|  \begin{bmatrix} 0 \\ 2 \end{bmatrix}, 
				\begin{bmatrix} 0.15^2 & 0 \\ 0 & 1 \end{bmatrix} \right) + 
	\frac{1}{4}\distNorm\left(x| \begin{bmatrix} -2 \\ 0 \end{bmatrix}, 
				\begin{bmatrix} 1 & 0 \\ 0 & 0.15^2  \end{bmatrix} \right)\\ 
	&+ \frac{1}{4}\distNorm\left(x| \begin{bmatrix} 2 \\ 0 \end{bmatrix}, 
	\begin{bmatrix} 1 & 0 \\ 0 & 0.15^2  \end{bmatrix} \right) + 
	\frac{1}{4}\distNorm\left(x| \begin{bmatrix} 0 \\ -2 \end{bmatrix}, 
	\begin{bmatrix} 0.15^2 & 0 \\ 0 & 1 \end{bmatrix} \right),
\]
we have that for all $x \in \reals^2$,
\[
    L_{p, \eps}(x) \leq \frac{1}{0.15^2}\left( 2 + \|x\| +\eps \right).
\]
\eprop

\bprfof{\cref{prop:cross}}
For notational convenience, denote the density of 4 Gaussian component by 
$p_1, p_2, p_3, p_4$, hence
\[
    \log p(x) = \log\left\{ p_1(x) + p_2(x) + p_3(x) + p_4(x)\right\} - \log 4.
\]
Examining $\nabla \log p(x)$ directly yields that  for all $x\in \reals^2$,
\[
    \nabla \log p(x) 
    &= \frac{1}{p_1(x) + p_2(x) + p_3(x) + p_4(x)}\sum_{i=1}^{4} p_i(x)\Sigma_i^{-1}(\mu_i - x) \\
		     &\in \conv\left\{ \Sigma_i^{-1}(\mu_i - x): i = 1, 2, 3, 4\right\}.
\]
Here $\mu_i$ and $\Sigma_i$ denotes the Gaussian mean and covariance of $p_i$.
Then by Jensen's inequality, 
\[
    \sup_{\|y -x\|\leq \eps} \|\nabla \log p(y)\| 
&    \leq \max\left\{ \sup_{\|y - x\|\leq \eps}\|\Sigma_i^{-1}(\mu_i - y)\|: i =
1, 2, 3, 4\right\} \\ 
& \leq \max_{i\in [4]}\left\{ \|\Sigma_i^{-1}\|\cdot\left(\|\mu_i\| + \sup_{\|y -x\|\leq \eps}\|y\|\right) \right\}\\ 
&\leq \frac{1}{0.15^2}\left( 2 + \|x\| +\eps \right).
\]
\eprfof

\captionsetup[subfigure]{labelformat=empty}

\captionsetup[subfigure]{labelformat=empty}

\captionsetup[subfigure]{labelformat=empty}

\captionsetup[subfigure]{labelformat=empty}

\captionsetup[subfigure]{labelformat=empty}

\captionsetup[subfigure]{labelformat=empty}

\captionsetup[subfigure]{labelformat=empty}

\section{Additional experimental details} \label{apdx:expts}

All experiments were conducted on a machine with an AMD Ryzen 9 3900X and 32G
of RAM.

\subsection{Model description} \label{apdx:model}
The two synthetic distributions tested in this experiment were
\bitems 
\item  the banana distribution \citep{haario2001adaptive}:
\[
y = \begin{bmatrix}y_1 \\ y_2\end{bmatrix} \distas 
\distNorm\left( 0, \begin{bmatrix}100 & 0 \\ 0 & 1\end{bmatrix} \right), \quad 
x = \begin{bmatrix} y_1 \\ y_2 + by_1^2 - 100b \end{bmatrix}, \quad b = 0.1;
\]
\item  a cross-shaped distribution: a Gaussian mixture of the form
\[
x &\distas \frac{1}{4}\distNorm\left( \begin{bmatrix} 0 \\ 2 \end{bmatrix}, 
				\begin{bmatrix} 0.15^2 & 0 \\ 0 & 1 \end{bmatrix} \right) + 
	\frac{1}{4}\distNorm\left( \begin{bmatrix} -2 \\ 0 \end{bmatrix}, 
				\begin{bmatrix} 1 & 0 \\ 0 & 0.15^2  \end{bmatrix} \right)\\ 
	&+ \frac{1}{4}\distNorm\left( \begin{bmatrix} 2 \\ 0 \end{bmatrix}, 
	\begin{bmatrix} 1 & 0 \\ 0 & 0.15^2  \end{bmatrix} \right) + 
	\frac{1}{4}\distNorm\left( \begin{bmatrix} 0 \\ -2 \end{bmatrix}, 
	\begin{bmatrix} 0.15^2 & 0 \\ 0 & 1 \end{bmatrix} \right);
\]
\eitems

The two real-data experiments are described below.
\paragraph{Linear regression.} We consider a Bayesian linear regression problem 
where the model takes the form
\[
\textbf{Lin. Reg.: }&
\beta \distiid \distNorm(0, 1),  \log\sigma^2 \distiid \distNorm(0,1),
\quad y_j \given \beta, \sigma^2 \distind \distNorm\left(x_j^T\beta, \sigma^2\right),
\label{eq:linreg_model}
\]
where $y_j$ is the response and $x_j\in\reals^p$ is the feature vector for data point $j$.
For this problem, we use the Oxford Parkinson's Disease Telemonitoring Dataset
\citep{Tsanas2010AccurateTO}. 
The original dataset is available at
\url{http://archive.ics.uci.edu/dataset/189/parkinsons+telemonitoring}, 
composed of $16$ biomedical voice measurements from $42$ patients with
early-stage Parkinson's disease. The goal is to use these voice measurements,
as well as subject age and gender information, to predict the total UPDRS scores.
We standardize all features and subsample the original dataset down to $500$
data points.
The posterior dimension of the linear regression inference problems is $21$.

\paragraph{Logistic regression.} We then consider a Bayesian hierarchical logistic regression:
\[
\textbf{Logis. Reg.: }&
\alpha \sim \distGam(1, 0.01), \beta \mid \alpha  \sim \distNorm(0, \alpha^{-1}
I), \quad y_{j} \given \beta \distind \distBern\left(\frac{1}{1+e^{-x_{j}^{T} \beta}}\right),
\label{eq:logreg_model}
\]
We use a bank marketing dataset \citep{Moro14} downsampled to $400$ data
points. The original dataset is  
available at \url{https://archive.ics.uci.edu/ml/datasets/bank+marketing}. 
The goal is to use client information to predict subscription to a term deposit. 
We include 8 features from dataset: client age, marital status, balance, housing loan status,
duration of last contact, number of contacts during campaign, number of days
since last contact, and number of contacts before the current campaign. For each
of the binary variables (marital status and housing loan status), all unknown
entries are removed. All included features are standardized.
The resulting posterior dimension of the logistic regression problem is $9$.

\subsection{MixFlow parameter settings} \label{apdx:mixflowparameters}

For all four examples, we use the uncorrected Hamiltonian MixFlow map $F$
composed of a discretized Hamiltonian dynamic simulation (via leapfrog
integrator) followed by a sinusoidal momentum pseudorefreshment as used in
\citet{Xu22mixflow} (see \cite[Section 5, Appendices E.1 and E.2]{Xu22mixflow} for details).  
However, we use a Gaussian momentum distribution, which is standard in Hamiltonian Monte Carlo methods
\cite{Neal11} as opposed to the Laplace momentum used previously in MixFlows
\cite{Xu22mixflow}.
In terms of the settings of leapfrog integrators for each target,  
we used 200 leapfrog steps of size 0.02 and 
60 leapfrog steps of size 0.005 for the banana and cross target distributions respectively, 
and used $40$ leapfrog steps of size $0.0006$
and $50$ leapfrog steps of size $0.002$ for the linear regression and logistic
regression examples, respectively.
The reference distribution $q_0$ for each target is chosen to be the mean-field
Gaussian approximation as used in \cite{Xu22mixflow}. 

As for the \texttt{NUTS} used for assessing the log-density evaluation quality
on two real data examples, we use the Julia package
\texttt{AdvancedHMC.jl} \citep{xu2020advancedhmc} with all default settings.
\texttt{NUTS} is initialized with the learned
mean of the mean-field Gaussian approximation $q_0$, 
and burns in the first $10{,}000$ samples before collecting samples; 
the burn-in samples are used for adapting the
hyperparameters of \texttt{NUTS} with target acceptance rate set to be $0.7$.

\subsection{Additional MixFlow results} \label{apdx:moreresults}

\captionsetup[subfigure]{labelformat=empty}
\begin{figure*}[t!]
\centering 
\begin{subfigure}[b]{0.24\columnwidth} 
    \scalebox{1}{\includegraphics[width=\columnwidth]{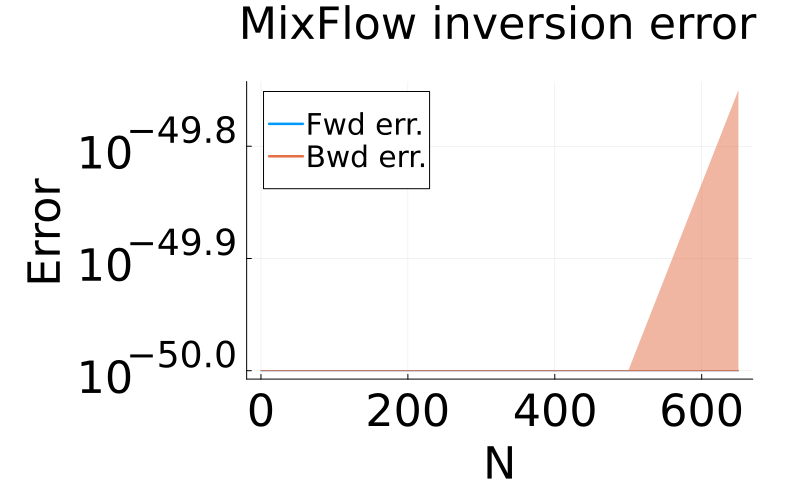}}
    \caption{(a) Banana \label{fig:banana_inverr}}
\end{subfigure}
\hfill
\centering 
\begin{subfigure}[b]{0.24\columnwidth} 
    \scalebox{1}{\includegraphics[width=\columnwidth]{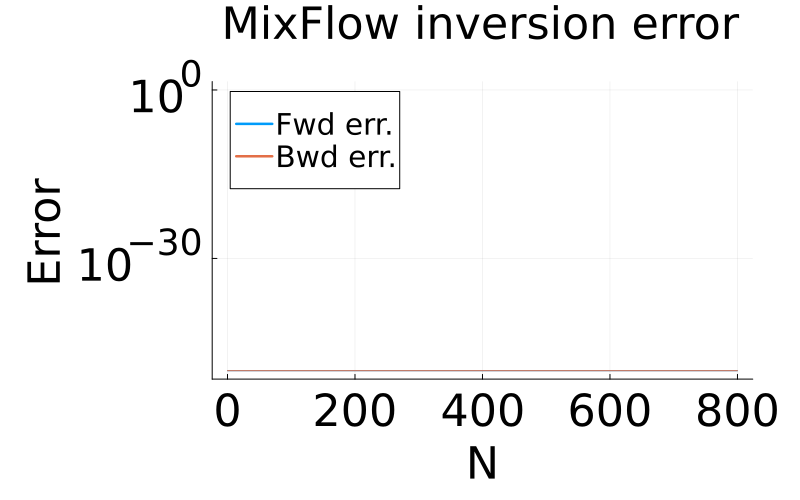}}
\caption{(b) Cross \label{fig:cross_inverr}}
\end{subfigure}
\hfill
\centering 
\begin{subfigure}[b]{0.24\columnwidth} 
    \scalebox{1}{\includegraphics[width=\columnwidth]{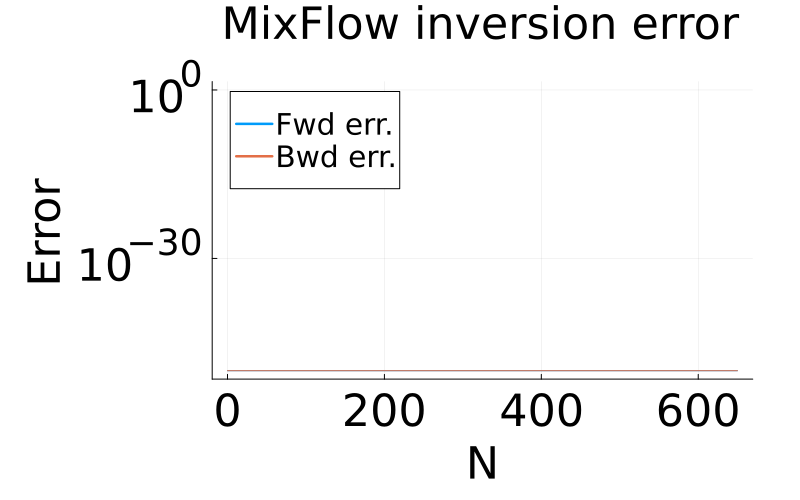}}
\caption{(c) Linear regression \label{fig:linreg_inverr}}
\end{subfigure}
\hfill
\centering 
\begin{subfigure}[b]{0.24\columnwidth} 
    \scalebox{1}{\includegraphics[width=\columnwidth]{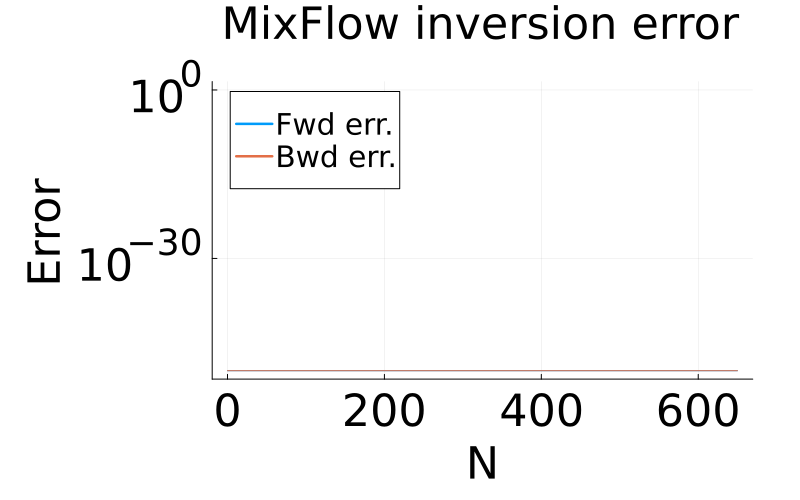}}
\caption{(d) Logistic regression \label{fig:logreg_inverr}}
\end{subfigure}
\hfill
\caption{MixFlow inversion error when using 2048-bit \texttt{BigFloat}
representation. Verticle axis shows the reconstruction error $\|B^N\circ F^N(x_0) - x_0\|$
(\texttt{Fwd})
and $\|F^N\circ B^N(x_0) - x_0\|$ (\texttt{Bwd}) for the two synthetic examples
(\cref{fig:banana_inverr,fig:cross_inverr}) and the two real data examples
(\cref{fig:linreg_inverr,fig:logreg_inverr}). $F$ and $B$ are implemented
using 2048-bit \texttt{BigFloat} representation, and $x_0$ is sampled from the
reference distribution $q_0$. The lines indicate the median, and the upper and
lower quantiles over $10$ independent runs. It can be seen that in all four
examples, the inversion error of MixFlow is ignorable when using 2048-bit
computation.} 
\label{fig:mfbigfloaterr}
\end{figure*}

\captionsetup[subfigure]{labelformat=empty}
\begin{figure*}[t!]
\centering 
\begin{subfigure}[b]{0.24\columnwidth} 
    \scalebox{1}{\includegraphics[width=\columnwidth]{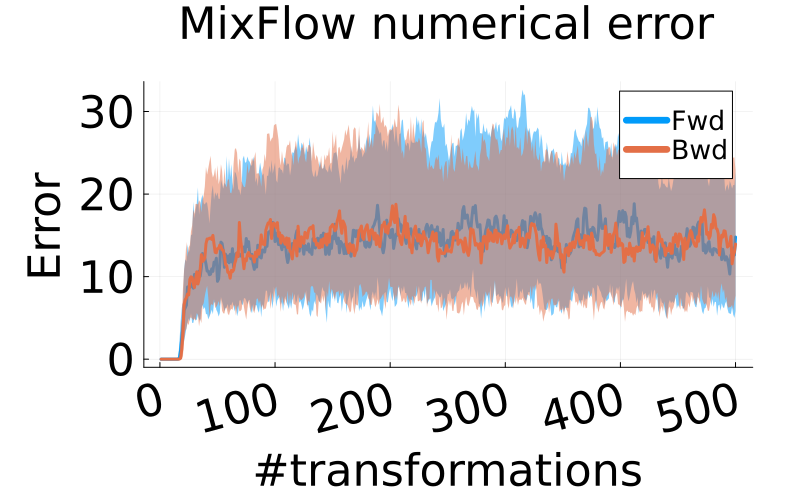}}
\caption{(a) Banana}
\end{subfigure}
\hfill
\centering 
\begin{subfigure}[b]{0.24\columnwidth} 
    \scalebox{1}{\includegraphics[width=\columnwidth]{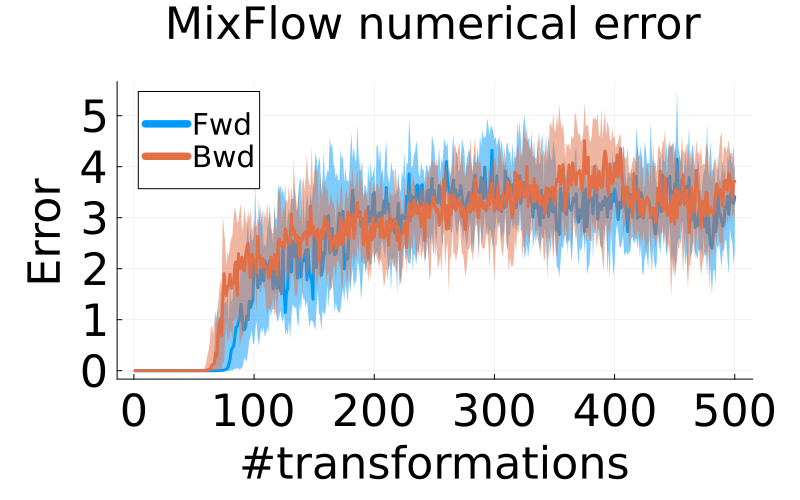}}
\caption{(b) Cross }
\end{subfigure}
\hfill
\centering 
\begin{subfigure}[b]{0.24\columnwidth} 
    \scalebox{1}{\includegraphics[width=\columnwidth]{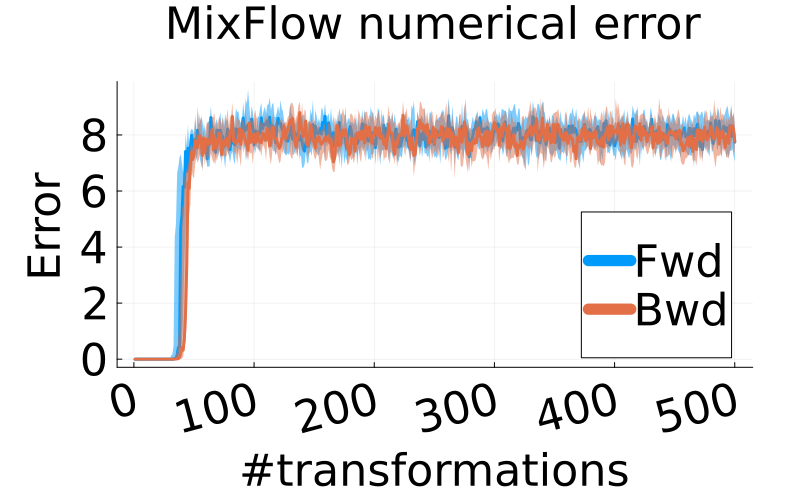}}
\caption{(c) Linear regression}
\end{subfigure}
\hfill
\centering 
\begin{subfigure}[b]{0.24\columnwidth} 
    \scalebox{1}{\includegraphics[width=\columnwidth]{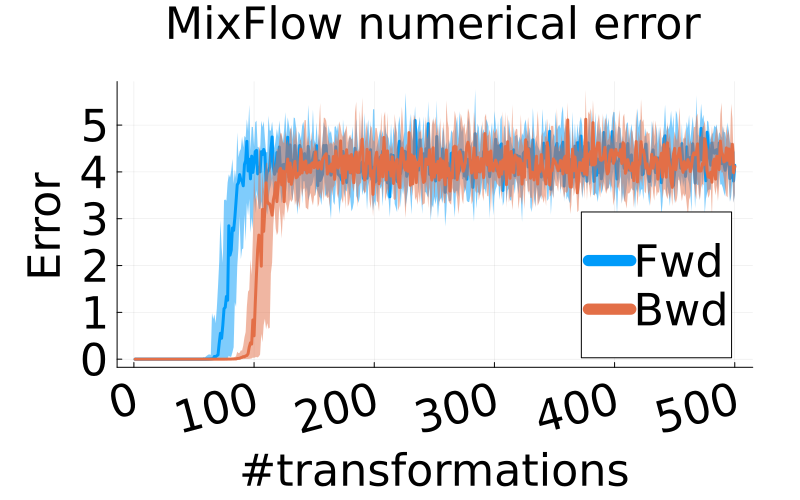}}
\caption{(d) Logistic regression}
\end{subfigure}
\hfill
\caption{MixFlow forward (\texttt{fwd}) and backward (\texttt{bwd}) orbit
errors on both synthetic examples and real data examples. 
(a)---(d) show the median and upper/lower quartile 
forward error $\|F^k x - \hF^k x\|$ and backward error $\|B^k x - \hB^{k}x\|$
comparing $k$ transformations of the forward exact/approximate maps $F$, $\hF$
or backward exact/approximate maps $B$, $\hB$.
Note that \cref{fig:banana_err,fig:cross_err,fig:linreg_flow_err_log,fig:logreg_flow_err_log} 
respectively display the initial segments of (a)---(d) in a logarithmic scale, intended to provide a clearer
illustration of the exponential growth of the error.
}
\label{fig:trajerrallex}
\end{figure*}

\captionsetup[subfigure]{labelformat=empty}
\begin{figure*}[t!]
\centering 
\begin{subfigure}[b]{0.24\columnwidth} 
    \scalebox{1}{\includegraphics[width=\columnwidth]{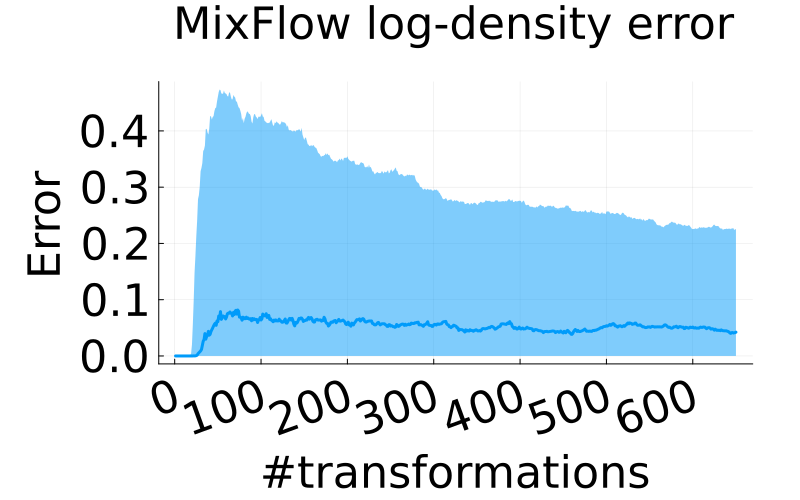}}
    \caption{(a) Banana} 
    \label{fig:banana_lpdfs_abs_err}
\end{subfigure}
\hfill
\centering 
\begin{subfigure}[b]{0.24\columnwidth} 
    \scalebox{1}{\includegraphics[width=\columnwidth]{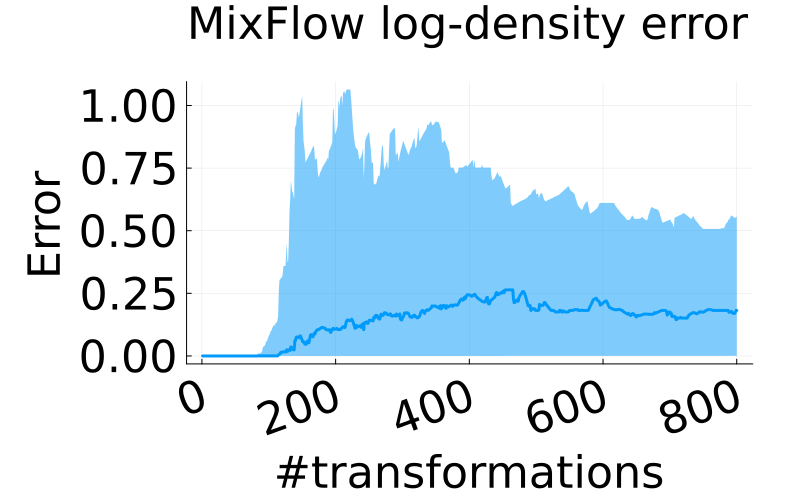}}
    \caption{(b) Cross}
    \label{fig:cross_lpdfs_abs_err}
\end{subfigure}
\hfill
\centering 
\begin{subfigure}[b]{0.24\columnwidth} 
    \scalebox{1}{\includegraphics[width=\columnwidth]{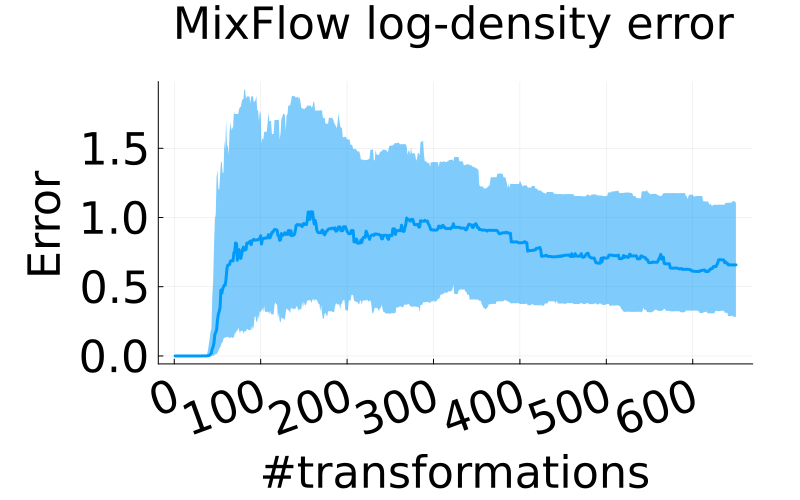}}
    \caption{(c) Lin. Reg.}
    \label{fig:linreg_lpdfs_abs_err}
\end{subfigure}
\hfill
\centering 
\begin{subfigure}[b]{0.24\columnwidth} 
    \scalebox{1}{\includegraphics[width=\columnwidth]{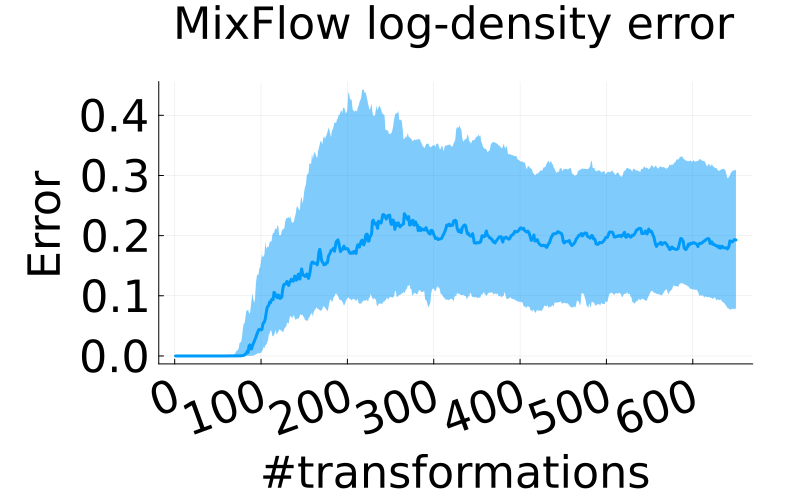}}
    \caption{(d) Log. Reg.}
    \label{fig:logreg_lpdfs_abs_err}
\end{subfigure}
\hfill
\caption{MixFlow log-density evaluation error. 
For synthetic examples (\cref{fig:banana_lpdfs_abs_err,fig:cross_lpdfs_abs_err}), 
we assessed densities at 100 evenly distributed points
within the target region (\cref{fig:banana_scatter,fig:cross_scatter}). For real
data examples (\cref{fig:linreg_lpdfs_abs_err,fig:logreg_lpdfs_abs_err}), we
evaluated densities at the locations of 100 \texttt{NUTS} samples.
The lines indicate the median, and error regions indicate $25^\text{th}$ to
$75^\text{th}$ percentile from $100$ evaluations on different positions.}
\label{fig:mflpdfserr}
\end{figure*}

\captionsetup[subfigure]{labelformat=empty}
\begin{figure*}[t!]
\centering 
\begin{subfigure}[b]{0.24\columnwidth} 
    \scalebox{1}{\includegraphics[width=\columnwidth]{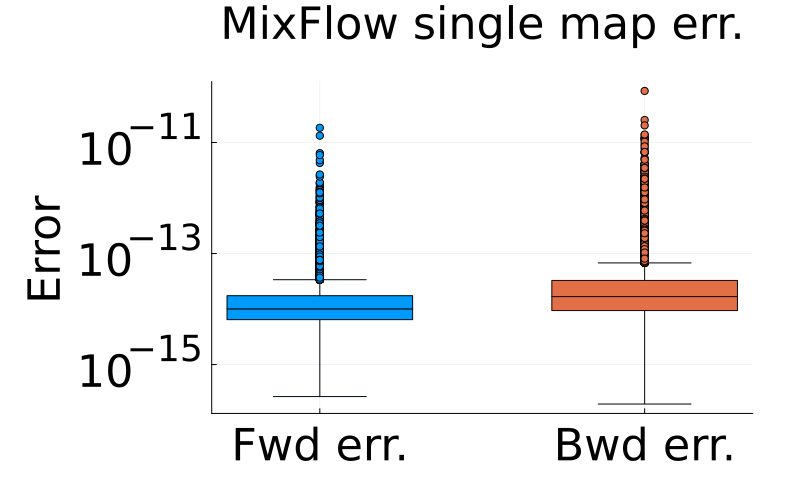}}
\caption{(a) Banana}
\end{subfigure}
\hfill
\centering 
\begin{subfigure}[b]{0.24\columnwidth} 
    \scalebox{1}{\includegraphics[width=\columnwidth]{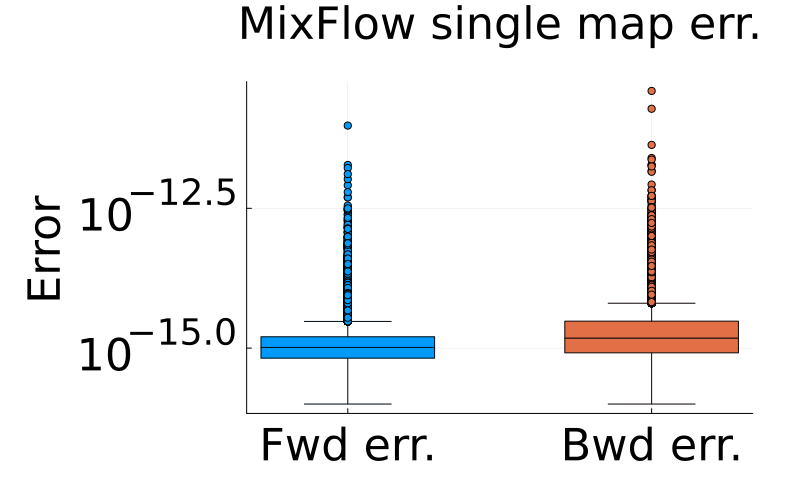}}
\caption{(b) Cross }
\end{subfigure}
\hfill
\centering 
\begin{subfigure}[b]{0.24\columnwidth} 
    \scalebox{1}{\includegraphics[width=\columnwidth]{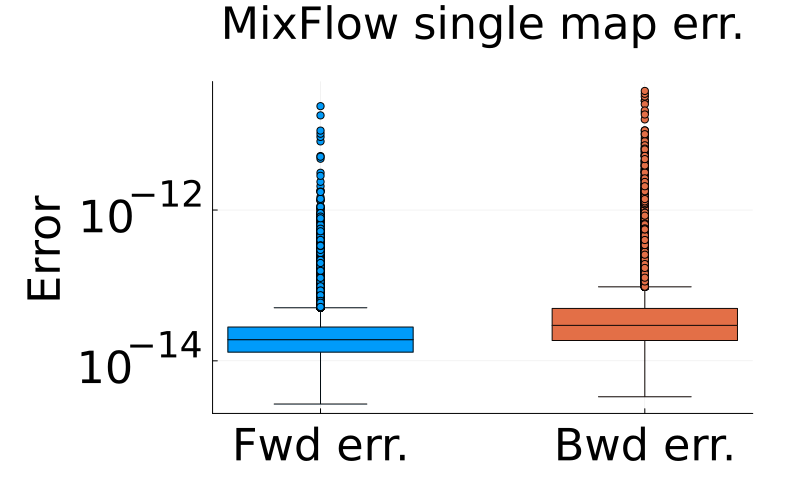}}
\caption{(c) Linear regression}
\end{subfigure}
\hfill
\centering 
\begin{subfigure}[b]{0.24\columnwidth} 
    \scalebox{1}{\includegraphics[width=\columnwidth]{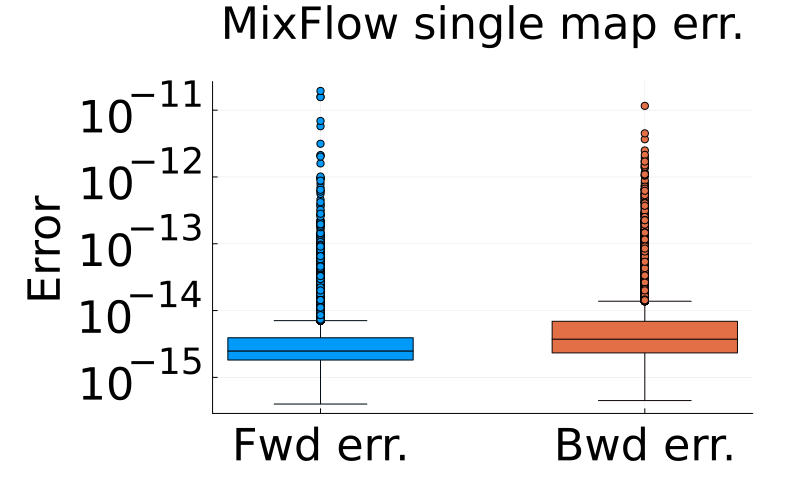}}
\caption{(d) Logistic regression}
\end{subfigure}
\hfill
\caption{
MixFlow forward (\texttt{fwd}) and backward (\texttt{bwd}) single transformation
error $\delta$ on both synthetic examples and real data examples. 
(a)---(d)  
show the distribution of forward error  
$\left\{\|F^n x_0 - \hF\circ F^{n-1} x_0\|: n = 1, \dots, N\right\}$ (\texttt{Fwd err.}) 
and backward error
$\left\{ \|F^{n} x_0 - \hB \circ F^{n+1} x_0\|: n = 1, \dots, N\right\}$
(\texttt{Bwd err.}).
These errors are computed along precise forward trajectories that originate at
$x_0$, which is sampled 50 times independently and identically from the
reference distribution $q_0$. $N$ here denotes the flow length for each example.
}
\label{fig:delta}
\end{figure*}

\captionsetup[subfigure]{labelformat=empty}
\begin{figure*}[t!]
\centering 
\begin{subfigure}[b]{0.24\columnwidth} 
    \scalebox{1}{\includegraphics[width=\columnwidth]{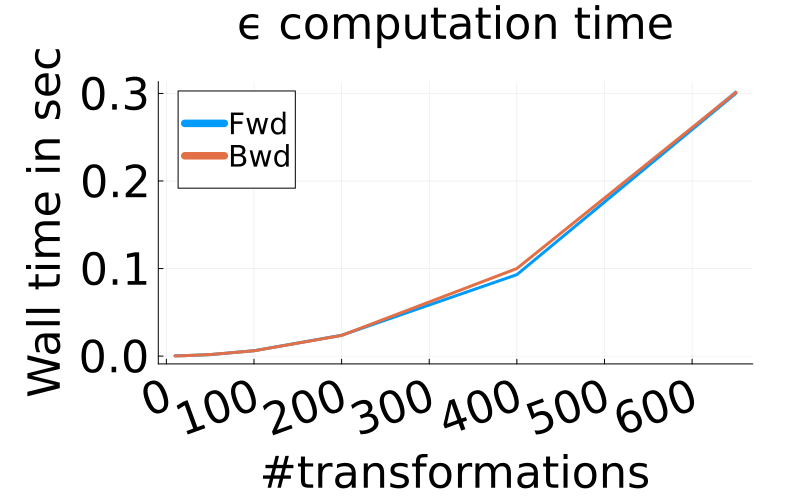}}
\caption{(a) Banana}
\end{subfigure}
\hfill
\centering 
\begin{subfigure}[b]{0.24\columnwidth} 
    \scalebox{1}{\includegraphics[width=\columnwidth]{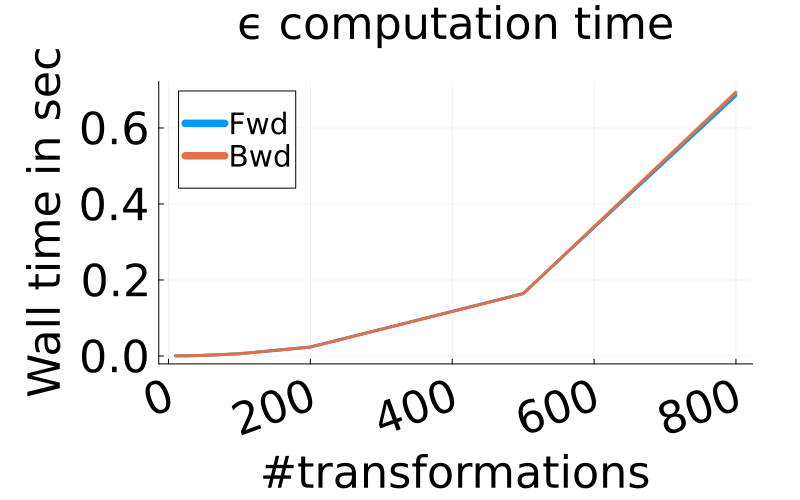}}
\caption{(b) Cross }
\end{subfigure}
\hfill
\centering 
\begin{subfigure}[b]{0.24\columnwidth} 
    \scalebox{1}{\includegraphics[width=\columnwidth]{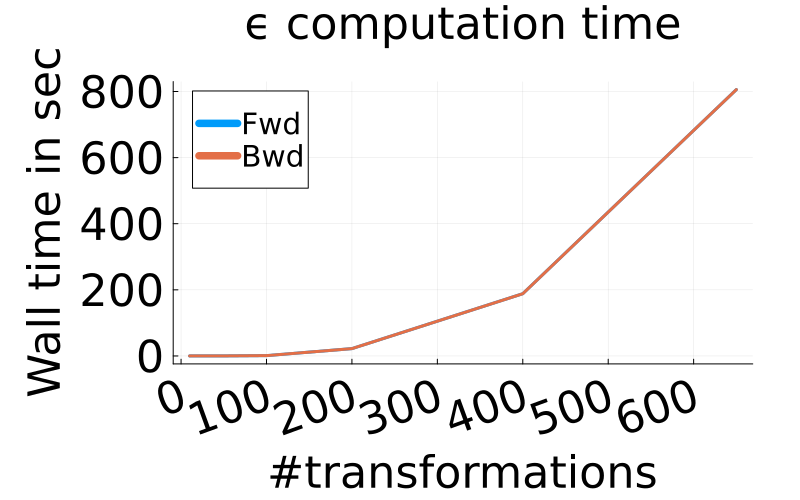}}
\caption{(c) Linear regression}
\end{subfigure}
\hfill
\centering 
\begin{subfigure}[b]{0.24\columnwidth} 
    \scalebox{1}{\includegraphics[width=\columnwidth]{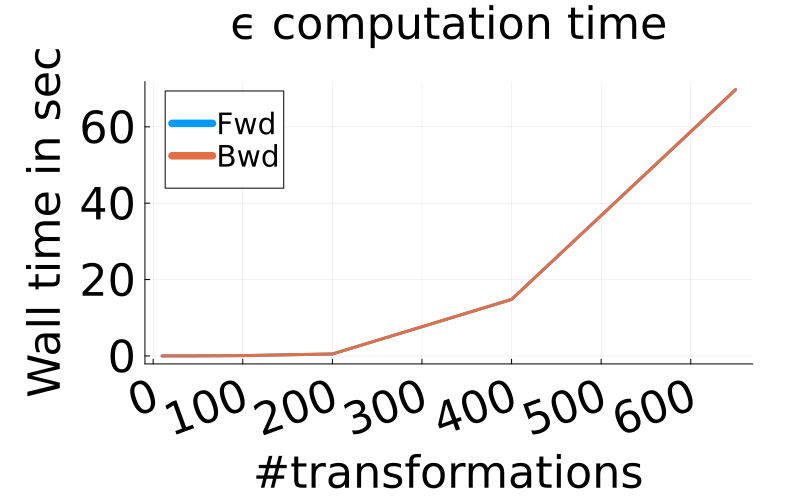}}
\caption{(d) Logistic regression}
\end{subfigure}
\hfill
\caption{
    MixFlow shadowing window computation time (wall time in second) for both the forward
    (\texttt{Fwd}) and backward (\texttt{Bwd}) orbits over increasing flow
    length.
The lines indicate the median, and error regions indicate $25^\text{th}$ to
$75^\text{th}$ percentile from $10$ runs. Note that here the Monte Carlo error
between different runs is so small that the error bar is too thin to be
visualized.
}
\label{fig:time}
\end{figure*}

\end{document}